%% file: main.tex
\documentclass{article}

\usepackage{PRIMEarxiv}

\usepackage[utf8]{inputenc} 
\usepackage[T1]{fontenc}    
\usepackage{hyperref}       
\hypersetup{hidelinks}
\usepackage{url}            
\usepackage{booktabs}       
\usepackage{amsfonts}       
\usepackage{nicefrac}       
\usepackage{microtype}      
\usepackage{lipsum}
\usepackage{fancyhdr}       
\usepackage{graphicx}       
\graphicspath{{media/}}     

\usepackage{microtype} 
\usepackage{booktabs}  
\usepackage{amsmath}
\usepackage{amsthm}
\usepackage{tikz}
\usepackage{tikzscale}
\usepackage{caption}
\usepackage{subcaption}
\usepackage{pgfplots}
\pgfplotsset{compat=1.17}
\usepackage{wrapfig}
\usepackage{algorithm}
\usepackage{algorithmic}
\usepackage{array,colortbl,multirow,multicol,booktabs,ctable}

\usetikzlibrary{patterns}

\DeclareMathOperator*{\argmax}{arg\,max}

\usepackage{natbib}
\usepackage{xspace}
\usepackage{enumitem}

\newcommand{\libname}{ARLO\@\xspace}
\newcommand{\lib}{\libname{}: Automated Reinforcement Learning Optimizer\@\xspace}
\newcommand{\eg}{\emph{e.g.}}

\DeclareRobustCommand{\eg}{e.g.,\@\xspace}                         
\DeclareRobustCommand{\ie}{i.e.,\@\xspace}

\DeclareRobustCommand{\quotes}[1]{``#1''}

\newcommand{\block}{stage\@\xspace}
\newcommand{\blocks}{stages\@\xspace}
\newcommand{\unit}{unit\@\xspace}
\newcommand{\units}{units\@\xspace}

\usepackage{mathtools}

\usepackage{array}
\newcolumntype{L}[1]{>{\raggedright\let\newline\\\arraybackslash\hspace{0pt}}m{#1}}
\newcolumntype{C}[1]{>{\centering\let\newline\\\arraybackslash\hspace{0pt}}m{#1}}
\newcolumntype{R}[1]{>{\raggedleft\let\newline\\\arraybackslash\hspace{0pt}}m{#1}}

\title{\libname{}: A Framework for Automated Reinforcement Learning
}

\author{
  Marco Mussi \\
  Politecnico di Milano \\
  \texttt{marco.mussi@polimi.it} \\
  \And
  Davide Lombarda \\
  Politecnico di Milano, ML cube \\
  \texttt{davide.lombarda@mlcube.com} \\
  \And
  Alberto Maria Metelli \\
  Politecnico di Milano \\
  \texttt{albertomaria.metelli@polimi.it} \\
  \And
  Francesco Trov\`{o} \\
  Politecnico di Milano \\
  \texttt{francesco1.trovo@polimi.it} \\
  \And
  Marcello Restelli \\
  Politecnico di Milano \\
  \texttt{marcello.restelli@polimi.it} \\
}

\begin{document}
\maketitle

\begin{abstract}
Automated Reinforcement Learning (AutoRL) is a relatively new area of research that is gaining increasing attention. The objective of AutoRL consists in easing the employment of Reinforcement Learning (RL) techniques for the broader public by alleviating some of its main challenges, including data collection, algorithm selection, and hyper-parameter tuning.
In this work, we propose a general and flexible framework, namely \lib, to construct automated pipelines for AutoRL. Based on this, we propose a pipeline for offline and one for online RL, discussing the components, interaction, and highlighting the difference between the two settings. Furthermore, we provide a Python implementation of such pipelines, released as an open-source library. Our implementation has been tested on an illustrative LQG domain and on classic MuJoCo environments, showing the ability to reach competitive performances requiring limited human intervention. We also showcase the full pipeline on a realistic dam environment, automatically performing the feature selection and the model generation tasks.
\end{abstract}


\input{chapters/01_introduction}
\input{chapters/02_preliminaries}
\input{chapters/03_framework}
\input{chapters/04_pipeline}
\input{chapters/05_components}
\input{chapters/06_experiments}

\input{chapters/07_conclusion}

\bibliography{biblio.bib}

\clearpage
\appendix
\input{chapters/09_appendix}

\end{document}

%% file: chapters/01_introduction.tex
\section{Introduction}
\label{sec:intro}

Reinforcement Learning~\citep[RL,][]{sutton2018reinforcement} has recently achieved successful results in solving several complex control problems, including autonomous driving~\citep{wang2018deep}, robot manipulators~\citep{NguyenL19}, and finance~\citep{zhang2020deep}. These outstanding achievements are rooted in the employment of powerful training algorithms combined with complex model representations, such as deep neural networks~\citep{abs-1708-05866}. Unfortunately, empirical experience suggests that this class of approaches heavily depends on fine-tuning, where an inaccurate choice of the hyper-parameters makes the difference between learning the optimal policy and not learning at all~\citep{bucsoniu2018reinforcement}. This represents an indubitable limitation, making this powerful tool not immediately usable by non-expert users. While this scenario is common even in general Machine Learning~\citep[ML,][]{bishop2006pattern}, the inherent complexity of RL, due to the sequential nature of the problem, exacerbates this issue even more.

The research effort towards the democratization of ML has reached a mature level of development for supervised learning. Indeed, several Automated Machine Learning (AutoML) frameworks and corresponding libraries have been developed and tested, such as the ones proposed by~\citet{DBLP:conf/nips/FeurerKESBH15,feurer2020auto,H2OAutoML20,OlsonGECCO2016}. AutoML is intended to automate the whole ML pipeline, starting from the preliminary operations on the data, ending with the trained and evaluated final model. This way, the complete ML process can be regarded, by the non-expert user, as a black-box, abstracting from the unnecessary details and favoring the adoption of ML as a production tool. For a detailed review of the currently available AutoML frameworks, we refer the reader to the recent survey by~\citet{he2021automl}. Conversely, RL is currently far from being a tool usable by a non-expert user since a complete and reliable Automated Reinforcement Learning (AutoRL) pipeline is currently missing. This automation gap between RL and supervised learning is even more severe from a theoretical perspective since, to the best of our knowledge, a general and flexible notion of AutoRL pipeline has not been formalized yet.

Recently, a surge of scientific works in the RL field~\citep{DBLP:journals/corr/abs-2201-03916,DBLP:journals/corr/abs-2201-05000} attempted to tackle either specific stages of the RL pipeline \emph{individually} (\eg feature construction, policy generation), or focus on specific application scenarios.
While providing a vast analysis of the available approaches for each single stage, they review the state-of-the-art to solve single tasks individually and do not propose a full pipeline and do not study the peculiarities characterizing the \emph{interaction} between such stages.
On the other hand, a na\"ive adaptation of the existing automated pipelines designed for AutoML to the RL setting is not a viable approach since they fail to capture the unique characteristics of RL related to the presence of an interacting environment and the sequential nature of the learning problem.

\textbf{Contributions}~~
In this paper, we make a step towards the formalization of an AutoRL framework. The contributions of this work can be synthesized as follows.
\begin{itemize}[noitemsep, topsep=-1pt]
    \item We propose a general and flexible formalization of a \emph{pipeline} for AutoRL. Grounding on such a definition, we instantiate it for two different scenarios: \emph{offline} and \emph{online} RL.\footnote{The reader might be tempted to address the offline RL setting with AutoML, given the fixed available dataset and, thus, the similarity with supervised learning. We stress that this choice is inappropriate as the peculiarities of RL are still crucial, especially the sequential properties of the problem.} The former assumes that the RL process is carried out based on a fixed \emph{batch} of data. The latter takes into account the availability of an interactive environment.
    \item We describe the individual \emph{\blocks} of the two pipelines and their respective characteristics, highlighting the \emph{interactions} between them and focusing on their inputs and outputs. Furthermore, we discuss the corresponding \emph{units}, \ie possible implementations of \blocks, and introduce a general approach to select the best-tuned unit in a finite set.
    \item We provide an implementation of the framework in an open-source Python library, called \libname{}.\footnote{The library is available at \url{https://github.com/arlo-lib/ARLO}.} The library contains the implementation of all the \blocks, the two RL pipelines, and the needed tools to run, optimize, and evaluate the pipelines.
    \item Finally, we test the implementation on an illustrative LQG and a MuJoCo environment, showing the ability to reach optimal performances without requiring any manual adjustment by humans. At last, we provide an experiment on a realistic dam environment with a pipeline performing the data generation, feature selection, policy generation, and policy evaluation \blocks.
\end{itemize}

Given the wide variety of RL problems and solutions, we restrict our formalization to the case of \emph{stationary} and \emph{fully observable} environments. We leave the extension to more complex settings (\eg multi-objective, multi-agent, lifelong) as a future work.

\textbf{Limitations and Broader Impact Statement}~~The goal of AutoRL is to bring RL closer to the non-expert user. This represents a source of opportunities and risks. On the one hand, making RL usable to a wide audience contributes to the \emph{democratization} of the field, overcoming the need for specific education and opening it to the large public. On the other hand, such an abstract approach tends to compromise the transparency of the learning process and traceability of the resulting model. Shadowing the underlying principles, AutoRL might pose the risk of misuse of RL approaches, leading to results not in line with expectations. Furthermore, AutoRL, even more than RL, requires huge amounts of data and computation that might represent a limit of the framework.

\textbf{Outline}~~The paper is structured as follows. In Section~\ref{sec:preliminaries}, we present the fundamental notions of Markov Decision Processes and the basics of RL. In Section~\ref{sec:framework}, we introduce a general notion of pipeline, \block, and unit. In Section~\ref{sec:pipelines}, we present the online and offline pipelines for RL. In Section~\ref{sec:components}, we describe the details of the components included in the two pipelines. In Section~\ref{sec:experiments}, we report the results of the tests performed on standard benchmarks and on a realistic environment. In Section~\ref{sec:conclusions}, we highlight the conclusions of our works and we propose future research lines.

%% file: chapters/02_preliminaries.tex
\section{Preliminaries}
\label{sec:preliminaries}

A Markov Decision Process~\citep[MDP,][]{puterman2014markov} is defined as a tuple $\mathcal{M}= (\mathcal{S}, \mathcal{A}, P, R, \gamma, \mu_0)$, where $\mathcal{S}$ is the set of states, $\mathcal{A}$ is the set of actions, $P(s'|s, a)$ is the state transition model, specifying the probability to land in state $s'$ starting from state $s$ and performing action $a$, $R(s, a)$ is the reward function, defining the expected reward when the agent is in state $s$ and performs action $a$, $\gamma \in [0,1]$ is the discount factor, and $\mu_0(s)$ is the initial state distribution.
The agent's behavior is defined in terms of a policy $\pi(a | s)$ defining the probability of performing action $a$ in state $s$.

\textbf{Interaction Protocol}~~
The initial state is sampled from the initial-state distribution $s_0 \sim \mu_0$, the agent selects an action based on its policy $a_0 \sim \pi(\cdot | s)$, the environment provides the agent with the reward $R(s_0, a_0)$, and the state evolves according to the transition model $s_1 \sim P(\cdot | s_0, a_0)$. The process is repeated for $T$ steps, where $T \in \mathbb{N} \cup\{+\infty\}$ is the (possibly infinite) horizon.

\textbf{Objective}~~
The goal of RL consists in learning an \emph{optimal} policy $\pi(a | s)$, \ie a policy maximizing the expected discounted sum of the rewards, a.k.a.~the \emph{expected return}~\citep{sutton2018reinforcement}:
\vspace{-0.2cm}
\begin{equation} \label{eq:discrew}
    J(\pi) \coloneqq \mathbb{E}^{\pi} \bigg[ \sum_{t=0}^{T-1} \gamma^{t} R(s_t,a_t)  \bigg],
\end{equation}\vspace{-0.2cm}where the expectation $\mathbb{E}^{\pi}[\cdot]$ is computed w.r.t.~the randomness of environment and policy.

\textbf{Environments and Datasets}~~
We introduce the notion of \emph{environment} and \emph{dataset}.
Formally, an environment $\mathcal{E}$ is a device to interact with the underlying MDP, that, given a state $s_t$ and an action $a_t$, it provides the next state $s'_t \sim P(\cdot|s_t,a_t)$ and the reward $r_t=R(s_t,a_t)$.
An environment is a \emph{generative model} if it allows to freely choose the state $s_t$ at each step, or a \emph{forward model} if, instead, we can perform steps in the MDP ($s_{t+1} = s'_t$) or start again sampling $s_t$ from the initial-state distribution $\mu_0$.
A dataset $\mathcal{D} \coloneqq \{\tau_i\}_{i=1}^n$ is a set of trajectories $\tau_i$, where each \emph{trajectory} is a sequence $\tau_i = (s_i^0, a_i^0, r_i^1, \ldots, s_i^{T_i-1}, a_i^{T_i-1}, r_i^{T_i}, s_i^{T_i})$ and $T_i$ is the length of the trajectory.

\textbf{Online vs.~Offline RL}~~
We distinguish between two main groups of RL algorithms: \emph{online} and \emph{offline} RL. The online RL algorithms~\citep{sutton2018reinforcement} aim at learning a policy $\pi$ by directly interacting with an environment $\mathcal{E}$. Typically they employ the last available policy to collect data and leverage the experience to improve it.
Conversely, the offline RL paradigm~\citep{levine2020offline} consists in carrying out the policy learning on a dataset $\mathcal{D}$ previously collected.\footnote{Even in this case, we may have an environment $\mathcal{E}$ to test the performance of the learned policy. Commonly, it is a less costly version, \eg in terms of computational or real costs, of the environment where the final policy will be applied.}
The ability to learn a (near-)optimal policy heavily depends on the exploration properties of the dataset $\mathcal{D}$.

Regarding offline RL, several works covered its peculiarities. In~\citet{levine2020offline} the authors survey the field of offline RL, presenting open problems, unique challenges and limitations. 
In~\citet{paine2020hyperparameter} the authors focused on the evaluation problem present in offline RL, namely on the evaluation of a learnt policy without resorting to an environment. Furthermore, this work showcases how offline RL algorithms are not robust with respect to hyper-parameters tuning.

%% file: chapters/03_framework.tex
\section{Framework}
\label{sec:framework}

In this section, we present the abstract formalization of the proposed AutoRL pipeline, detailing the notions of \emph{pipeline}, \emph{\block}, and \emph{unit}.

\textbf{Stages and Pipelines}~~
A \block $\psi$ represents a single component of the pipeline with a specific \emph{purpose}. For instance, the portion of the pipeline in charge of performing feature engineering is regarded as a stage. A stage $\psi$ interacts with the other stages of the pipeline by means of an \emph{interface}, defining its inputs and outputs.
We denote a stage's inputs with $\mathrm{In}(\psi)$ and its outputs with $\mathrm{Out}(\psi)$. A pipeline is a sequence of $m \in \mathbb{N}$ \blocks $\Psi = (\psi_1, \dots, \psi_m)$. The possibility of staking specific \blocks in sequence depends, in general, on problem-dependent constraints. 

\textbf{Units}~~
A \unit constitutes the actual \emph{implementation} of the \blocks corresponding to algorithms that are in charge of generating the output required by the corresponding \block.\footnote{From a software engineering perspective, a \block is an abstract class, while a unit a concrete class.} We define three relevant types of \units: \emph{fixed}, \emph{tunable}, and \emph{automatic}.

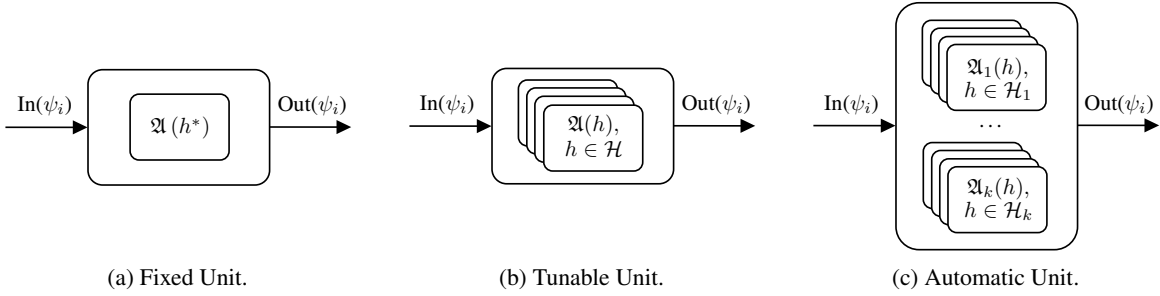
\begin{figure}[t!]
    \centering
    \begin{subfigure}[b]{0.32\textwidth}
        \centering
        \resizebox{.9\linewidth}{!}{\input{images/unit_fixed}}
        \caption{Fixed Unit.}
        \label{fig:unit_fixed}
    \end{subfigure}
    \begin{subfigure}[b]{0.32\textwidth}
        \centering
        \resizebox{.9\linewidth}{!}{\input{images/unit_tunable}}
        \caption{Tunable Unit.}
        \label{fig:unit_tunable}
    \end{subfigure}
    \begin{subfigure}[b]{0.32\textwidth}
        \centering
        \resizebox{.9\linewidth}{!}{\input{images/unit_automatic}}
        \caption{Automatic Unit.}
        \label{fig:unit_automatic}
    \end{subfigure}
    \caption{The three types of units.}
    \label{fig:units}
\end{figure}

\underline{\emph{Fixed Unit}}~~
A fixed unit (Figure~\ref{fig:unit_fixed}) corresponds to an algorithm $\psi = \mathfrak{A}(h)$, where $\mathfrak{A}(h)$ denotes algorithm $\mathfrak{A}$ that generates the \block output, instanced with hyper-parameters $h \in \mathcal{H}$ selected from an hyper-parameter set $\mathcal{H}$.

\underline{\emph{Tunable Unit}}~~
A tunable unit (Figure~\ref{fig:unit_tunable}) is described by a tuple $\psi = (\mathfrak{A}, \mathcal{H}, \mathfrak{T}, \ell)$ where $\mathfrak{A}(\cdot)$ is an algorithm, $\mathcal{H}$ is a \emph{hyper-parameters set}, $\mathfrak{T}$ is a \emph{tuner} (\eg genetic algorithm, particle swarm, Bayesian optimizer), and $\ell(\mathfrak{A},h) \in \mathbb{R}$ is a tuning \emph{performance index} mapping an algorithm $\mathfrak{A}(\cdot)$ and hyper-parameters $h \in \mathcal{H}$ pair to a real number. The \emph{tuning optimization problem} can be formulated as finding the hyper-parameters $h^*\in \mathcal{H}$ maximizing the performance index $\ell$. Formally:
\vspace{-0.2cm}
\begin{align*}
    h^* \in \argmax_{h \in \mathcal{H}} \ell(\mathfrak{A},h).
\end{align*}
This optimization is addressed by the tuner $\mathfrak{T}$. When the \block corresponding to the tunable unit is executed, it reduces to the fixed unit $\mathfrak{A}(h^*)$, and, subsequently, it generates the block outputs.

\underline{\emph{Automatic Unit}}~~
An \emph{automatic unit} (Figure~\ref{fig:unit_automatic}) is a set of tunable units paired with a performance index, \ie $\psi = (\{\psi_j\}_{j=1}^k, \ell)$, where $\psi_j = (\mathfrak{A}_j, \mathcal{H}_j, \mathfrak{T}_j, \ell_j)$, for $j \in \{1, \ldots, k\}$, and $\ell(\mathfrak{A}, h) \in \mathbb{R}$ is a performance index for algorithm $\mathfrak{A}$ with hyper-parameters $h$.
The goal of an automatic unit consists in selecting the best tuned algorithm among the available ones, by ranking them based on the additional performance index $\ell$. We define the \emph{automatic optimization problem} as follows:
\vspace{-0.2cm}
\begin{align*}
    j^* \in \argmax_{j \in \{1, \dots, k\}} \ell(\mathfrak{A}_j, h^*_j), \qquad \text{where} \qquad h^*_j \in \argmax_{h \in \mathcal{H}_j} \ell_j(\mathfrak{A}_j, h), \quad j\in\{1,\dots,k\}.
\end{align*}
When the \block corresponding to the automatic unit is executed, it reduces the automatic unit a fixed one $\mathfrak{A}_{j^*}(h^*_{j^*})$, and, subsequently, it generates the corresponding output.

The problem of jointly finding the best algorithm and its related hyper-parameter configuration is also referred in the AutoML community as CASH~\citep[Combined Algorithm Selection and Hyper-parameter Optimization Problem, ][]{thornton2013auto}.

Intuitively, a fixed unit is a human hand-crafted unit in which an algorithm is selected and the related hyper-parameters are specified. No automatic operations nor evaluation are performed here. In a tunable unit, the algorithm is specified but the task to find the best hyper-parameter configuration is demanded to the pipeline. In an automatic unit, both the choice of the best algorithm and the best hyper-parameter configuration is demanded to the pipeline.


%% file: images/unit_fixed.tex
\tikzset{every picture/.style={line width=0.75pt}} 

\begin{tikzpicture}[x=0.75pt,y=0.75pt,yscale=-1,xscale=1]

\draw  [fill={rgb, 255:red, 255; green, 255; blue, 255 }  ,fill opacity=1 ] (75.89,306.03) .. controls (75.89,303.05) and (78.31,300.63) .. (81.29,300.63) -- (130.56,300.63) .. controls (133.54,300.63) and (135.96,303.05) .. (135.96,306.03) -- (135.96,335.23) .. controls (135.96,338.21) and (133.54,340.63) .. (130.56,340.63) -- (81.29,340.63) .. controls (78.31,340.63) and (75.89,338.21) .. (75.89,335.23) -- cycle ;
\draw   (50.33,294.67) .. controls (50.33,289.76) and (54.31,285.78) .. (59.22,285.78) -- (151.67,285.78) .. controls (156.58,285.78) and (160.56,289.76) .. (160.56,294.67) -- (160.56,346.94) .. controls (160.56,351.85) and (156.58,355.83) .. (151.67,355.83) -- (59.22,355.83) .. controls (54.31,355.83) and (50.33,351.85) .. (50.33,346.94) -- cycle ;
\draw    (0.21,320.76) -- (47.04,320.81) ;
\draw [shift={(50.04,320.81)}, rotate = 180.05] [fill={rgb, 255:red, 0; green, 0; blue, 0 }  ][line width=0.08]  [draw opacity=0] (8.93,-4.29) -- (0,0) -- (8.93,4.29) -- cycle    ;
\draw    (160.28,320.59) -- (207.11,320.64) ;
\draw [shift={(210.11,320.64)}, rotate = 180.05] [fill={rgb, 255:red, 0; green, 0; blue, 0 }  ][line width=0.08]  [draw opacity=0] (8.93,-4.29) -- (0,0) -- (8.93,4.29) -- cycle    ;
\draw [color={rgb, 255:red, 255; green, 255; blue, 255 }  ,draw opacity=1 ]   (74.71,395) -- (145,392.14) ;

\draw (87,313) node [anchor=north west][inner sep=0.75pt]    {$\mathfrak{A}\left( h^{*}\right)$};
\draw (6,300) node [anchor=north west][inner sep=0.75pt]   [align=left] {In($\displaystyle \psi _{i}$)};
\draw (164,300) node [anchor=north west][inner sep=0.75pt]   [align=left] {Out($\displaystyle \psi _{i}$)};

\end{tikzpicture}

%% file: images/unit_tunable.tex
\tikzset{every picture/.style={line width=0.75pt}} 

\begin{tikzpicture}[x=0.75pt,y=0.75pt,yscale=-1,xscale=1]

\draw  [fill={rgb, 255:red, 255; green, 255; blue, 255 }  ,fill opacity=1 ] (286.36,297.77) .. controls (286.36,294.79) and (288.77,292.37) .. (291.76,292.37) -- (341.02,292.37) .. controls (344,292.37) and (346.42,294.79) .. (346.42,297.77) -- (346.42,326.97) .. controls (346.42,329.96) and (344,332.37) .. (341.02,332.37) -- (291.76,332.37) .. controls (288.77,332.37) and (286.36,329.96) .. (286.36,326.97) -- cycle ;
\draw  [fill={rgb, 255:red, 255; green, 255; blue, 255 }  ,fill opacity=1 ] (291.46,302.57) .. controls (291.46,299.59) and (293.87,297.17) .. (296.86,297.17) -- (346.12,297.17) .. controls (349.1,297.17) and (351.52,299.59) .. (351.52,302.57) -- (351.52,331.77) .. controls (351.52,334.76) and (349.1,337.17) .. (346.12,337.17) -- (296.86,337.17) .. controls (293.87,337.17) and (291.46,334.76) .. (291.46,331.77) -- cycle ;
\draw  [fill={rgb, 255:red, 255; green, 255; blue, 255 }  ,fill opacity=1 ] (296.36,307.64) .. controls (296.36,304.66) and (298.77,302.24) .. (301.76,302.24) -- (351.02,302.24) .. controls (354,302.24) and (356.42,304.66) .. (356.42,307.64) -- (356.42,336.84) .. controls (356.42,339.82) and (354,342.24) .. (351.02,342.24) -- (301.76,342.24) .. controls (298.77,342.24) and (296.36,339.82) .. (296.36,336.84) -- cycle ;
\draw  [fill={rgb, 255:red, 255; green, 255; blue, 255 }  ,fill opacity=1 ] (301.76,313.04) .. controls (301.76,310.06) and (304.17,307.64) .. (307.16,307.64) -- (356.42,307.64) .. controls (359.4,307.64) and (361.82,310.06) .. (361.82,313.04) -- (361.82,342.24) .. controls (361.82,345.22) and (359.4,347.64) .. (356.42,347.64) -- (307.16,347.64) .. controls (304.17,347.64) and (301.76,345.22) .. (301.76,342.24) -- cycle ;
\draw   (270.33,294.24) .. controls (270.33,289.33) and (274.31,285.35) .. (279.22,285.35) -- (371.67,285.35) .. controls (376.58,285.35) and (380.56,289.33) .. (380.56,294.24) -- (380.56,346.51) .. controls (380.56,351.42) and (376.58,355.4) .. (371.67,355.4) -- (279.22,355.4) .. controls (274.31,355.4) and (270.33,351.42) .. (270.33,346.51) -- cycle ;
\draw    (220.21,320.34) -- (267.04,320.38) ;
\draw [shift={(270.04,320.38)}, rotate = 180.05] [fill={rgb, 255:red, 0; green, 0; blue, 0 }  ][line width=0.08]  [draw opacity=0] (8.93,-4.29) -- (0,0) -- (8.93,4.29) -- cycle    ;
\draw    (380.28,320.16) -- (427.11,320.21) ;
\draw [shift={(430.11,320.21)}, rotate = 180.05] [fill={rgb, 255:red, 0; green, 0; blue, 0 }  ][line width=0.08]  [draw opacity=0] (8.93,-4.29) -- (0,0) -- (8.93,4.29) -- cycle    ;
\draw [color={rgb, 255:red, 255; green, 255; blue, 255 }  ,draw opacity=1 ]   (290.29,395.43) -- (360.57,395.14) ;

\draw (315,312) node [anchor=north west][inner sep=0.75pt]    {$\mathfrak{A}(h),$};
\draw (225,300) node [anchor=north west][inner sep=0.75pt]   [align=left] {In($\displaystyle \psi _{i}$)};
\draw (384,300) node [anchor=north west][inner sep=0.75pt]   [align=left] {Out($\displaystyle \psi _{i}$)};
\draw (313,328) node [anchor=north west][inner sep=0.75pt]    {$h\in \mathcal{H}$};

\end{tikzpicture}

%% file: images/unit_automatic.tex
\tikzset{every picture/.style={line width=0.75pt}} 

\begin{tikzpicture}[x=0.75pt,y=0.75pt,yscale=-1,xscale=1]

\draw  [fill={rgb, 255:red, 255; green, 255; blue, 255 }  ,fill opacity=1 ] (516.1,261.49) .. controls (516.1,258.51) and (518.52,256.09) .. (521.5,256.09) -- (570.77,256.09) .. controls (573.75,256.09) and (576.17,258.51) .. (576.17,261.49) -- (576.17,290.69) .. controls (576.17,293.67) and (573.75,296.09) .. (570.77,296.09) -- (521.5,296.09) .. controls (518.52,296.09) and (516.1,293.67) .. (516.1,290.69) -- cycle ;
\draw  [fill={rgb, 255:red, 255; green, 255; blue, 255 }  ,fill opacity=1 ] (521.2,266.29) .. controls (521.2,263.31) and (523.62,260.89) .. (526.6,260.89) -- (575.87,260.89) .. controls (578.85,260.89) and (581.27,263.31) .. (581.27,266.29) -- (581.27,295.49) .. controls (581.27,298.47) and (578.85,300.89) .. (575.87,300.89) -- (526.6,300.89) .. controls (523.62,300.89) and (521.2,298.47) .. (521.2,295.49) -- cycle ;
\draw  [fill={rgb, 255:red, 255; green, 255; blue, 255 }  ,fill opacity=1 ] (526.1,271.36) .. controls (526.1,268.37) and (528.52,265.96) .. (531.5,265.96) -- (580.77,265.96) .. controls (583.75,265.96) and (586.17,268.37) .. (586.17,271.36) -- (586.17,300.56) .. controls (586.17,303.54) and (583.75,305.96) .. (580.77,305.96) -- (531.5,305.96) .. controls (528.52,305.96) and (526.1,303.54) .. (526.1,300.56) -- cycle ;
\draw  [fill={rgb, 255:red, 255; green, 255; blue, 255 }  ,fill opacity=1 ] (531.07,276.17) .. controls (531.07,273.19) and (533.48,270.77) .. (536.47,270.77) -- (585.73,270.77) .. controls (588.72,270.77) and (591.13,273.19) .. (591.13,276.17) -- (591.13,305.37) .. controls (591.13,308.35) and (588.72,310.77) .. (585.73,310.77) -- (536.47,310.77) .. controls (533.48,310.77) and (531.07,308.35) .. (531.07,305.37) -- cycle ;
\draw   (500.08,259.42) .. controls (500.08,251.69) and (506.34,245.43) .. (514.07,245.43) -- (596.31,245.43) .. controls (604.04,245.43) and (610.3,251.69) .. (610.3,259.42) -- (610.3,381.44) .. controls (610.3,389.17) and (604.04,395.43) .. (596.31,395.43) -- (514.07,395.43) .. controls (506.34,395.43) and (500.08,389.17) .. (500.08,381.44) -- cycle ;
\draw    (449.95,320.05) -- (496.78,320.09) ;
\draw [shift={(499.78,320.1)}, rotate = 180.05] [fill={rgb, 255:red, 0; green, 0; blue, 0 }  ][line width=0.08]  [draw opacity=0] (8.93,-4.29) -- (0,0) -- (8.93,4.29) -- cycle    ;
\draw    (610.03,319.88) -- (656.86,319.92) ;
\draw [shift={(659.86,319.92)}, rotate = 180.05] [fill={rgb, 255:red, 0; green, 0; blue, 0 }  ][line width=0.08]  [draw opacity=0] (8.93,-4.29) -- (0,0) -- (8.93,4.29) -- cycle    ;
\draw  [fill={rgb, 255:red, 255; green, 255; blue, 255 }  ,fill opacity=1 ] (516.67,335.77) .. controls (516.67,332.79) and (519.09,330.37) .. (522.07,330.37) -- (571.34,330.37) .. controls (574.32,330.37) and (576.74,332.79) .. (576.74,335.77) -- (576.74,364.97) .. controls (576.74,367.96) and (574.32,370.37) .. (571.34,370.37) -- (522.07,370.37) .. controls (519.09,370.37) and (516.67,367.96) .. (516.67,364.97) -- cycle ;
\draw  [fill={rgb, 255:red, 255; green, 255; blue, 255 }  ,fill opacity=1 ] (521.77,340.57) .. controls (521.77,337.59) and (524.19,335.17) .. (527.17,335.17) -- (576.44,335.17) .. controls (579.42,335.17) and (581.84,337.59) .. (581.84,340.57) -- (581.84,369.77) .. controls (581.84,372.76) and (579.42,375.17) .. (576.44,375.17) -- (527.17,375.17) .. controls (524.19,375.17) and (521.77,372.76) .. (521.77,369.77) -- cycle ;
\draw  [fill={rgb, 255:red, 255; green, 255; blue, 255 }  ,fill opacity=1 ] (526.67,345.64) .. controls (526.67,342.66) and (529.09,340.24) .. (532.07,340.24) -- (581.34,340.24) .. controls (584.32,340.24) and (586.74,342.66) .. (586.74,345.64) -- (586.74,374.84) .. controls (586.74,377.82) and (584.32,380.24) .. (581.34,380.24) -- (532.07,380.24) .. controls (529.09,380.24) and (526.67,377.82) .. (526.67,374.84) -- cycle ;
\draw  [fill={rgb, 255:red, 255; green, 255; blue, 255 }  ,fill opacity=1 ] (531.64,350.46) .. controls (531.64,347.48) and (534.06,345.06) .. (537.04,345.06) -- (586.3,345.06) .. controls (589.29,345.06) and (591.7,347.48) .. (591.7,350.46) -- (591.7,379.66) .. controls (591.7,382.64) and (589.29,385.06) .. (586.3,385.06) -- (537.04,385.06) .. controls (534.06,385.06) and (531.64,382.64) .. (531.64,379.66) -- cycle ;

\draw (542,277) node [anchor=north west][inner sep=0.75pt]    {$\mathfrak{A}_{1}(h),$};
\draw (542,350) node [anchor=north west][inner sep=0.75pt]    {$\mathfrak{A}_{k}(h),$};
\draw (613,300) node [anchor=north west][inner sep=0.75pt]   [align=left] {Out($\displaystyle \psi _{i}$)};
\draw (455,300) node [anchor=north west][inner sep=0.75pt]   [align=left] {In($\displaystyle \psi _{i}$)};
\draw (549,318) node [anchor=north west][inner sep=0.75pt]   [align=left] {\ldots};
\draw (540,365) node [anchor=north west][inner sep=0.75pt]    {$h\in \mathcal{H}_{k}$};
\draw (540,292) node [anchor=north west][inner sep=0.75pt]    {$h\in \mathcal{H}_{1}$};

\end{tikzpicture}

%% file: chapters/04_pipeline.tex
\section{AutoRL Pipelines}
\label{sec:pipelines}

\begin{figure}[t!]
    \centering
    \begin{subfigure}[b]{\textwidth}
        \centering
        \resizebox{\linewidth}{!}{\input{images/pipeline_online}}
        \caption{Online Pipeline.}
        \label{fig:pipeline_online}
    \end{subfigure}
    \begin{subfigure}[b]{\textwidth}
        \centering
        \resizebox{\linewidth}{!}{\input{images/pipeline_offline}}
        \caption{Offline Pipeline.}
        \label{fig:pipeline_offline}
    \end{subfigure}
    \caption{The Online (a) and Offline (b) AutoRL Pipelines.}
    \label{fig:pipelines}
\end{figure}
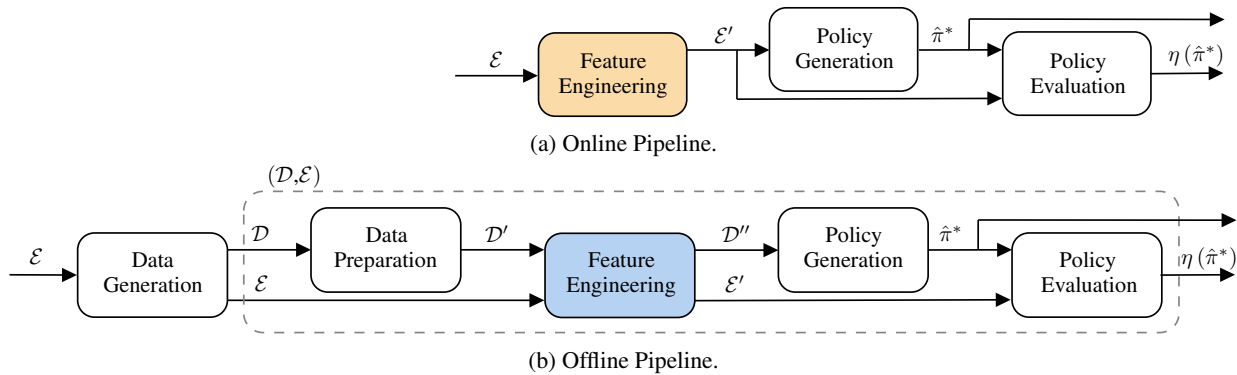

In this section, we present the main methodological contribution of the paper, discussing the two AutoRL pipelines: \emph{online} and \emph{offline}. We focus on how to build these pipelines describing the stages' interactions. The detailed description of each individual stage is reported in Section~\ref{sec:components}. A graphical representation of the pipelines is provided in Figure~\ref{fig:pipelines}.

\textbf{Online Pipeline}~~
The \emph{Online AutoRL Pipeline} (Figure~\ref{fig:pipeline_online}) takes as input an environment $\mathcal{E}$ that is fed to the \texttt{Feature Engineering} stage, which modifies its state/action representations and the reward to facilitate the learning performed in the next stages. It outputs a transformed environment $\mathcal{E}'$, based on the features created in this stage. Subsequently, the environment $\mathcal{E}'$ is used to learn an estimate $\hat{\pi}^*$ of the optimal policy through the \texttt{Policy Generation}. Finally, the \texttt{Policy Evaluation} phase provides an estimate of the performance $\eta(\hat{\pi}^*)$, based on a performance index $\eta$.

\textbf{Offline Pipeline}~~
In the \emph{Offline AutoRL Pipeline} (Figure~\ref{fig:pipeline_offline}), differently from the online one, two additional preliminary stages are included: \texttt{Data Generation} and \texttt{Data Preparation}.
If an environment $\mathcal{E}$ is provided as input, the \texttt{Data Generation} stage creates a dataset $\mathcal{D}$. This stage is omitted if a dataset $\mathcal{D}$ is already available, \eg in the case the dataset $\mathcal{D}$ comes from a real process. In such a case, the environment $\mathcal{E}$ is employed for the evaluation of the policy performance only.
The \texttt{Data Preparation} stage modifies the dataset $\mathcal{D}$, by applying corrections over the individual instances (\ie the rows of the dataset) obtaining $\mathcal{D}'$. Then, the environment $\mathcal{E}$ and dataset $\mathcal{D}'$ pass through the \texttt{Feature Engineering} stage, which, similarly to its online counterpart, generates a dataset $\mathcal{D}''$ and an environment $\mathcal{E}'$ with transformed states, actions, and reward. After that, the dataset $\mathcal{D}''$ is used for learning an estimate of the optimal policy $\hat{\pi}^*$ through the \texttt{Policy Generation} stage. Differently from the online one, this stage uses the dataset $\mathcal{D}''$, while the environment $\mathcal{E}'$ is employed for estimating $\eta(\hat{\pi}^*)$ in the \texttt{Policy Evaluation} stage.

%% file: images/pipeline_online.tex
\tikzset{every picture/.style={line width=0.75pt}} 

\begin{tikzpicture}[x=0.75pt,y=0.75pt,yscale=-1,xscale=1]

\draw  [fill={rgb, 255:red, 245; green, 166; blue, 35 }  ,fill opacity=0.4 ] (310.2,410.34) .. controls (310.2,404.77) and (314.71,400.26) .. (320.28,400.26) -- (389.72,400.26) .. controls (395.28,400.26) and (399.8,404.77) .. (399.8,410.34) -- (399.8,440.58) .. controls (399.8,446.15) and (395.28,450.66) .. (389.72,450.66) -- (320.28,450.66) .. controls (314.71,450.66) and (310.2,446.15) .. (310.2,440.58) -- cycle ;
\draw    (680.45,424.92) -- (721.23,424.56) ;
\draw [shift={(724.23,424.54)}, rotate = 179.5] [fill={rgb, 255:red, 0; green, 0; blue, 0 }  ][line width=0.08]  [draw opacity=0] (8.93,-4.29) -- (0,0) -- (8.93,4.29) -- cycle    ;
\draw    (259.8,426.14) -- (306.8,426.14) ;
\draw [shift={(309.8,426.14)}, rotate = 180] [fill={rgb, 255:red, 0; green, 0; blue, 0 }  ][line width=0.08]  [draw opacity=0] (8.93,-4.29) -- (0,0) -- (8.93,4.29) -- cycle    ;
\draw   (449.8,395.14) .. controls (449.8,389.57) and (454.31,385.06) .. (459.88,385.06) -- (529.32,385.06) .. controls (534.88,385.06) and (539.4,389.57) .. (539.4,395.14) -- (539.4,425.38) .. controls (539.4,430.95) and (534.88,435.46) .. (529.32,435.46) -- (459.88,435.46) .. controls (454.31,435.46) and (449.8,430.95) .. (449.8,425.38) -- cycle ;
\draw    (399.8,410.59) -- (446.8,410.59) ;
\draw [shift={(449.8,410.59)}, rotate = 180] [fill={rgb, 255:red, 0; green, 0; blue, 0 }  ][line width=0.08]  [draw opacity=0] (8.93,-4.29) -- (0,0) -- (8.93,4.29) -- cycle    ;
\draw   (590.8,409.74) .. controls (590.8,404.17) and (595.31,399.66) .. (600.88,399.66) -- (670.32,399.66) .. controls (675.88,399.66) and (680.4,404.17) .. (680.4,409.74) -- (680.4,439.98) .. controls (680.4,445.55) and (675.88,450.06) .. (670.32,450.06) -- (600.88,450.06) .. controls (595.31,450.06) and (590.8,445.55) .. (590.8,439.98) -- cycle ;
\draw [color={rgb, 255:red, 255; green, 255; blue, 255 }  ,draw opacity=1 ]   (-9.73,423.03) -- (27.62,423.17) ;
\draw [shift={(30.62,423.19)}, rotate = 180.22] [fill={rgb, 255:red, 255; green, 255; blue, 255 }  ,fill opacity=1 ][line width=0.08]  [draw opacity=0] (8.93,-4.29) -- (0,0) -- (8.93,4.29) -- cycle    ;
\draw    (430.13,440.33) -- (587,439.87) -- (587.44,439.87) ;
\draw [shift={(590.44,439.87)}, rotate = 180] [fill={rgb, 255:red, 0; green, 0; blue, 0 }  ][line width=0.08]  [draw opacity=0] (8.93,-4.29) -- (0,0) -- (8.93,4.29) -- cycle    ;
\draw    (540.15,410.59) -- (587.45,410.49) ;
\draw [shift={(590.45,410.49)}, rotate = 179.88] [fill={rgb, 255:red, 0; green, 0; blue, 0 }  ][line width=0.08]  [draw opacity=0] (8.93,-4.29) -- (0,0) -- (8.93,4.29) -- cycle    ;
\draw    (570.56,391.84) -- (721.94,392.12) ;
\draw [shift={(724.94,392.12)}, rotate = 180.1] [fill={rgb, 255:red, 0; green, 0; blue, 0 }  ][line width=0.08]  [draw opacity=0] (8.93,-4.29) -- (0,0) -- (8.93,4.29) -- cycle    ;
\draw    (570.56,391.84) -- (570.43,410.71) ;
\draw    (430.17,410.67) -- (430.13,440.33) ;
\draw [color={rgb, 255:red, 255; green, 255; blue, 255 }  ,draw opacity=1 ]   (691.33,440.33) -- (733.33,439.67) ;

\draw (279,410) node [anchor=north west][inner sep=0.75pt]    {$\mathcal{E}$};
\draw (321,426) node [anchor=north west][inner sep=0.75pt]   [align=left] {Engineering};
\draw (334,411) node [anchor=north west][inner sep=0.75pt]   [align=left] {Feature};
\draw (415,395) node [anchor=north west][inner sep=0.75pt]    {$\mathcal{E} '$};
\draw (477,396) node [anchor=north west][inner sep=0.75pt]   [align=left] {Policy};
\draw (618,411) node [anchor=north west][inner sep=0.75pt]   [align=left] {Policy};
\draw (606,426) node [anchor=north west][inner sep=0.75pt]   [align=left] {Evaluation};
\draw (464,411) node [anchor=north west][inner sep=0.75pt]   [align=left] {Generation};
\draw (546,395) node [anchor=north west][inner sep=0.75pt]    {$\hat{\pi }^{*}$};
\draw (689.36,406) node [anchor=north west][inner sep=0.75pt]    {$\eta\left(\hat{\pi }^{*}\right)$};

\end{tikzpicture}

%% file: images/pipeline_offline.tex
\tikzset{every picture/.style={line width=0.75pt}} 

\begin{tikzpicture}[x=0.75pt,y=0.75pt,yscale=-1,xscale=1]

\draw    (-9.63,315.03) -- (27.71,315.17) ;
\draw [shift={(30.71,315.19)}, rotate = 180.22] [fill={rgb, 255:red, 0; green, 0; blue, 0 }  ][line width=0.08]  [draw opacity=0] (8.93,-4.29) -- (0,0) -- (8.93,4.29) -- cycle    ;
\draw  [fill={rgb, 255:red, 74; green, 144; blue, 226 }  ,fill opacity=0.4 ] (310.74,300.28) .. controls (310.74,294.71) and (315.25,290.2) .. (320.82,290.2) -- (390.26,290.2) .. controls (395.83,290.2) and (400.34,294.71) .. (400.34,300.28) -- (400.34,330.52) .. controls (400.34,336.09) and (395.83,340.6) .. (390.26,340.6) -- (320.82,340.6) .. controls (315.25,340.6) and (310.74,336.09) .. (310.74,330.52) -- cycle ;
\draw    (678.91,315.21) -- (720.54,315.01) ;
\draw [shift={(723.54,315)}, rotate = 179.72] [fill={rgb, 255:red, 0; green, 0; blue, 0 }  ][line width=0.08]  [draw opacity=0] (8.93,-4.29) -- (0,0) -- (8.93,4.29) -- cycle    ;
\draw   (170.74,285.08) .. controls (170.74,279.51) and (175.25,275) .. (180.82,275) -- (250.26,275) .. controls (255.83,275) and (260.34,279.51) .. (260.34,285.08) -- (260.34,315.32) .. controls (260.34,320.89) and (255.83,325.4) .. (250.26,325.4) -- (180.82,325.4) .. controls (175.25,325.4) and (170.74,320.89) .. (170.74,315.32) -- cycle ;
\draw   (31.14,300.08) .. controls (31.14,294.51) and (35.65,290) .. (41.22,290) -- (110.66,290) .. controls (116.23,290) and (120.74,294.51) .. (120.74,300.08) -- (120.74,330.32) .. controls (120.74,335.89) and (116.23,340.4) .. (110.66,340.4) -- (41.22,340.4) .. controls (35.65,340.4) and (31.14,335.89) .. (31.14,330.32) -- cycle ;
\draw    (120.74,300.08) -- (167.74,300.08) ;
\draw [shift={(170.74,300.08)}, rotate = 180] [fill={rgb, 255:red, 0; green, 0; blue, 0 }  ][line width=0.08]  [draw opacity=0] (8.93,-4.29) -- (0,0) -- (8.93,4.29) -- cycle    ;
\draw    (260.34,300.08) -- (307.34,300.08) ;
\draw [shift={(310.34,300.08)}, rotate = 180] [fill={rgb, 255:red, 0; green, 0; blue, 0 }  ][line width=0.08]  [draw opacity=0] (8.93,-4.29) -- (0,0) -- (8.93,4.29) -- cycle    ;
\draw    (120.74,330.32) -- (307.34,330.32) ;
\draw [shift={(310.34,330.32)}, rotate = 180] [fill={rgb, 255:red, 0; green, 0; blue, 0 }  ][line width=0.08]  [draw opacity=0] (8.93,-4.29) -- (0,0) -- (8.93,4.29) -- cycle    ;
\draw   (450.34,285.08) .. controls (450.34,279.51) and (454.85,275) .. (460.42,275) -- (529.86,275) .. controls (535.43,275) and (539.94,279.51) .. (539.94,285.08) -- (539.94,315.32) .. controls (539.94,320.89) and (535.43,325.4) .. (529.86,325.4) -- (460.42,325.4) .. controls (454.85,325.4) and (450.34,320.89) .. (450.34,315.32) -- cycle ;
\draw    (400.34,300.28) -- (447.34,300.28) ;
\draw [shift={(450.34,300.28)}, rotate = 180] [fill={rgb, 255:red, 0; green, 0; blue, 0 }  ][line width=0.08]  [draw opacity=0] (8.93,-4.29) -- (0,0) -- (8.93,4.29) -- cycle    ;
\draw    (400.34,330.52) -- (586.33,330.01) ;
\draw [shift={(589.33,330)}, rotate = 179.84] [fill={rgb, 255:red, 0; green, 0; blue, 0 }  ][line width=0.08]  [draw opacity=0] (8.93,-4.29) -- (0,0) -- (8.93,4.29) -- cycle    ;
\draw   (590.14,300.08) .. controls (590.14,294.51) and (594.65,290) .. (600.22,290) -- (669.66,290) .. controls (675.23,290) and (679.74,294.51) .. (679.74,300.08) -- (679.74,330.32) .. controls (679.74,335.89) and (675.23,340.4) .. (669.66,340.4) -- (600.22,340.4) .. controls (594.65,340.4) and (590.14,335.89) .. (590.14,330.32) -- cycle ;
\draw  [color={rgb, 255:red, 132; green, 132; blue, 132 }  ,draw opacity=1 ][dash pattern={on 4.5pt off 4.5pt}] (130.51,276.58) .. controls (130.51,270.23) and (135.66,265.09) .. (142,265.09) -- (678.51,265.09) .. controls (684.86,265.09) and (690,270.23) .. (690,276.58) -- (690,338.14) .. controls (690,344.48) and (684.86,349.63) .. (678.51,349.63) -- (142,349.63) .. controls (135.66,349.63) and (130.51,344.48) .. (130.51,338.14) -- cycle ;
\draw    (539.85,300.18) -- (587.14,300.09) ;
\draw [shift={(590.14,300.08)}, rotate = 179.88] [fill={rgb, 255:red, 0; green, 0; blue, 0 }  ][line width=0.08]  [draw opacity=0] (8.93,-4.29) -- (0,0) -- (8.93,4.29) -- cycle    ;
\draw    (569.89,282.22) -- (721.44,282.34) ;
\draw [shift={(724.44,282.34)}, rotate = 180.04] [fill={rgb, 255:red, 0; green, 0; blue, 0 }  ][line width=0.08]  [draw opacity=0] (8.93,-4.29) -- (0,0) -- (8.93,4.29) -- cycle    ;
\draw    (569.89,282.22) -- (569.86,299.57) ;

\draw (0.87,298) node [anchor=north west][inner sep=0.75pt]    {$\mathcal{E}$};
\draw (45,315) node [anchor=north west][inner sep=0.75pt]   [align=left] {Generation};
\draw (62,300) node [anchor=north west][inner sep=0.75pt]   [align=left] {Data};
\draw (133.66,284) node [anchor=north west][inner sep=0.75pt]    {$\mathcal{D}$};
\draw (183,300) node [anchor=north west][inner sep=0.75pt]   [align=left] {Preparation};
\draw (202,285) node [anchor=north west][inner sep=0.75pt]   [align=left] {Data};
\draw (135.68,314) node [anchor=north west][inner sep=0.75pt]    {$\mathcal{E}$};
\draw (273.43,284) node [anchor=north west][inner sep=0.75pt]    {$\mathcal{D} '$};
\draw (322,315) node [anchor=north west][inner sep=0.75pt]   [align=left] {Engineering};
\draw (335,300) node [anchor=north west][inner sep=0.75pt]   [align=left] {Feature};
\draw (417,314) node [anchor=north west][inner sep=0.75pt]    {$\mathcal{E} '$};
\draw (415,284) node [anchor=north west][inner sep=0.75pt]    {$\mathcal{D} ''$};
\draw (477,285) node [anchor=north west][inner sep=0.75pt]   [align=left] {Policy};
\draw (617,300) node [anchor=north west][inner sep=0.75pt]   [align=left] {Policy};
\draw (545,284) node [anchor=north west][inner sep=0.75pt]    {$\hat{\pi }^{*}$};
\draw (606,315) node [anchor=north west][inner sep=0.75pt]   [align=left] {Evaluation};
\draw (142.94,248) node [anchor=north west][inner sep=0.75pt]   [align=left] {$\displaystyle (\mathcal{D}$,$\displaystyle \mathcal{E})$};
\draw (464,300) node [anchor=north west][inner sep=0.75pt]   [align=left] {Generation};
\draw (689.79,296.9) node [anchor=north west][inner sep=0.75pt]    {$\eta\left(\hat{\pi }^{*}\right)$};

\end{tikzpicture}

%% file: chapters/05_components.tex
\section{Stages and Units}
\label{sec:components}

We now provide examples of units for each of the \blocks, highlighting the differences between the online and offline pipelines.
For each \block, we define its goal, performance index for tunable or automatic units, and implementation selected from the state-of-the-art methodologies.

\subsection{Data Generation}
The \texttt{Data Generation} stage takes as input an environment $\mathcal{E}$ and returns the unaltered environment $\mathcal{E}$ and a dataset $\mathcal{D}$ generated by interacting with the environment. 
The goal of this stage is to create a dataset that is retrieved by exploring the state space as much as possible.
Based on the type of environment, \ie generative or forward model, the resulting dataset is made of transitions or trajectories.

In principle, this \block should output a dataset as \quotes{informative} as possible, \ie that represents exhaustively the corresponding environment. As performance index for evaluating the quality of a \texttt{Data Generation} unit, we adopt the \emph{entropy} of the state-action visitation distribution $d_\pi(s,a)$ generated by the policy $\pi(a|s)$, that is proportional to: 
$$-\int_{s\in \mathcal{S}} \int_{a\in \mathcal{A}} {d_\pi(s,a) \log d_\pi(s,a)} da \ ds.\footnote{In this stage, we rely on the Particle Based Entropy estimation developed by~\citet{singh2003nearest}.}
$$
A straightforward implementation of \texttt{Data Generation} consists in collecting data with the random uniform policy. However, this approach is not guaranteed to explore the state space effectively~\citep{DBLP:conf/aaai/MuttiPR21,DBLP:conf/icra/EndrawisLJNT21}.
In the pipeline, we consider the state-of-the-art solutions proposed by~\cite{pathak2019self}, and~\cite{DBLP:conf/aaai/MuttiPR21}.
The former employs Proximal Policy Optimization~\citep[PPO,][]{schulman2017proximal} using the estimated variance of the MDP dynamics as reward, as a proxy for the entropy.
Instead, the latter provides a novel policy search algorithm maximizing a $K$-nearest neighbours-based estimate of the state distribution entropy.

\subsection{Data Preparation}

This phase uses a dataset $\mathcal{D}$, coming either from a real-world environment or generated in the \texttt{Data Generation} stage, and returns a dataset $\mathcal{D}'$ with the same state-action features and reward, but with a possibly different number of entries. 
The goal of this phase is to optimize an existing dataset in order to be processed better from in subsequent stages.
\texttt{Data Preparation} includes data augmentation, data imputation, and data scaling, and can embed further domain-specific sub-stages (\eg for images, audio data), and/or consistency checks (\eg filling missing values).

No single automatic unit is deemed adoptable due to the difficulty of defining a general enough performance index for this stage. However, domain-specific performance indexes are available, \eg for the data imputation sub-stages, we may rely on the indexes defined by~\citet{jadhav2019comparison}.

Possible implementations of this stage include the techniques for classical ML preprocessing, such as imputation from a dataset of trajectories via KNN imputation or Bayesian Multiple Imputation~\citep{lizotte2008missing}. 
Moreover, for pixel-based observations (\eg the Gym Atari environments) data augmentation techniques, \eg cropping, reflection, scaling, were employed in~\cite{DBLP:conf/cig/YeKBT20}. Other approaches viable for feature-based representations are presented in~\cite{DBLP:conf/nips/LaskinLSPAS20}, where experiments on the OpenAI Procgen Benchmark and on the MuJoCo environments.

\begin{figure}[t!]
     \centering
     \begin{subfigure}[b]{0.85\textwidth}
         \centering
         \resizebox{\linewidth}{!}{\input{images/fs_online}}
         \caption{Feature Engineering \block in Online Pipeline.}
         \label{fig:fs_online}
     \end{subfigure}
     \begin{subfigure}[b]{0.67\textwidth}
         \centering
         \resizebox{\linewidth}{!}{\input{images/fs_offline}}
         \caption{Feature Engineering \block in Offline Pipeline.}
         \label{fig:fs_offline}
     \end{subfigure}
     \caption{The offline and online \texttt{Feature Engineering} \blocks.} \label{fig:fs}
\end{figure}
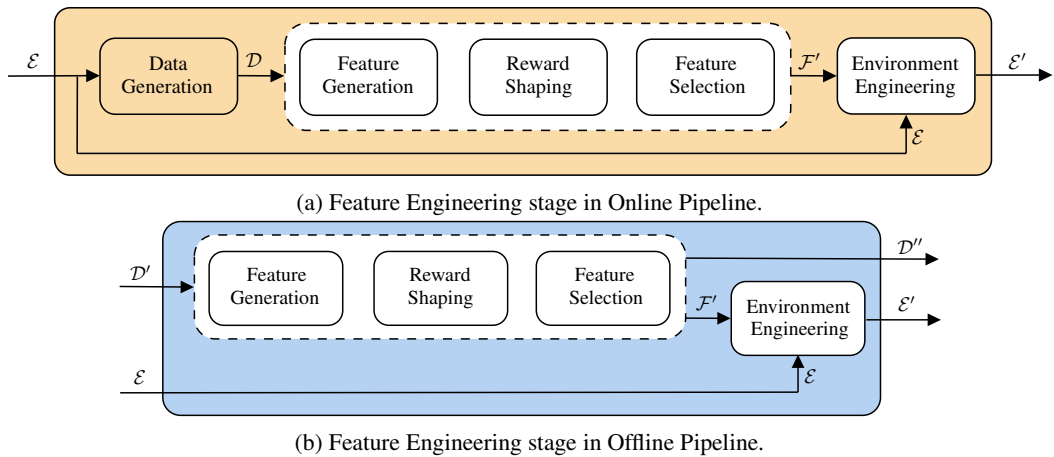

\subsection{Feature Engineering}

The \texttt{Feature Engineering} stage displays significant differences between online and offline pipelines (Figure~\ref{fig:fs}). Offline pipelines (Figure~\ref{fig:fs_offline}) take as input an environment $\mathcal{E}$ and a dataset $\mathcal{D}'$ and return a feature-adjusted environment $\mathcal{E}'$ and dataset $\mathcal{D}''$. Conversely, online pipelines (Figure~\ref{fig:fs_online}) take as input an environment $\mathcal{E}$ and return a feature-adjusted environment $\mathcal{E}'$.
In both cases, this stage requires an internal dataset for feature engineering that, for the online case, has to be generated.

The core task of this stage is to select and generate a set of features that properly model the state-action space of the problem and perform reward shaping actions to facilitate the following learning phase. \texttt{Feature Engineering} \block includes one or more of the following \emph{sub-stages}:
\begin{itemize}[noitemsep, topsep=-1pt]
    \item \texttt{Feature Generation}, in charge of creating new features. This sub-stage makes use of techniques such as radial basis functions, tile coding, or coarse coding~\citep{sutton2018reinforcement}.
    \item \texttt{Feature Selection}, aimed at selecting a meaningful subset of features, either to reduce the computation requirements or to regularize the following policy learning phase. Viable options are Mutual Information-based selection~\citep{DBLP:conf/ijcnn/BerahaMPTR19}, correlation-based filtering methods, and tree-based variable selection~\citep{DBLP:conf/adprl/CastellettiGRS11}.
    \item \texttt{Reward Shaping}, performing specific transformations on the reward function, possibly preserving the optimal policy, to speed up the convergence of an RL algorithm~\citep{ng1999policy}. For instance, in presence of sparse reward functions, reward shaping can be regarded as a form of \emph{curriculum learning}~\citep{portelas2020automatic}.
\end{itemize}
These sub-stages return a transformation that is applied to the environment through the \texttt{Environment Engineering} \block. In the offline case, the same transformation is applied to the dataset, while for the online case the internal dataset is disregarded.

We consider as a performance index for the complete feature engineering stage the mutual information between the current state-action pair $(s, a)$ features and the next-state reward $(s',r)$ features~\citep{mutualinfo,NIPS2017_ef72d539} regularized, \eg by the number of selected features.\footnote{For instance, one may use the ratio between the mutual information and the number of selected features.}

\subsection{Policy Generation}

The \texttt{Policy Generation} stage is in charge of the training phase of the RL learning algorithm. More specifically, it takes as input an environment $\mathcal{E}'$ or a dataset $\mathcal{D}''$, in the online and offline RL pipelines, respectively, to output an estimate $\hat{\pi}^*$ of the optimal policy.

Among the most common choices of performance indexes for this stage, we mention the \emph{expected return}, \ie the expected discounted sum of the rewards, the \emph{average reward}, \ie the long-term expected average reward, and the \emph{total reward} \ie expected cumulative sum of the rewards~\citep[in the case the environment is episodic, ][]{puterman2014markov}. For specific applications, \eg risk-averse setting, one may adopt the mean-variance, mean-volatility, and CVaR~\citep{pratt1978risk, bisi2021risk}.

Many works deal with hyper-parameter optimization for RL algorithms. In~\cite{DBLP:conf/iclr/FrankeKBH21} a framework based on Population Based Training \citep[PBT, ][]{jaderberg2017population} is proposed to tune off-policy RL algorithms. In~\cite{DBLP:journals/corr/abs-2106-15883} a new time-varying bandit algorithm was presented for tuning RL algorithms. Hyper-parameter tuning is a widely researched topic and the techniques developed by ML algorithms can be used for RL algorithms as well. Nevertheless, the sample inefficiency of tuning techniques is a common problem, not unique to RL. 
Another issue is the sensitivity to hyper-parameters configurations, which increases the difficulty of benchmarking tuning algorithms due to the difficulty of obtaining reproducible results.
Further methods were proposed by~\cite{DBLP:conf/aistats/ZhangRPLBCHC21,DBLP:conf/icml/LeeLSA21,DBLP:journals/corr/abs-2107-12808,DBLP:conf/atal/SaphalRMAK21,DBLP:conf/icml/FalknerKH18}.

The specific implementation of the \texttt{Policy Generation} stage depends on the selected RL algorithm. For offline pipelines, we mention, among the others, Least Squares Policy Iteration~\cite[LSPI,][]{lagoudakis2003least}, Fitted Q-Iteration~\citep[FQI,][]{DBLP:journals/jmlr/ErnstGW05}. For online pipelines, a large surge of RL algorithms have been developed in the recent years. We mention, among the most popular ones, Deep Q-Networks~\citep[DQN,][]{DBLP:journals/corr/SchaulQAS15}, Deep Deterministic Policy Gradient~\citep[DDPG,][]{DBLP:journals/corr/LillicrapHPHETS15}, Trust Region Policy Optimization~\citep[TRPO,][]{DBLP:conf/icml/SchulmanLAJM15}, Soft Actor Critic~\citep[SAC,][]{DBLP:conf/icml/HaarnojaZAL18}, and Proximal Policy Optimization~\citep[PPO,][]{schulman2017proximal}.

\subsection{Policy Evaluation}

The \texttt{Policy Evaluation} \block takes as input the policy $\hat{\pi}^*$ produced by the \texttt{Policy Generation} phase and an environment $\mathcal{E}'$, and produces as output an estimation of a performance index $\eta(\hat{\pi}^*)$. 

Regarding the performance index used in this \block, the options are the same we mentioned for \texttt{Policy Generation}. Notice that the performance index chosen in this \block may differ from the one of the \texttt{Policy Generation} one. For instance, it is a common practice to train RL algorithms using a discounted objective and evaluate the resulting policies using an undiscounted one~\citep{duan2016benchmarking}.
Notice that, due to the nature of the task, only fixed units are used in this \block.

%% file: images/fs_online.tex
\tikzset{every picture/.style={line width=0.75pt}} 

\begin{tikzpicture}[x=0.75pt,y=0.75pt,yscale=-1,xscale=1]

\draw  [fill={rgb, 255:red, 245; green, 166; blue, 35 }  ,fill opacity=0.4 ] (19.98,39.81) .. controls (19.98,34.28) and (24.46,29.8) .. (29.99,29.8) -- (620.36,29.8) .. controls (625.89,29.8) and (630.38,34.28) .. (630.38,39.81) -- (630.38,128.91) .. controls (630.38,134.44) and (625.89,138.93) .. (620.36,138.93) -- (29.99,138.93) .. controls (24.46,138.93) and (19.98,134.44) .. (19.98,128.91) -- cycle ;
\draw    (-10.42,74.33) -- (40.17,74.08) -- (46.47,74.08) ;
\draw [shift={(49.47,74.08)}, rotate = 180] [fill={rgb, 255:red, 0; green, 0; blue, 0 }  ][line width=0.08]  [draw opacity=0] (8.93,-4.29) -- (0,0) -- (8.93,4.29) -- cycle    ;
\draw    (620.18,73.66) -- (667.18,73.66) ;
\draw [shift={(670.18,73.66)}, rotate = 180] [fill={rgb, 255:red, 0; green, 0; blue, 0 }  ][line width=0.08]  [draw opacity=0] (8.93,-4.29) -- (0,0) -- (8.93,4.29) -- cycle    ;
\draw  [color={rgb, 255:red, 0; green, 0; blue, 0 }  ,draw opacity=1 ][fill={rgb, 255:red, 255; green, 255; blue, 255 }  ,fill opacity=1 ][dash pattern={on 4.5pt off 4.5pt}] (170.07,53.93) .. controls (170.07,46.22) and (176.32,39.98) .. (184.02,39.98) -- (485.63,39.98) .. controls (493.33,39.98) and (499.58,46.22) .. (499.58,53.93) -- (499.58,95.78) .. controls (499.58,103.48) and (493.33,109.73) .. (485.63,109.73) -- (184.02,109.73) .. controls (176.32,109.73) and (170.07,103.48) .. (170.07,95.78) -- cycle ;
\draw  [fill={rgb, 255:red, 255; green, 255; blue, 255 }  ,fill opacity=1 ] (529.98,59.4) .. controls (529.98,53.93) and (534.42,49.49) .. (539.9,49.49) -- (609.67,49.49) .. controls (615.14,49.49) and (619.58,53.93) .. (619.58,59.4) -- (619.58,89.14) .. controls (619.58,94.62) and (615.14,99.05) .. (609.67,99.05) -- (539.9,99.05) .. controls (534.42,99.05) and (529.98,94.62) .. (529.98,89.14) -- cycle ;
\draw   (290.38,60.2) .. controls (290.38,54.73) and (294.82,50.29) .. (300.3,50.29) -- (370.07,50.29) .. controls (375.54,50.29) and (379.98,54.73) .. (379.98,60.2) -- (379.98,89.94) .. controls (379.98,95.42) and (375.54,99.85) .. (370.07,99.85) -- (300.3,99.85) .. controls (294.82,99.85) and (290.38,95.42) .. (290.38,89.94) -- cycle ;
\draw   (399.23,60.2) .. controls (399.23,54.73) and (403.67,50.29) .. (409.14,50.29) -- (478.92,50.29) .. controls (484.39,50.29) and (488.83,54.73) .. (488.83,60.2) -- (488.83,89.94) .. controls (488.83,95.42) and (484.39,99.85) .. (478.92,99.85) -- (409.14,99.85) .. controls (403.67,99.85) and (399.23,95.42) .. (399.23,89.94) -- cycle ;
\draw    (139.37,74.06) -- (166.58,74.21) ;
\draw [shift={(169.58,74.22)}, rotate = 180.32] [fill={rgb, 255:red, 0; green, 0; blue, 0 }  ][line width=0.08]  [draw opacity=0] (8.93,-4.29) -- (0,0) -- (8.93,4.29) -- cycle    ;
\draw    (499.38,74.06) -- (526.58,74.21) ;
\draw [shift={(529.58,74.22)}, rotate = 180.32] [fill={rgb, 255:red, 0; green, 0; blue, 0 }  ][line width=0.08]  [draw opacity=0] (8.93,-4.29) -- (0,0) -- (8.93,4.29) -- cycle    ;
\draw    (34.64,124.76) -- (574.78,124.76) ;
\draw    (574.78,124.76) -- (574.78,102.56) ;
\draw [shift={(574.78,99.56)}, rotate = 90] [fill={rgb, 255:red, 0; green, 0; blue, 0 }  ][line width=0.08]  [draw opacity=0] (8.93,-4.29) -- (0,0) -- (8.93,4.29) -- cycle    ;
\draw    (34.78,74.63) -- (34.64,124.76) ;
\draw  [color={rgb, 255:red, 0; green, 0; blue, 0 }  ,draw opacity=1 ] (49.54,59.58) .. controls (49.54,54.11) and (53.98,49.67) .. (59.45,49.67) -- (129.23,49.67) .. controls (134.7,49.67) and (139.14,54.11) .. (139.14,59.58) -- (139.14,89.32) .. controls (139.14,94.79) and (134.7,99.23) .. (129.23,99.23) -- (59.45,99.23) .. controls (53.98,99.23) and (49.54,94.79) .. (49.54,89.32) -- cycle ;
\draw   (179.76,60.25) .. controls (179.76,54.77) and (184.2,50.34) .. (189.67,50.34) -- (259.45,50.34) .. controls (264.92,50.34) and (269.36,54.77) .. (269.36,60.25) -- (269.36,89.99) .. controls (269.36,95.46) and (264.92,99.9) .. (259.45,99.9) -- (189.67,99.9) .. controls (184.2,99.9) and (179.76,95.46) .. (179.76,89.99) -- cycle ;

\draw (0.57,58.17) node [anchor=north west][inner sep=0.75pt]    {$\mathcal{E}$};
\draw (639.93,58.24) node [anchor=north west][inner sep=0.75pt]    {$\mathcal{E} '$};
\draw (203,60) node [anchor=north west][inner sep=0.75pt]   [align=left] {Feature};
\draw (193,75) node [anchor=north west][inner sep=0.75pt]   [align=left] {Generation};
\draw (313,60) node [anchor=north west][inner sep=0.75pt]   [align=left] {Reward};
\draw (312,75) node [anchor=north west][inner sep=0.75pt]   [align=left] {Shaping};
\draw (423,60) node [anchor=north west][inner sep=0.75pt]   [align=left] {Feature};
\draw (419,75) node [anchor=north west][inner sep=0.75pt]   [align=left] {Selection};
\draw (143,58) node [anchor=north west][inner sep=0.75pt]    {$\mathcal{D}$};
\draw (503,58) node [anchor=north west][inner sep=0.75pt]    {$\mathcal{F} '$};
\draw (80,60) node [anchor=north west][inner sep=0.75pt]   [align=left] {Data};
\draw (62,75) node [anchor=north west][inner sep=0.75pt]   [align=left] {Generation};
\draw (538,60) node [anchor=north west][inner sep=0.75pt]   [align=left] {Environment};
\draw (540,75) node [anchor=north west][inner sep=0.75pt]   [align=left] {Engineering};
\draw (577,108) node [anchor=north west][inner sep=0.75pt]    {$\mathcal{E}$};

\end{tikzpicture}

%% file: images/fs_offline.tex
\tikzset{every picture/.style={line width=0.75pt}} 

\begin{tikzpicture}[x=0.75pt,y=0.75pt,yscale=-1,xscale=1]

\draw  [fill={rgb, 255:red, 74; green, 144; blue, 226 }  ,fill opacity=0.4 ] (71.05,192.26) .. controls (71.05,185.66) and (76.4,180.32) .. (82.99,180.32) -- (540.5,180.32) .. controls (547.1,180.32) and (552.44,185.66) .. (552.44,192.26) -- (552.44,298.49) .. controls (552.44,305.08) and (547.1,310.43) .. (540.5,310.43) -- (82.99,310.43) .. controls (76.4,310.43) and (71.05,305.08) .. (71.05,298.49) -- cycle ;
\draw    (543,246.29) -- (590,246.29) ;
\draw [shift={(593,246.29)}, rotate = 180] [fill={rgb, 255:red, 0; green, 0; blue, 0 }  ][line width=0.08]  [draw opacity=0] (8.93,-4.29) -- (0,0) -- (8.93,4.29) -- cycle    ;
\draw  [color={rgb, 255:red, 0; green, 0; blue, 0 }  ,draw opacity=1 ][fill={rgb, 255:red, 255; green, 255; blue, 255 }  ,fill opacity=1 ][dash pattern={on 4.5pt off 4.5pt}] (92.23,203.3) .. controls (92.23,195.6) and (98.47,189.35) .. (106.18,189.35) -- (407.78,189.35) .. controls (415.49,189.35) and (421.73,195.6) .. (421.73,203.3) -- (421.73,245.15) .. controls (421.73,252.85) and (415.49,259.1) .. (407.78,259.1) -- (106.18,259.1) .. controls (98.47,259.1) and (92.23,252.85) .. (92.23,245.15) -- cycle ;
\draw  [fill={rgb, 255:red, 255; green, 255; blue, 255 }  ,fill opacity=1 ] (452.34,230.5) .. controls (452.34,225.02) and (456.78,220.59) .. (462.25,220.59) -- (532.03,220.59) .. controls (537.5,220.59) and (541.94,225.02) .. (541.94,230.5) -- (541.94,260.23) .. controls (541.94,265.71) and (537.5,270.15) .. (532.03,270.15) -- (462.25,270.15) .. controls (456.78,270.15) and (452.34,265.71) .. (452.34,260.23) -- cycle ;
\draw   (212.74,210.3) .. controls (212.74,204.82) and (217.18,200.39) .. (222.65,200.39) -- (292.43,200.39) .. controls (297.9,200.39) and (302.34,204.82) .. (302.34,210.3) -- (302.34,240.03) .. controls (302.34,245.51) and (297.9,249.95) .. (292.43,249.95) -- (222.65,249.95) .. controls (217.18,249.95) and (212.74,245.51) .. (212.74,240.03) -- cycle ;
\draw   (321.58,210.3) .. controls (321.58,204.82) and (326.02,200.39) .. (331.5,200.39) -- (401.27,200.39) .. controls (406.75,200.39) and (411.18,204.82) .. (411.18,210.3) -- (411.18,240.03) .. controls (411.18,245.51) and (406.75,249.95) .. (401.27,249.95) -- (331.5,249.95) .. controls (326.02,249.95) and (321.58,245.51) .. (321.58,240.03) -- cycle ;
\draw    (42.09,223.91) -- (88.93,224.29) ;
\draw [shift={(91.93,224.32)}, rotate = 180.47] [fill={rgb, 255:red, 0; green, 0; blue, 0 }  ][line width=0.08]  [draw opacity=0] (8.93,-4.29) -- (0,0) -- (8.93,4.29) -- cycle    ;
\draw    (421.73,245.15) -- (448.93,245.3) ;
\draw [shift={(451.93,245.32)}, rotate = 180.32] [fill={rgb, 255:red, 0; green, 0; blue, 0 }  ][line width=0.08]  [draw opacity=0] (8.93,-4.29) -- (0,0) -- (8.93,4.29) -- cycle    ;
\draw    (42.43,295.29) -- (497.13,294.85) ;
\draw    (497.13,294.85) -- (497.13,272.65) ;
\draw [shift={(497.13,269.65)}, rotate = 90] [fill={rgb, 255:red, 0; green, 0; blue, 0 }  ][line width=0.08]  [draw opacity=0] (8.93,-4.29) -- (0,0) -- (8.93,4.29) -- cycle    ;
\draw    (421.71,204.86) -- (588.57,205.56) ;
\draw [shift={(591.57,205.57)}, rotate = 180.24] [fill={rgb, 255:red, 0; green, 0; blue, 0 }  ][line width=0.08]  [draw opacity=0] (8.93,-4.29) -- (0,0) -- (8.93,4.29) -- cycle    ;
\draw   (102.12,210.34) .. controls (102.12,204.87) and (106.55,200.43) .. (112.03,200.43) -- (181.8,200.43) .. controls (187.28,200.43) and (191.72,204.87) .. (191.72,210.34) -- (191.72,240.08) .. controls (191.72,245.55) and (187.28,249.99) .. (181.8,249.99) -- (112.03,249.99) .. controls (106.55,249.99) and (102.12,245.55) .. (102.12,240.08) -- cycle ;

\draw (50,279) node [anchor=north west][inner sep=0.75pt]    {$\mathcal{E}$};
\draw (562.75,229) node [anchor=north west][inner sep=0.75pt]    {$\mathcal{E} '$};
\draw (126,210) node [anchor=north west][inner sep=0.75pt]   [align=left] {Feature};
\draw (115,225) node [anchor=north west][inner sep=0.75pt]   [align=left] {Generation};
\draw (235,210) node [anchor=north west][inner sep=0.75pt]   [align=left] {Reward};
\draw (234,225) node [anchor=north west][inner sep=0.75pt]   [align=left] {Shaping};
\draw (346,210) node [anchor=north west][inner sep=0.75pt]   [align=left] {Feature};
\draw (342,225) node [anchor=north west][inner sep=0.75pt]   [align=left] {Selection};
\draw (47.62,208.26) node [anchor=north west][inner sep=0.75pt]    {$\mathcal{D} '$};
\draw (427,228) node [anchor=north west][inner sep=0.75pt]    {$\mathcal{F} '$};
\draw (461,231) node [anchor=north west][inner sep=0.75pt]   [align=left] {Environment};
\draw (464,246) node [anchor=north west][inner sep=0.75pt]   [align=left] {Engineering};
\draw (500,278) node [anchor=north west][inner sep=0.75pt]    {$\mathcal{E}$};
\draw (562.32,190.3) node [anchor=north west][inner sep=0.75pt]    {$\mathcal{D} ''$};

\end{tikzpicture}

%% file: chapters/06_experiments.tex
\section{Experimental Results}
\label{sec:experiments}

In this section, we use the Python implementation of \libname{} on $3$ RL problems.
In addition to the presented stages, the library also allows to create newly defined \blocks, if needed, and a set of analysis tools. 
The implementation of the framework is available at \url{https://github.com/arlo-lib/ARLO}. The implemented methods are reported in Appendix~\ref{apx:implementation}.
The \texttt{Policy Generation} stages have been integrated with the MushroomRL~\citep{DBLP:journals/jmlr/DEramoTBRP21} library.\footnote{The \libname{} library includes an easy procedure to integrate algorithms coming from other RL libraries.}
The optimization of the tunable units has been performed using a genetic algorithm as described in~\ref{apx:details_exp}.
A comprehensive description of all the features available in the \libname{} library, as well as details on how to integrate already developed methods for RL, are provided at \url{https://arlo-lib.github.io/arlo-lib}.

In Sections~\ref{sec:lqg} and~\ref{sec:HalfCheetah}, we present the results our online pipelines whose \texttt{Policy Generation} stages contain tunable units to select the best hyper-parameters over two simulated problems. In Section~\ref{sec:dam}, we consider an offline pipeline including tunable \texttt{Feature Engineering} and \texttt{Policy Generation} stages on a realistic dam control problem.
The experimental details are reported in the supplementary material.

For the different experiments we considered different seeds. These did not only influence the RL algorithm, but were used for all the components making up the considered pipeline.

\subsection{Linear Quadratic Gaussian Regulator}\label{sec:lqg}
In this experiment, we address a Linear-Quadratic Gaussian Regulator~\citep[LQG,][]{dorato1994linear} by the state dynamics $s_{t+1} = A s_t + B a_t + \sigma$, where $s_{t}$ is the state at time $t$, $a_{t}$ is the action at time $t$, $A$ is the state dynamic matrix, $B$ is the action dynamic matrix, and $\sigma$ is a Gaussian white noise. The reward function is $r_{t+1} = - s_t^T Q s_t - a_t^T R a_t$, where $Q$ and $R$ are the state and action cost weight matrices respectively\footnote{The details about the hyper-parameters configuration space, the tuning procedure, and the compute requirements for the LQG experiment can be found in the Appendix~\ref{sec:supplementary_lqg}.}.
The discount factor is equal to $\gamma = 0.9$ and the time horizon is $T = 15$.

We employ the \emph{Soft-Actor Critic}~\citep[SAC,][]{DBLP:conf/icml/HaarnojaZAL18} algorithm.
To tune its hyper-parameters, we create an online RL pipeline, using the expected return (Eq.~\eqref{eq:discrew}) as performance index, and a genetic algorithm~\citep[like in][]{sehgal2019deep} as tuning algorithm. The results are obtained after $50$ generations of the genetic algorithm, each using a population of $20$ agents.

\begin{table}[h!]
  \captionof{table}{Results achieved tuning SAC hyper-parameters on an LQG environment.}
  \label{table:saclqgresults}
  \centering
  \small
  \begin{tabular}{c|cc}
    \toprule
    Method & Default & Tuned  \\
    \toprule
    \citet{van1981generalized} & \multicolumn{2}{c}{$-7.2 \ (4.9)$} \\
    \midrule
    $1^{st}$ Seed & $-59.0 \ (24.0)$ & $-8.6 \ (4.7)$ \\
    $2^{nd}$ Seed & $-67.4 \ (16.1)$ & $-8.2 \ (5.1)$ \\
    $3^{rd}$ Seed & $-52.4 \ (12.5)$  & $-8.7 \ (4.7)$ \\
    \bottomrule
  \end{tabular}
\end{table}
\begin{table}[h!]
\centering
 \captionof{table}{Results achieved tuning DDPG hyper-parameters on HalfCheetah-v3 environment.}
  \label{table:ddpghalfcheetahresults}
  \small
  \begin{tabular}{c|cc}
    \toprule
    Method & Default & Tuned \\
    \toprule
    \citet{islam2017reproducibility}& \multicolumn{2}{c}{$3725.3 \ (512.8)$}\\
    \toprule
    $1^{st}$ Seed & $1157.7 \ (45.6)$ & $3407.2 \ (952.1)$ \\
    $2^{nd}$ Seed & $850.8 \ (78.9)$ & $4624.6 \ (110.9)$ \\
    $3^{rd}$ Seed & $956.2 \ (34.2)$ & $3076.9 \ (77.9) $ \\
    \bottomrule
  \end{tabular}
\end{table}

\textbf{Results}~~
We compare the results provided by the \libname framework with the optimal solution~\citep{van1981generalized}. In Table~\ref{table:saclqgresults}, we report the estimated expected return, averaged over $100$ episodes (with the standard deviation in brackets), for the default configuration and the corresponding tuned policy over three different seeds.
Even if the performance of the tuned algorithms does not match the one of the optimal solution, the \emph{default} hyper-parameter configuration of SAC is notably under-performing ($\approx 5$ times worse) compared to the tuned configuration ($\approx 1.2$ times worse).
This result suggests that the proposed framework can generate solutions compatible with the optimal one without exploiting specific domain knowledge about the problem.

\subsection{HalfCheetah}\label{sec:HalfCheetah}
In the second experiment, we apply the online RL pipeline to the MuJoCo HalfCheetah-v3 environment from OpenAI Gym~\citep{openAIGym}.\footnote{\url{https://www.gymlibrary.ml/pages/environments/mujoco/half_cheetah}.}
As learning algorithm for the \texttt{Policy Generation} we employ the \emph{Deep Deterministic Policy Gradient}~\citep[DDPG,][]{DBLP:journals/corr/LillicrapHPHETS15}, whose hyper-parameter tuning is known to be a challenging task~\citep{islam2017reproducibility}\footnote{The details about the hyper-parameters configuration space, the tuning procedure, and the compute requirements for the HalfCheetah experiment can be found in the Appendix~\ref{sec:supplementary_halfcheetah}.}.
The hyper-parameters of DDPG have been tuned using the undiscounted cumulative reward as a performance index and, as a tuner, a genetic algorithm. We employ a discount factor $\gamma = 1$ and the time horizon to $T = 1000$.

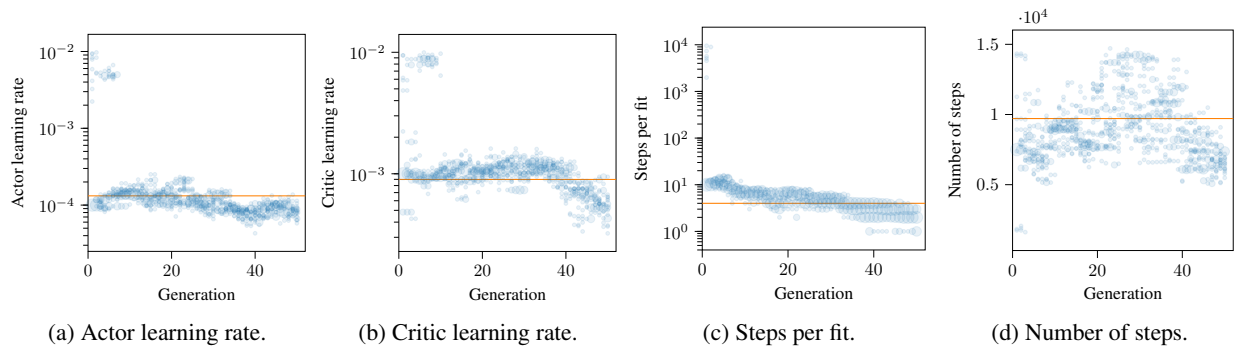
\begin{figure}[t!]
    \centering
    \begin{subfigure}[b]{0.245\textwidth}
        \centering
        \resizebox{\linewidth}{!}{\input{images/ddpg_actor_lr}}
        \caption{Actor learning rate.}
        \label{fig:ddpg_actor_lr}
    \end{subfigure}
    \begin{subfigure}[b]{0.245\textwidth}
        \centering
        \resizebox{\linewidth}{!}{\input{images/ddpg_critic_lr}}
        \caption{Critic learning rate.}
        \label{fig:ddpg_critic_lr}
    \end{subfigure}
    \begin{subfigure}[b]{0.245\textwidth}
        \centering
        \resizebox{.99\linewidth}{!}{\input{images/ddpg_n_steps_per_fit}}
        \caption{Steps per fit.}
        \label{fig:ddpg_n_steps_per_fit}
    \end{subfigure}
    \begin{subfigure}[b]{0.245\textwidth}
        \centering
        \resizebox{.97\linewidth}{!}{\input{images/ddpg_n_steps}}
        \caption{Number of steps.}
        \label{fig:ddpg_n_steps}
    \end{subfigure}
    \caption{Values of the hyper-parameters generated by the genetic optimization procedure over the $50$ generations. The orange line corresponds to the best found value.}
    \label{fig:ddpg_hyperparams}
\end{figure}

\textbf{Results}~~
In Table~\ref{table:ddpghalfcheetahresults}, we report the estimated total reward, averaged over $100$ episodes, for the default and tuned configurations over three different seeds (the standard deviation is provided in brackets). The provided performances are in line with the literature ones~\citep{islam2017reproducibility} and show that the proposed pipeline provides an automatic way of achieving competitive performance.
In Figure~\ref{fig:ddpg_hyperparams}, we report the different hyper-parameters selected during the learning phase by individuals (agents) used in the genetic algorithm optimization procedure throughout the tuning procedure.
These results show how some of the parameters have a strong influence on the reward obtained by the agents, \ie the actor and critic learning rate and the steps per fit (Figures~\ref{fig:ddpg_actor_lr}, \ref{fig:ddpg_critic_lr}, and \ref{fig:ddpg_n_steps_per_fit}, respectively), which implies that the value of the parameter concentrates around the optimal value after a few generations of the genetic algorithm.
Conversely, those which do not influence the outcome of the optimization procedure, \ie the steps (Figure~\ref{fig:ddpg_n_steps}), continue to explore the available range until the end of the generations.

\subsection{Dam}\label{sec:dam}
To showcase the capabilities of our framework, we propose an experiment with a more complex offline RL pipeline that includes \texttt{Data Generation}, \texttt{Feature Engineering}, \texttt{Policy Generation}, and \texttt{Policy Evaluation} \blocks\footnote{The details about the hyper-parameters configuration space, the tuning procedure, and the compute requirements for the Dam experiment can be found in the Appendix~\ref{sec:dam_supplementary}.}.
The selected environment consists of the control of a water reservoir (dam) that models the dynamics of a real alpine lake~\citep{DBLP:conf/adprl/CastellettiGRS11}. The agent observes the current level of the lake and the sequence of the most recent $30$ daily inflows. The actuation consists of the amount of daily water release. The goal of the agent is to trade-off between avoiding floods and fulfilling the downstream water demand. The dataset is generated using a random uniform policy.
The \texttt{Feature Engineering} \block performs forward feature selection via mutual information~\citep[as presented in][]{DBLP:conf/ijcnn/BerahaMPTR19} to identify a subset of the available inflows features. The \texttt{Policy Generation} \block uses the Fitted Q-Iteration~\citep[FQI,][]{DBLP:journals/jmlr/ErnstGW05} algorithm. The hyper-parameters of FQI are fixed to a hand-tuned configuration as the one presented by~\citet{tirinzoni2018importance}. The objective of this experiment is to show in a realistic environment that tuning the hyper-parameters of a \texttt{Feature Engineering} \block is beneficial for the final performance.
\begin{table}
\captionof{table}{Results achieved tuning the hyper-parameters of a Feature Engineering \block.}\label{tab:dam}
  \centering
  \small
  \begin{tabular}{c|cc}
    \toprule
    Method & Discounted Reward \\
    \toprule
    Baseline & $-1649.85$ \ ($112.88$)\\
    Tuned Configuration & $-1224.67$ \ ($124.41$) \\
    \bottomrule
  \end{tabular}
\end{table}
\textbf{Results}~~
In Table~\ref{tab:dam}, we report the estimated expected return over $10$ episodes for the baseline configuration (standard deviation in brackets), in which all the features have been considered, and for the tuned configuration, in which only a subset of the features was selected automatically by the pipeline. We observe that the result achieved by the tuned agent significantly outperforms the baseline one, meaning that the feature selection techniques select only the most informative feature for the problem, with beneficial effects on the successive learning phase.

%% file: images/ddpg_actor_lr.tex
\begin{tikzpicture}
\definecolor{color0}{rgb}{0.12156862745098,0.466666666666667,0.705882352941177}
\begin{axis}[
height=6cm,
width=6cm,
tick align=outside,
tick pos=left,
x grid style={white!69.0196078431373!black},
xtick style={color=black},
y grid style={white!69.0196078431373!black},
ytick style={color=black},
xlabel={Generation},
ylabel={Actor learning rate},
xmin=0, 
xmax=52,
ymode=log]
\addplot [semithick, color0, mark size=3.4965075614664802, opacity=0.1, only marks] 
table {
2.0 0.0001
};
\addplot [semithick, color0, mark size=1.0986122886681098, opacity=0.1, only marks] 
table {
1.0 0.002232290931272232
1.0 0.003249595288885752
1.0 0.004149171684023501
1.0 0.00588159411507698
1.0 0.00787635269194737
1.0 0.008324199312714956
1.0 0.009238101440579955
1.0 0.009435499592876631
2.0 9.65732599960901e-05
2.0 0.005161747394760756
2.0 0.005301465849619686
2.0 0.008324199312714956
2.0 0.009724378100781937
3.0 9.132332230671343e-05
3.0 9.970544757152773e-05
3.0 0.00010928392955711663
3.0 0.00011225326036629982
3.0 0.00011739293285173736
3.0 0.0001375012881895844
3.0 0.004778558262399909
3.0 0.005301465849619686
4.0 8.181367591803508e-05
4.0 8.987104770907539e-05
4.0 0.0001
4.0 0.00013081950577048583
4.0 0.00013216669477478693
4.0 0.0001375012881895844
4.0 0.004307227418769194
4.0 0.0055805960444488675
5.0 8.81939094883297e-05
5.0 9.132332230671343e-05
5.0 0.00013216669477478693
5.0 0.0001444369576156475
5.0 0.00014578171508542305
5.0 0.0046526208715321005
5.0 0.004764482902992046
5.0 0.004999479483199271
5.0 0.005564158263276714
5.0 0.0055805960444488675
5.0 0.006098694650082737
6.0 9.65732599960901e-05
6.0 0.00011773973810961614
6.0 0.00013081950577048583
6.0 0.00013190462833334883
6.0 0.00013216669477478693
6.0 0.0001345220342732922
6.0 0.0001399388500926737
6.0 0.0001444369576156475
6.0 0.00014578171508542305
6.0 0.0001545055658219591
6.0 0.00016323413216508186
6.0 0.00016379044864238315
6.0 0.004351084077747682
6.0 0.0046526208715321005
6.0 0.004793645796160264
6.0 0.004999479483199271
6.0 0.0050845669158048465
6.0 0.006657747406071013
7.0 0.00011719705742765398
7.0 0.00011979981848237813
7.0 0.0001331084415243712
7.0 0.0001361727805237919
7.0 0.00013832935385624638
7.0 0.0001444369576156475
7.0 0.00014578171508542305
7.0 0.00015986579626749966
7.0 0.00016379044864238315
8.0 0.000126326961666934
8.0 0.0001361727805237919
8.0 0.00014224443037645667
8.0 0.0001444369576156475
8.0 0.00014696415655682764
8.0 0.00015100967496502052
8.0 0.0001568644708566683
8.0 0.00015986579626749966
8.0 0.00016323413216508186
8.0 0.00016379044864238315
8.0 0.00019200758868989713
9.0 0.00013633097025719055
9.0 0.00014389947949626654
9.0 0.00014696415655682764
9.0 0.00015202887047866882
9.0 0.0001568644708566683
9.0 0.00015802711174739965
9.0 0.0001590717889299049
9.0 0.00019200758868989713
10.0 0.00010188282876413368
10.0 0.00013029490369656122
10.0 0.00014389947949626654
10.0 0.00014696415655682764
10.0 0.00015633167855672627
10.0 0.0001568644708566683
10.0 0.00015802711174739965
10.0 0.0001809235547223602
11.0 0.00011984096580279573
11.0 0.00015396174915622245
11.0 0.00015802711174739965
11.0 0.0001590717889299049
11.0 0.0001797735555121323
11.0 0.0001809235547223602
12.0 0.00010581307041854353
12.0 0.00011520210790921837
12.0 0.0001252208059420531
12.0 0.00012789820441704656
12.0 0.00014593466009700612
12.0 0.00015396174915622245
12.0 0.00015633167855672627
12.0 0.0001809235547223602
12.0 0.00018892202164104725
12.0 0.00021487432781493748
13.0 0.00010531041398801784
13.0 0.00011520210790921837
13.0 0.00011833910803913332
13.0 0.00012310252016238155
13.0 0.00012789820441704656
13.0 0.00013862316219211
13.0 0.0001405799448698526
13.0 0.00014389741780465845
13.0 0.00016377863920176238
13.0 0.00016428721496517266
13.0 0.00016728045885175107
13.0 0.00017865185006629393
13.0 0.0001809235547223602
13.0 0.00018892202164104725
14.0 9.79046142929549e-05
14.0 9.855402569415879e-05
14.0 0.00010014585813330336
14.0 0.00010664339590157236
14.0 0.00012310252016238155
14.0 0.00013194424770824925
14.0 0.00013832935385624638
14.0 0.00014054574780574664
14.0 0.00014310643215658503
14.0 0.00015466616249384368
14.0 0.00016172220880600883
14.0 0.00018317488906114527
14.0 0.00021137307401579528
15.0 8.569531264435382e-05
15.0 8.829692582679221e-05
15.0 9.855402569415879e-05
15.0 0.00010014585813330336
15.0 0.00011700045312546532
15.0 0.00012220178495762332
15.0 0.00012762816444402021
15.0 0.00013832935385624638
15.0 0.00015456030684571419
15.0 0.00018317488906114527
15.0 0.00021137307401579528
15.0 0.00021188120601111282
15.0 0.0002513302999247348
16.0 8.829692582679221e-05
16.0 0.00012041522796308707
16.0 0.00013053044254563066
16.0 0.00013120525168399742
16.0 0.00013194424770824925
16.0 0.00013796475781527618
16.0 0.00014571012666753138
16.0 0.00016106841248860566
16.0 0.00018317488906114527
16.0 0.0002513302999247348
17.0 9.910299950487275e-05
17.0 0.00011000830722539617
17.0 0.00011166529419845281
17.0 0.00012041522796308707
17.0 0.00012444414374329594
17.0 0.00013796475781527618
17.0 0.00014489403042862117
17.0 0.00014755447383239638
18.0 0.00011000830722539617
18.0 0.00014571012666753138
18.0 0.00014755447383239638
18.0 0.0001550437851502175
18.0 0.00016064388288674163
18.0 0.00016303141828904699
19.0 8.822965004711161e-05
19.0 9.910299950487275e-05
19.0 0.00010517822123378575
19.0 0.00011911167170586487
19.0 0.00013120525168399742
19.0 0.00013194424770824925
19.0 0.00013796475781527618
19.0 0.00014050968216575995
19.0 0.0001710156525068311
19.0 0.00018705108114263464
19.0 0.00018850619908858008
20.0 8.822965004711161e-05
20.0 9.35386921411449e-05
20.0 9.601193321402941e-05
20.0 9.910299950487275e-05
20.0 0.00010531041398801784
20.0 0.00011770525337363637
20.0 0.00013463778931105606
20.0 0.00013796475781527618
20.0 0.000143378080607407
20.0 0.00014923595227999453
20.0 0.00016303141828904699
20.0 0.00016899637475395422
20.0 0.0001710156525068311
20.0 0.0001934277863562061
21.0 7.111896883404553e-05
21.0 9.735658373527421e-05
21.0 0.00010531041398801784
21.0 0.00010594390673142216
21.0 0.00011326015349139949
21.0 0.0001242290750040107
21.0 0.00012556626131355765
21.0 0.00012988129670732208
21.0 0.00013186815002183547
21.0 0.00014050968216575995
21.0 0.0001559156666966264
21.0 0.0001934277863562061
21.0 0.0002187771001439042
22.0 8.582790643103824e-05
22.0 9.735658373527421e-05
22.0 0.00010313933215859469
22.0 0.00010742765489313817
22.0 0.00010794940117875568
22.0 0.00010859876869818604
22.0 0.00011326015349139949
22.0 0.00012195147401284796
22.0 0.0001284682412202206
22.0 0.00013194424770824925
22.0 0.00015437051302115017
22.0 0.0001559156666966264
22.0 0.00015909856661133382
22.0 0.00015982277741498861
22.0 0.00019114387596524523
22.0 0.00021068516683451125
22.0 0.0002187771001439042
22.0 0.00023250062939198082
23.0 0.00010531041398801784
23.0 0.00011798093826315882
23.0 0.00015140798130059137
23.0 0.0001559156666966264
23.0 0.00015982277741498861
23.0 0.00016308543802748677
23.0 0.0001654453615365683
23.0 0.00017284356342285462
23.0 0.00018282943191467419
23.0 0.0001930514626346724
23.0 0.00019827491809936763
23.0 0.00020220805610108837
23.0 0.00020902870932161261
23.0 0.0002187771001439042
23.0 0.00022214073797623861
23.0 0.00022750460422219632
24.0 9.23821854316651e-05
24.0 9.735658373527421e-05
24.0 0.00012258969468062554
24.0 0.0001232097868316819
24.0 0.00012418546677275482
24.0 0.00012705705575337541
24.0 0.00015140798130059137
24.0 0.0001559156666966264
24.0 0.00016308543802748677
24.0 0.00020220805610108837
24.0 0.0002187771001439042
25.0 8.423492841414218e-05
25.0 0.00010037406062603295
25.0 0.00011135462394444664
25.0 0.00012564137771652696
25.0 0.00015140798130059137
25.0 0.00015766706837185948
25.0 0.00016308543802748677
25.0 0.0002187771001439042
26.0 8.907316949795429e-05
26.0 9.135803232820327e-05
26.0 0.00010037406062603295
26.0 0.00010543449160561072
26.0 0.00010754020877066652
26.0 0.00011743441265524748
26.0 0.00011772242587931603
27.0 8.092247541458926e-05
27.0 8.923902078216735e-05
27.0 9.525620986516717e-05
27.0 9.621536823876857e-05
27.0 0.00010054858617538097
27.0 0.00010858663766133265
27.0 0.00011351509810598251
27.0 0.00012705705575337541
27.0 0.00012910755973378533
27.0 0.00013116574355119328
28.0 7.413659252474155e-05
28.0 8.092247541458926e-05
28.0 9.020904829951265e-05
28.0 9.100091314255529e-05
28.0 9.135803232820327e-05
28.0 9.904431540444362e-05
28.0 0.00012105630417352174
28.0 0.00012705705575337541
28.0 0.00013396221898309414
29.0 7.534411007160387e-05
29.0 7.57430855950245e-05
29.0 8.98319464152465e-05
29.0 9.904431540444362e-05
29.0 9.929550023747365e-05
29.0 0.00010056754954482767
29.0 0.00010858663766133265
29.0 0.00012105630417352174
29.0 0.00014926004055249945
29.0 0.00015723362551970914
30.0 6.56681895725633e-05
30.0 7.24505094428155e-05
30.0 8.092247541458926e-05
30.0 8.45056365010008e-05
30.0 8.787889093255165e-05
30.0 8.98319464152465e-05
30.0 9.23414307359999e-05
30.0 9.929550023747365e-05
30.0 0.00010313933215859469
30.0 0.00010858663766133265
30.0 0.00010902071387683043
30.0 0.00011491733279285505
30.0 0.00011599058871619842
30.0 0.00011822210476035585
30.0 0.00012538805525846034
31.0 8.196720959425165e-05
31.0 8.98319464152465e-05
31.0 9.23414307359999e-05
31.0 0.00011822210476035585
31.0 0.00012472757563072352
31.0 0.00014784913578218283
32.0 7.70637033874294e-05
32.0 8.024146032307536e-05
32.0 8.98319464152465e-05
32.0 0.00010163177645962385
32.0 0.00010781095311887003
32.0 0.00010870967980988223
32.0 0.00012160192925898209
32.0 0.00014193536852746504
32.0 0.00014745577963181088
32.0 0.00015990110509500827
33.0 8.196720959425165e-05
33.0 8.557812310139968e-05
33.0 8.595701380510732e-05
33.0 8.983087643604993e-05
33.0 8.98319464152465e-05
33.0 9.046417111466723e-05
33.0 9.095230908014027e-05
33.0 9.517855494463388e-05
33.0 0.00010163177645962385
33.0 0.00010413510224852084
33.0 0.00010768068605759726
33.0 0.00010902071387683043
33.0 0.00013033405870996459
33.0 0.0001503830631647694
34.0 7.70637033874294e-05
34.0 8.595701380510732e-05
34.0 9.06579342105893e-05
34.0 9.735840691295019e-05
34.0 9.973603860478442e-05
34.0 0.00010084338236949504
34.0 0.000101001217962313
34.0 0.0001017103288491744
34.0 0.00010242737194811882
34.0 0.00010902071387683043
34.0 0.00011229240653413242
34.0 0.00013033405870996459
34.0 0.00013396221898309414
34.0 0.0001533253140006044
35.0 7.204301992289131e-05
35.0 7.221295103310732e-05
35.0 7.988189768776343e-05
35.0 8.122283924044832e-05
35.0 8.188668065633386e-05
35.0 8.245203049251439e-05
35.0 8.458955514795607e-05
35.0 8.595701380510732e-05
35.0 9.095230908014027e-05
35.0 9.470333241489451e-05
35.0 9.704416565265975e-05
35.0 0.0001017103288491744
35.0 0.0001038061875222359
35.0 0.00010567726864976252
36.0 6.54192141066889e-05
36.0 6.612401958046832e-05
36.0 6.875896212664494e-05
36.0 7.221295103310732e-05
36.0 7.294098435877829e-05
36.0 7.692762409072822e-05
36.0 8.033367423312653e-05
36.0 8.041406903829664e-05
36.0 8.072518792492939e-05
36.0 8.188668065633386e-05
36.0 8.585248830939923e-05
36.0 8.595701380510732e-05
36.0 8.632171065835403e-05
36.0 8.660200540073634e-05
36.0 9.06579342105893e-05
36.0 9.470333241489451e-05
36.0 9.592735185972662e-05
36.0 0.0001017103288491744
37.0 5.9903124476194216e-05
37.0 6.42412136006354e-05
37.0 6.529315025681559e-05
37.0 6.612401958046832e-05
37.0 7.221295103310732e-05
37.0 7.5378155368089e-05
37.0 7.666856319395434e-05
37.0 7.692762409072822e-05
37.0 7.976993769936579e-05
37.0 8.122283924044832e-05
37.0 8.480533681388527e-05
37.0 8.632171065835403e-05
37.0 8.711347791024877e-05
37.0 9.299889870713407e-05
38.0 5.415353322973853e-05
38.0 5.6873135681846295e-05
38.0 6.69527221396892e-05
38.0 7.120525752291272e-05
38.0 7.208961856413506e-05
38.0 7.270490467072313e-05
38.0 8.122283924044832e-05
38.0 8.52924205955906e-05
38.0 8.585248830939923e-05
38.0 8.711347791024877e-05
38.0 8.75056175511191e-05
38.0 8.807655301607075e-05
38.0 8.942194473267834e-05
38.0 9.51776471460337e-05
39.0 5.362212557253515e-05
39.0 5.6873135681846295e-05
39.0 6.465242973610266e-05
39.0 6.548146071174147e-05
39.0 6.577659442055225e-05
39.0 6.69527221396892e-05
39.0 6.768561695282121e-05
39.0 7.270490467072313e-05
39.0 7.694611838660021e-05
39.0 8.072518792492939e-05
39.0 8.124386018973075e-05
39.0 8.221676073359709e-05
39.0 0.00010485138473452816
40.0 4.294571242253117e-05
40.0 5.950810591438441e-05
40.0 6.297948925713136e-05
40.0 6.53265031995642e-05
40.0 6.69527221396892e-05
40.0 7.270490467072313e-05
40.0 7.349901580417634e-05
40.0 7.391817850191002e-05
40.0 7.558480601089475e-05
40.0 7.976993769936579e-05
40.0 7.988209517195367e-05
40.0 8.124386018973075e-05
40.0 8.557549005421104e-05
40.0 8.587347864643554e-05
40.0 8.942194473267834e-05
41.0 5.531189166165099e-05
41.0 6.61117622847418e-05
41.0 7.141190340177083e-05
41.0 7.160291724152806e-05
41.0 7.349901580417634e-05
41.0 7.36068085414538e-05
41.0 7.391817850191002e-05
41.0 7.803352187848149e-05
41.0 8.385726306862653e-05
41.0 8.587347864643554e-05
41.0 9.330087004953989e-05
41.0 9.389045896981091e-05
41.0 0.00010485138473452816
41.0 0.00010626203435356991
42.0 6.61117622847418e-05
42.0 6.7693197498074e-05
42.0 6.811781534391669e-05
42.0 7.073432357194657e-05
42.0 7.257583120922443e-05
42.0 7.36068085414538e-05
42.0 7.429534136220413e-05
42.0 0.00010137141344610754
42.0 0.00010357946380538417
42.0 0.00010664937052242066
42.0 0.00012207814601425507
43.0 6.7693197498074e-05
43.0 6.981190083177892e-05
43.0 7.183313499415472e-05
43.0 8.195981409913551e-05
43.0 8.235575622676602e-05
43.0 8.343497521397876e-05
43.0 8.550997409007913e-05
43.0 8.807655301607075e-05
43.0 8.938334006119437e-05
43.0 8.945316738557475e-05
43.0 9.065442507575844e-05
43.0 9.389045896981091e-05
43.0 0.00010031196598025599
43.0 0.00010137141344610754
43.0 0.00011623447581283757
43.0 0.00011895119298539571
44.0 7.073432357194657e-05
44.0 7.563893554668167e-05
44.0 8.159464458728374e-05
44.0 8.771951257567202e-05
44.0 8.807655301607075e-05
44.0 8.945316738557475e-05
44.0 0.00010031196598025599
44.0 0.00010137141344610754
44.0 0.00010187981677678319
44.0 0.00010239460539099423
44.0 0.00010649018550437218
44.0 0.00010837616471000283
44.0 0.0001332767582911107
45.0 7.35181369017568e-05
45.0 8.03465140766104e-05
45.0 8.110538013492155e-05
45.0 8.945316738557475e-05
45.0 9.5390160705338e-05
45.0 9.541066952543166e-05
45.0 0.00010779362300908624
45.0 0.00010837616471000283
45.0 0.0001105531487611431
46.0 5.7689226472529514e-05
46.0 9.126491918307179e-05
46.0 9.438584568795705e-05
46.0 9.814090326700159e-05
46.0 9.877426443684169e-05
46.0 0.00011165025470896115
46.0 0.0001255288062664337
47.0 6.291443928147892e-05
47.0 7.073432357194657e-05
47.0 8.860104871413853e-05
47.0 9.07477850710642e-05
47.0 9.438584568795705e-05
47.0 9.814090326700159e-05
47.0 0.0001093242464278081
47.0 0.00011801964154024959
48.0 6.291443928147892e-05
48.0 6.654217537098718e-05
48.0 6.997455297936258e-05
48.0 7.120754226056868e-05
48.0 7.640537313200285e-05
48.0 7.700148196207629e-05
48.0 9.146189791166281e-05
48.0 0.00010413886656658941
48.0 0.0001093242464278081
48.0 0.00011489777761535653
48.0 0.00011628662063953487
48.0 0.00012833797821514632
49.0 7.120754226056868e-05
49.0 7.640537313200285e-05
49.0 7.6872789591965e-05
49.0 8.380277423549461e-05
49.0 8.514729136830472e-05
49.0 8.5463425611679e-05
49.0 8.752765206830821e-05
49.0 8.806454904880255e-05
49.0 8.930471302832915e-05
49.0 9.162408610030839e-05
49.0 9.588925770374551e-05
49.0 9.814090326700159e-05
49.0 0.00011311684028450333
49.0 0.00013694577658547558
50.0 5.79522214419836e-05
50.0 6.338600034701161e-05
50.0 6.465655909041154e-05
50.0 7.077995007269614e-05
50.0 7.952911638907225e-05
50.0 7.959689585455697e-05
50.0 8.266344670695314e-05
50.0 8.5463425611679e-05
50.0 8.815260832299626e-05
50.0 9.19899567675475e-05
50.0 9.588925770374551e-05
50.0 0.00010510031873010678
50.0 0.00011366144654134841
};
\addplot [semithick, color0, mark size=3.58351893845611, opacity=0.1, only marks] 
table {
1.0 0.0001
};
\addplot [semithick, color0, mark size=1.791759469228055, opacity=0.1, only marks] 
table {
2.0 0.00011694935511743725
2.0 0.00011773973810961614
3.0 9.013654709749211e-05
3.0 9.65732599960901e-05
3.0 0.00011694935511743725
4.0 9.013654709749211e-05
4.0 9.132332230671343e-05
5.0 0.00011773973810961614
5.0 0.0001286706124564465
5.0 0.00013081950577048583
6.0 0.004764482902992046
7.0 0.00013216669477478693
7.0 0.0001345220342732922
7.0 0.00014389947949626654
7.0 0.004999479483199271
8.0 9.65732599960901e-05
8.0 0.00013832935385624638
9.0 9.65732599960901e-05
12.0 0.00014389741780465845
14.0 0.00011833910803913332
15.0 0.00011819865536185713
15.0 0.00013194424770824925
16.0 0.00010531041398801784
16.0 0.00013832935385624638
16.0 0.00021188120601111282
17.0 0.00013194424770824925
18.0 9.910299950487275e-05
18.0 0.00013194424770824925
19.0 0.00014755447383239638
19.0 0.00016303141828904699
20.0 0.00011178614563607751
20.0 0.00011911167170586487
20.0 0.00013194424770824925
21.0 0.00013194424770824925
21.0 0.00019827491809936763
22.0 0.00010531041398801784
23.0 9.735658373527421e-05
23.0 0.00010859876869818604
24.0 0.00010313933215859469
24.0 0.00010531041398801784
24.0 0.00010859876869818604
26.0 0.00012705705575337541
26.0 0.00013396221898309414
27.0 0.00010754020877066652
27.0 0.00012258969468062554
28.0 0.00010858663766133265
29.0 7.413659252474155e-05
29.0 8.092247541458926e-05
30.0 0.00014926004055249945
31.0 8.45056365010008e-05
32.0 8.45056365010008e-05
32.0 0.00010902071387683043
33.0 7.70637033874294e-05
33.0 0.00013396221898309414
33.0 0.00014745577963181088
34.0 8.412696605760815e-05
34.0 8.983087643604993e-05
34.0 9.095230908014027e-05
36.0 8.807655301607075e-05
37.0 8.072518792492939e-05
37.0 8.585248830939923e-05
37.0 9.06579342105893e-05
38.0 6.529315025681559e-05
38.0 7.976993769936579e-05
38.0 8.072518792492939e-05
40.0 0.00010485138473452816
42.0 8.587347864643554e-05
42.0 9.389045896981091e-05
42.0 0.00010485138473452816
43.0 7.073432357194657e-05
43.0 0.00010485138473452816
44.0 8.235575622676602e-05
44.0 9.389045896981091e-05
45.0 0.00010485138473452816
46.0 9.5390160705338e-05
46.0 0.00010485138473452816
47.0 0.00010239460539099423
47.0 0.00010767581774502335
50.0 6.654217537098718e-05
50.0 8.03465140766104e-05
50.0 8.806454904880255e-05
};
\addplot [semithick, color0, mark size=2.1972245773362196, opacity=0.1, only marks] 
table {
3.0 0.0001
3.0 0.00011773973810961614
4.0 0.004778558262399909
5.0 9.65732599960901e-05
7.0 9.65732599960901e-05
15.0 0.00010531041398801784
17.0 0.00013120525168399742
17.0 0.00014571012666753138
18.0 0.00010531041398801784
18.0 0.00013120525168399742
21.0 0.00011178614563607751
24.0 0.00020902870932161261
25.0 0.00010313933215859469
25.0 0.00010859876869818604
25.0 0.00012258969468062554
25.0 0.00012705705575337541
26.0 0.00010313933215859469
27.0 9.135803232820327e-05
27.0 0.00010313933215859469
28.0 8.923902078216735e-05
28.0 0.00010054858617538097
28.0 0.00010313933215859469
29.0 0.00010313933215859469
29.0 0.00013396221898309414
30.0 0.00013396221898309414
32.0 8.196720959425165e-05
32.0 0.00013396221898309414
35.0 8.983087643604993e-05
35.0 9.06579342105893e-05
39.0 8.942194473267834e-05
40.0 8.807655301607075e-05
41.0 5.950810591438441e-05
41.0 8.807655301607075e-05
42.0 8.807655301607075e-05
44.0 0.00010485138473452816
45.0 7.073432357194657e-05
45.0 9.389045896981091e-05
45.0 0.00010239460539099423
46.0 8.03465140766104e-05
46.0 8.945316738557475e-05
46.0 0.00010239460539099423
48.0 9.814090326700159e-05
};
\addplot [semithick, color0, mark size=3.7376696182833684, opacity=0.1, only marks] 
table {
11.0 0.00013832935385624638
};
\addplot [semithick, color0, mark size=2.4849066497880004, opacity=0.1, only marks] 
table {
9.0 0.0001345220342732922
10.0 0.0001590717889299049
16.0 0.00011819865536185713
17.0 0.00010531041398801784
18.0 0.00013796475781527618
31.0 0.00010902071387683043
39.0 8.807655301607075e-05
};
\addplot [semithick, color0, mark size=2.70805020110221, opacity=0.1, only marks] 
table {
4.0 0.00011773973810961614
8.0 0.0001345220342732922
14.0 0.00010531041398801784
19.0 0.00010531041398801784
48.0 8.03465140766104e-05
};
\addplot [semithick, color0, mark size=2.8903717578961645, opacity=0.1, only marks] 
table {
9.0 0.00013832935385624638
13.0 0.00013832935385624638
26.0 0.00012258969468062554
49.0 8.03465140766104e-05
};
\addplot [semithick, color0, mark size=3.1780538303479458, opacity=0.1, only marks] 
table {
10.0 0.00013832935385624638
12.0 0.00013832935385624638
31.0 0.00013396221898309414
47.0 8.03465140766104e-05
};
\addplot [semithick, orange]
table { 
0 0.00013194424770824925
52 0.00013194424770824925
};
\end{axis}
\end{tikzpicture}

%% file: images/ddpg_critic_lr.tex
\begin{tikzpicture}
\definecolor{color0}{rgb}{0.12156862745098,0.466666666666667,0.705882352941177}
\begin{axis}[
height=6cm,
width=6cm,
tick align=outside,
tick pos=left,
x grid style={white!69.0196078431373!black},
xtick style={color=black},
xlabel={Generation},
ylabel={Critic learning rate},
xmin=0, 
xmax=52,
y grid style={white!69.0196078431373!black},
ytick style={color=black},
ymode=log]
\addplot [semithick, color0, mark size=1.0986122886681098, opacity=0.1, only marks] 
table {
1.0 0.002232290931272232
1.0 0.004848823581217198
1.0 0.00588159411507698
1.0 0.006161471918476304
1.0 0.009424987550741464
1.0 0.009435499592876631
2.0 0.0009265657279404006
2.0 0.0010945965325373816
2.0 0.0018281493783056362
2.0 0.004848823581217198
2.0 0.00588159411507698
2.0 0.00787635269194737
2.0 0.008781889689134558
3.0 0.00042663572238007587
3.0 0.0009736453164422528
3.0 0.0010283025721661279
3.0 0.0013071629655515603
3.0 0.008781889689134558
4.0 0.00042663572238007587
4.0 0.0008824981239159401
4.0 0.0013071629655515603
4.0 0.0013300342646776638
4.0 0.0018281493783056362
4.0 0.0021831689501863338
4.0 0.008781889689134558
5.0 0.0008619844318554324
5.0 0.0009265657279404006
5.0 0.0014147198344652966
5.0 0.00705617524941659
5.0 0.009405910065026021
6.0 0.0008461706061105575
6.0 0.0009121513794293555
6.0 0.0009690645600369627
6.0 0.001
6.0 0.0010572417624411141
6.0 0.008014164443953524
6.0 0.009956841294966833
7.0 0.0006832509643783275
7.0 0.0008614645423809332
7.0 0.0008853166726104792
7.0 0.0009397994692018497
7.0 0.0009736453164422528
7.0 0.001052447504278087
7.0 0.007867538979744524
7.0 0.008923816271475737
7.0 0.009057136065674702
7.0 0.009509154799109854
8.0 0.0009121513794293555
8.0 0.0009416863082308665
8.0 0.0009682855962621306
8.0 0.000985976542579278
8.0 0.0010488965488341985
8.0 0.001122304908401232
8.0 0.007578670009868356
8.0 0.007867538979744524
8.0 0.008014164443953524
8.0 0.009537340603560306
9.0 0.0009121513794293555
9.0 0.0009443502689675849
9.0 0.0009917498296678113
9.0 0.0010127134645197334
9.0 0.0010254671809704608
9.0 0.001062105940867237
9.0 0.0010720290603939
9.0 0.008014164443953524
9.0 0.008192051256178567
9.0 0.00860949396687646
9.0 0.008948409925234135
10.0 0.0009281112005052334
10.0 0.0009460172151810901
10.0 0.001032227324667405
10.0 0.001039652243417012
10.0 0.0010465719090918986
10.0 0.0010720290603939
10.0 0.0010855153333157118
10.0 0.0011046205662347687
10.0 0.0011631869612520646
10.0 0.001211544027359025
10.0 0.006426789416773087
10.0 0.009749036673554454
11.0 0.0009072148020606427
11.0 0.0009222366456720418
11.0 0.0009460172151810901
11.0 0.0009513326110404276
11.0 0.0009736863964661819
11.0 0.0010855153333157118
11.0 0.0011079834806246806
11.0 0.001115180612804334
11.0 0.0011419724956972025
11.0 0.0014418106997854381
12.0 0.0007275216935462526
12.0 0.0008339903421663728
12.0 0.0009513326110404276
12.0 0.0009666782011786012
12.0 0.0009736863964661819
12.0 0.0009777453667366796
12.0 0.0010348597373166023
12.0 0.0010547711367238474
12.0 0.0012049353987602812
12.0 0.0012603113672806892
13.0 0.0007339316977815972
13.0 0.0007629655923575146
13.0 0.0008339903421663728
13.0 0.000885090976106997
13.0 0.0009594598930574996
13.0 0.0009603246598242327
13.0 0.000998979146633851
13.0 0.0010547711367238474
13.0 0.001098962749127339
13.0 0.0012586387776166363
14.0 0.0006157991915885332
14.0 0.0008012168225359779
14.0 0.0008336499179038149
14.0 0.0008339903421663728
14.0 0.000885090976106997
14.0 0.0008983532040235747
14.0 0.0010574610411818743
14.0 0.0010720290603939
14.0 0.001098962749127339
14.0 0.0011376249480354174
15.0 0.0006157991915885332
15.0 0.000637120054497633
15.0 0.0008339903421663728
15.0 0.0008494073795556314
15.0 0.0009257311020067219
15.0 0.0011790345885773013
15.0 0.0012040785821548195
15.0 0.0012090271174887241
15.0 0.0012201870441837985
15.0 0.0014143850480150084
15.0 0.0014414563829887156
16.0 0.0006529591451774829
16.0 0.0007755795812180338
16.0 0.0008494073795556314
16.0 0.0009257311020067219
16.0 0.0011542047362290487
16.0 0.0012049353987602812
16.0 0.0012201870441837985
16.0 0.001222496171959703
16.0 0.0015499076239106192
17.0 0.0009097242406342886
17.0 0.0009153336725130123
17.0 0.0010056474572907206
17.0 0.0010242100184749666
17.0 0.0010574610411818743
17.0 0.0010708460084306439
17.0 0.0012049353987602812
17.0 0.0012955914468852588
17.0 0.0013118753696233362
17.0 0.0014030068558791276
18.0 0.0008425336818518755
18.0 0.0009097242406342886
18.0 0.0010056474572907206
18.0 0.001072872517623288
18.0 0.0010949003001334352
18.0 0.0011257563326158655
18.0 0.0011513124236287298
19.0 0.0007058794005990084
19.0 0.0008580888449937513
19.0 0.001023768561439591
19.0 0.0010568026501619236
19.0 0.0010947640836248527
19.0 0.0011513124236287298
19.0 0.0011804223450196476
19.0 0.0012207875330683205
19.0 0.0012381730649030365
20.0 0.0008103415890308784
20.0 0.0008825191198098908
20.0 0.000923868218151944
20.0 0.0009535945999513619
20.0 0.0009863274468581928
20.0 0.0010183486102332215
20.0 0.0010443243883807801
20.0 0.0010708460084306439
20.0 0.0011513124236287298
20.0 0.0011579366197046067
21.0 0.000884949313119252
21.0 0.0011217873948632935
21.0 0.0011579366197046067
21.0 0.0011635178491293584
21.0 0.001208205897086899
21.0 0.0013659560316184997
22.0 0.0008522175929381002
22.0 0.0009535945999513619
22.0 0.0009538616399559033
22.0 0.0009586936215019439
22.0 0.0010242100184749666
22.0 0.001061152253689707
22.0 0.0010679747608678966
22.0 0.0010968712971835457
22.0 0.0011351970710096213
22.0 0.0011668503087980198
22.0 0.0011813703520567882
22.0 0.0013659560316184997
22.0 0.0014001856992652574
22.0 0.0016113454299346197
23.0 0.0009326753528725099
23.0 0.0010067341706232634
23.0 0.0010296027487301
23.0 0.0011217873948632935
23.0 0.0011507662895332564
23.0 0.0011767339780410291
24.0 0.000737657056613416
24.0 0.0008408869053071218
24.0 0.0009032055658328202
24.0 0.0009326753528725099
24.0 0.0009538616399559033
24.0 0.0009886317095860401
24.0 0.0010035194231890476
24.0 0.0010679747608678966
24.0 0.0011277857699609197
24.0 0.0011334976943654545
24.0 0.001187687326675419
24.0 0.0012080562186413827
24.0 0.001237195302277051
24.0 0.001249496015115102
24.0 0.0013767407904077614
25.0 0.0008862268631369493
25.0 0.0009326753528725099
25.0 0.0009744020952081105
25.0 0.0009886317095860401
25.0 0.0010418831974229174
25.0 0.001050393735716573
25.0 0.001088613315226202
25.0 0.001121172317368629
25.0 0.0011382222451991168
25.0 0.0011470926418172798
25.0 0.0011507662895332564
25.0 0.001249496015115102
25.0 0.0012658090669337924
25.0 0.0013146192434302837
25.0 0.0014981747056042147
26.0 0.0007354088146308797
26.0 0.0008862268631369493
26.0 0.0009625227748469075
26.0 0.0009648370311635863
26.0 0.000980324797615498
26.0 0.0010160636535191526
26.0 0.0010418831974229174
26.0 0.001121172317368629
26.0 0.0011470926418172798
26.0 0.0012658090669337924
26.0 0.001272221284403613
26.0 0.0012909649070802216
26.0 0.0013146192434302837
26.0 0.0013216487210022023
26.0 0.001373143930208306
27.0 0.0008599354563045317
27.0 0.0008639932591080704
27.0 0.0008962939577522072
27.0 0.0009625227748469075
27.0 0.0010236493989256301
27.0 0.0010453870567855102
27.0 0.001050393735716573
27.0 0.0010647216685715486
27.0 0.001121172317368629
27.0 0.0011533716660571906
27.0 0.0011813703520567882
27.0 0.0011925703621091807
27.0 0.0013800998746140358
27.0 0.0013817477350539676
28.0 0.0009387120326093644
28.0 0.0009931041308096579
28.0 0.0010236493989256301
28.0 0.0010886579718464003
28.0 0.0011117761161478236
28.0 0.0011807829549918426
28.0 0.0011822751750084537
28.0 0.0013800998746140358
28.0 0.0016133600351503326
29.0 0.0008855250785951432
29.0 0.0009022008333704658
29.0 0.0009387120326093644
29.0 0.0009557304705007423
29.0 0.0009931041308096579
29.0 0.000997968146927071
29.0 0.0010162896484319543
29.0 0.0010236493989256301
29.0 0.0010845748583618417
29.0 0.0011823813241360495
29.0 0.0012981106061527975
29.0 0.0013296964673679373
29.0 0.0013865935108078908
30.0 0.0008631612333784753
30.0 0.0009151616839364472
30.0 0.0009437105197820871
30.0 0.0010162896484319543
30.0 0.0010236493989256301
30.0 0.001099932959102841
30.0 0.001112562574953917
30.0 0.0011813703520567882
30.0 0.0011853972649626095
30.0 0.0012563952587863034
30.0 0.0013793688868879047
31.0 0.0008661619404253652
31.0 0.0008925901698559961
31.0 0.0009056470708748522
31.0 0.0009368353477304843
31.0 0.0010162896484319543
31.0 0.0010241094271701977
31.0 0.0010375381287446167
31.0 0.001099932959102841
31.0 0.001112562574953917
31.0 0.001150801546956867
31.0 0.0011697860475282762
31.0 0.001177469540903602
31.0 0.0012405939324083225
31.0 0.0012753235906499201
32.0 0.000988407456893675
32.0 0.0009931041308096579
32.0 0.001008338533637685
32.0 0.0010241094271701977
32.0 0.0010565860877858962
32.0 0.0010971141787349357
32.0 0.001152145791687945
32.0 0.0011687714839321212
32.0 0.001177469540903602
32.0 0.001227897900817441
32.0 0.0013179367536398077
32.0 0.0013767488392073869
33.0 0.0009457956445522159
33.0 0.000988407456893675
33.0 0.001000497542565353
33.0 0.0010241094271701977
33.0 0.001101837282542011
33.0 0.0011217126795161464
33.0 0.001150801546956867
33.0 0.001177469540903602
33.0 0.0012022913273078356
33.0 0.001227897900817441
33.0 0.0012563952587863034
33.0 0.0012995294893884282
33.0 0.0013102038288623453
33.0 0.0014013858646427565
33.0 0.0014797779597430982
33.0 0.0015974632660270655
34.0 0.000988407456893675
34.0 0.0010091706768807938
34.0 0.001101837282542011
34.0 0.001150801546956867
34.0 0.001177469540903602
34.0 0.001210844254534257
34.0 0.0014797779597430982
35.0 0.0009016722427666978
35.0 0.000954034825805542
35.0 0.0010496618448766937
35.0 0.0011395944486698075
35.0 0.0011490264837232785
36.0 0.0008626731548651891
36.0 0.0008759285446521544
36.0 0.0009181085820890172
36.0 0.0009299196404766084
36.0 0.000954034825805542
36.0 0.0009732866890092451
36.0 0.000976391753058253
36.0 0.0010496618448766937
36.0 0.0010565860877858962
36.0 0.0010773329192303586
37.0 0.0007819876401924652
37.0 0.0008626731548651891
37.0 0.0009571833100592241
37.0 0.0010048935884139306
37.0 0.0010565860877858962
37.0 0.001076864106649322
37.0 0.0011124665085349241
37.0 0.0011772707899069435
37.0 0.001177469540903602
37.0 0.0011970768578329388
37.0 0.001227744768850907
37.0 0.0012279953313989811
37.0 0.0012995294893884282
38.0 0.0007670964371527719
38.0 0.0007769647914900216
38.0 0.0007819876401924652
38.0 0.0008462149120575244
38.0 0.0008978495170463734
38.0 0.0009299196404766084
38.0 0.0009451690660005197
38.0 0.000976391753058253
38.0 0.001017019061789116
38.0 0.0010565860877858962
38.0 0.0011124665085349241
38.0 0.0011217126795161464
38.0 0.001134559579218452
38.0 0.0012279953313989811
38.0 0.0012479587623218404
39.0 0.0006409230404464548
39.0 0.0006896692458289338
39.0 0.0008156729435369525
39.0 0.0008638315041898824
39.0 0.0009126959244590924
39.0 0.000949278806556905
39.0 0.0009947561939742478
39.0 0.0010161828439030473
39.0 0.0010426922686121334
39.0 0.001087918724325249
39.0 0.0011417724059244812
39.0 0.0012479587623218404
39.0 0.001253936663109122
39.0 0.001326985953492392
40.0 0.0005697589192575825
40.0 0.0006028269258901445
40.0 0.0006236204849626541
40.0 0.0006705045786321587
40.0 0.0006896692458289338
40.0 0.0007717278832725021
40.0 0.0007769647914900216
40.0 0.0007791204763744415
40.0 0.0008428126256482717
40.0 0.0008638315041898824
40.0 0.0009311423462408373
40.0 0.0009499391406657239
40.0 0.0009758718648426006
40.0 0.0012051233393315566
41.0 0.00045993395963840427
41.0 0.00047069697255151886
41.0 0.0006236204849626541
41.0 0.0006543140953799401
41.0 0.0006896692458289338
41.0 0.0007382723268909424
41.0 0.0007413633621773439
41.0 0.0007550588826560657
41.0 0.0008490392686234794
41.0 0.0008638315041898824
41.0 0.0009499391406657239
41.0 0.001005233212031771
41.0 0.0010565860877858962
41.0 0.001193627033574503
41.0 0.0012578565790275188
42.0 0.0003799668860954795
42.0 0.00044814832666824193
42.0 0.0006525524389005695
42.0 0.0006775043968147175
42.0 0.0007029055362905106
42.0 0.0007058484961395284
42.0 0.0007216654347512949
42.0 0.0007500865139579422
42.0 0.0007530806147042696
42.0 0.0007537877061196698
42.0 0.0007582068462912466
42.0 0.0008490392686234794
42.0 0.0008638315041898824
42.0 0.0009920285758074575
42.0 0.0012578565790275188
43.0 0.00037113389069806635
43.0 0.0003799668860954795
43.0 0.000425262312379267
43.0 0.00044814832666824193
43.0 0.0005267684575896046
43.0 0.000616949008434544
43.0 0.0006525524389005695
43.0 0.0006789855882016266
43.0 0.0006977841457777357
43.0 0.000723236156346331
43.0 0.0007500865139579422
43.0 0.0007537877061196698
43.0 0.0007589123055247365
43.0 0.0007655692183640291
43.0 0.0008219177075048532
43.0 0.0008346534266507251
43.0 0.0009591610521003466
43.0 0.0010128768441208663
44.0 0.00033522112193179807
44.0 0.00039731893439123064
44.0 0.000425262312379267
44.0 0.00044814832666824193
44.0 0.0005015715860038022
44.0 0.0005838274012023748
44.0 0.0006242082555406571
44.0 0.0006424437572633326
44.0 0.0006775043968147175
44.0 0.0007446510543150231
44.0 0.0007471560045291319
44.0 0.0007537877061196698
44.0 0.0007589480964496132
44.0 0.0007599266985551554
44.0 0.000841116799404493
44.0 0.0008778168693711114
45.0 0.000425262312379267
45.0 0.0005273333461341199
45.0 0.0005362112521720838
45.0 0.0005838274012023748
45.0 0.0005942930282568171
45.0 0.0006470669028939313
45.0 0.0006902862655762402
45.0 0.0006911132647393354
45.0 0.000691821764174973
45.0 0.0007084441693167838
45.0 0.0007220077112011492
45.0 0.0007446510543150231
45.0 0.0007471560045291319
45.0 0.0007637857560515181
45.0 0.0008535223216222808
46.0 0.0004763059379746639
46.0 0.0006320247000488016
46.0 0.0006651883714674803
46.0 0.0007585974043412619
46.0 0.000829204724747581
46.0 0.0008485433071210534
47.0 0.0004785990819385618
47.0 0.0005198427178002984
47.0 0.0005514152175416718
47.0 0.0005570461769344
47.0 0.0005942930282568171
47.0 0.0006044318063515825
47.0 0.0006131308629307597
47.0 0.0006564069007327864
47.0 0.0006651883714674803
47.0 0.0007721538064174673
47.0 0.0008879070209126825
48.0 0.0004562963772457502
48.0 0.0004785990819385618
48.0 0.0005198427178002984
48.0 0.0005640498768491792
48.0 0.0005644268931393261
48.0 0.0005646651139577081
48.0 0.0005673162423101099
48.0 0.000576596842140903
48.0 0.0005895268714128103
48.0 0.0006039809360128481
48.0 0.0006044318063515825
48.0 0.0006227235558021036
48.0 0.0006242082555406571
48.0 0.0007131610263366232
48.0 0.0007298149850667213
48.0 0.0009444538634184373
49.0 0.0003664942218564387
49.0 0.0003948287841342799
49.0 0.0004302273920990679
49.0 0.0004880668315004159
49.0 0.0004910994960757726
49.0 0.0005283766759650661
49.0 0.0005498849655419078
49.0 0.0005640498768491792
49.0 0.0005673162423101099
49.0 0.0006686161667344241
49.0 0.0008269626802249006
50.0 0.00032319935814649714
50.0 0.0003683828104321384
50.0 0.000376066596156748
50.0 0.0004592596902419743
50.0 0.0005023218824771598
50.0 0.0005318247517006402
50.0 0.0006039809360128481
50.0 0.0006197561685949073
50.0 0.0006424457910717603
50.0 0.0006467388542033099
50.0 0.000673163293025825
50.0 0.0009444538634184373
};
\addplot [semithick, color0, mark size=3.58351893845611, opacity=0.1, only marks] 
table {
1.0 0.001
};
\addplot [semithick, color0, mark size=1.791759469228055, opacity=0.1, only marks] 
table {
1.0 0.00048344093978007583
2.0 0.00048344093978007583
2.0 0.0009937466899386264
2.0 0.0011773973810961613
3.0 0.00048344093978007583
3.0 0.0008853166726104792
3.0 0.0009937466899386264
3.0 0.0018281493783056362
3.0 0.00787635269194737
4.0 0.0008853166726104792
4.0 0.0009736453164422528
5.0 0.00787635269194737
5.0 0.008781889689134558
6.0 0.0007931272046205693
6.0 0.0008853166726104792
6.0 0.0009265657279404006
6.0 0.0009736453164422528
6.0 0.008923816271475737
7.0 0.0007931272046205693
7.0 0.008014164443953524
7.0 0.008781889689134558
8.0 0.0009281112005052334
8.0 0.008781889689134558
8.0 0.008923816271475737
9.0 0.0009774183364130596
11.0 0.0011046205662347687
11.0 0.001211544027359025
12.0 0.0008212441214368042
12.0 0.0009072148020606427
13.0 0.0009397994692018497
13.0 0.0009777453667366796
13.0 0.0010348597373166023
13.0 0.0010720290603939
13.0 0.0012049353987602812
15.0 0.0008983532040235747
15.0 0.0010574610411818743
15.0 0.0011376249480354174
16.0 0.0006157991915885332
17.0 0.001222496171959703
18.0 0.0008983532040235747
18.0 0.0009153336725130123
18.0 0.0010708460084306439
20.0 0.0008983532040235747
20.0 0.0010574610411818743
21.0 0.001033973700608016
21.0 0.0010574610411818743
22.0 0.001033973700608016
22.0 0.0010574610411818743
22.0 0.0011217873948632935
23.0 0.0009666731420730366
23.0 0.0013659560316184997
24.0 0.0011217873948632935
27.0 0.0007354088146308797
27.0 0.001272221284403613
27.0 0.0012909649070802216
28.0 0.0010647216685715486
28.0 0.001453477206746658
30.0 0.0009931041308096579
30.0 0.000997968146927071
30.0 0.0013865935108078908
31.0 0.0009931041308096579
31.0 0.0012268674926634517
31.0 0.0012563952587863034
32.0 0.0010375381287446167
33.0 0.0010565860877858962
33.0 0.0010971141787349357
34.0 0.0011217126795161464
34.0 0.0012563952587863034
34.0 0.0012995294893884282
35.0 0.001101837282542011
35.0 0.001177469540903602
35.0 0.0012279953313989811
36.0 0.001177469540903602
36.0 0.0012995294893884282
37.0 0.0011217126795161464
37.0 0.0011395944486698075
39.0 0.0007670964371527719
39.0 0.0008462149120575244
39.0 0.0010565860877858962
40.0 0.0010161828439030473
40.0 0.0010565860877858962
40.0 0.0011256591460684876
41.0 0.0008789347280509609
42.0 0.0011256591460684876
43.0 0.0008490392686234794
44.0 0.0006525524389005695
44.0 0.0006789855882016266
45.0 0.0006242082555406571
46.0 0.0006242082555406571
46.0 0.000691821764174973
46.0 0.0007446510543150231
46.0 0.0007637857560515181
47.0 0.0006242082555406571
47.0 0.0006686161667344241
48.0 0.0005273333461341199
48.0 0.0008879070209126825
49.0 0.0007298149850667213
49.0 0.0009444538634184373
50.0 0.0005273333461341199
50.0 0.0005640498768491792
};
\addplot [semithick, color0, mark size=2.1972245773362196, opacity=0.1, only marks] 
table {
5.0 0.0009736453164422528
6.0 0.008781889689134558
9.0 0.0009281112005052334
10.0 0.001062105940867237
11.0 0.0009397994692018497
11.0 0.0010720290603939
12.0 0.0009397994692018497
12.0 0.0010720290603939
15.0 0.0012049353987602812
16.0 0.0008983532040235747
18.0 0.0010242100184749666
19.0 0.0008983532040235747
19.0 0.0010574610411818743
20.0 0.0010242100184749666
20.0 0.0010568026501619236
21.0 0.0009535945999513619
21.0 0.0010568026501619236
23.0 0.0009538616399559033
23.0 0.0011813703520567882
24.0 0.0011813703520567882
28.0 0.0007354088146308797
29.0 0.0011813703520567882
30.0 0.0007354088146308797
32.0 0.001150801546956867
32.0 0.0012563952587863034
34.0 0.0010565860877858962
35.0 0.0010565860877858962
35.0 0.0011217126795161464
35.0 0.0012995294893884282
36.0 0.0011217126795161464
36.0 0.0012279953313989811
37.0 0.0009299196404766084
41.0 0.0011256591460684876
42.0 0.0006896692458289338
45.0 0.000841116799404493
46.0 0.000425262312379267
46.0 0.0005273333461341199
50.0 0.0004302273920990679
};
\addplot [semithick, color0, mark size=2.4849066497880004, opacity=0.1, only marks] 
table {
5.0 0.0008853166726104792
5.0 0.001
7.0 0.0009121513794293555
8.0 0.0009397994692018497
9.0 0.0009397994692018497
14.0 0.0010348597373166023
18.0 0.0010574610411818743
21.0 0.0010242100184749666
23.0 0.0010574610411818743
28.0 0.0011813703520567882
29.0 0.0007354088146308797
34.0 0.0010971141787349357
};
\addplot [semithick, color0, mark size=2.70805020110221, opacity=0.1, only marks] 
table {
3.0 0.001
10.0 0.0009397994692018497
19.0 0.0010242100184749666
25.0 0.0011813703520567882
26.0 0.0011813703520567882
38.0 0.0011395944486698075
47.0 0.0005273333461341199
49.0 0.0005273333461341199
};
\addplot [semithick, color0, mark size=2.8903717578961645, opacity=0.1, only marks] 
table {
14.0 0.0012049353987602812
16.0 0.0010574610411818743
};
\addplot [semithick, color0, mark size=3.044522437723423, opacity=0.1, only marks] 
table {
2.0 0.001
};
\addplot [semithick, color0, mark size=3.1780538303479458, opacity=0.1, only marks] 
table {
17.0 0.0008983532040235747
};
\addplot [semithick, color0, mark size=3.295836866004329, opacity=0.1, only marks] 
table {
4.0 0.001
};
\addplot [semithick, orange]
table { 
0 0.0008983532040235747
52 0.0008983532040235747
};
\end{axis}
\end{tikzpicture}

%% file: images/ddpg_n_steps_per_fit.tex
\begin{tikzpicture}
\definecolor{color0}{rgb}{0.12156862745098,0.466666666666667,0.705882352941177}
\begin{axis}[
height=6cm,
width=6cm,
tick align=outside,
tick pos=left,
x grid style={white!69.0196078431373!black},
xtick style={color=black},
y grid style={white!69.0196078431373!black},
ytick style={color=black},
xlabel={Generation},
ylabel={Steps per fit},
xmin=0, 
xmax=52,
ymode=log]
\addplot [semithick, color0, mark size=3.4965075614664802, opacity=0.1, only marks] 
table {
2.0 10.0
10.0 7.0
17.0 4.0
23.0 6.0
26.0 5.0
45.0 3.0
};
\addplot [semithick, color0, mark size=1.0986122886681098, opacity=0.1, only marks] 
table {
1.0 12.0
1.0 1983.0
1.0 4557.0
1.0 5102.0
1.0 5658.0
1.0 7329.0
1.0 9525.0
2.0 8967.0
3.0 8.0
4.0 8.0
5.0 12.0
5.0 16.0
6.0 6.0
6.0 11.0
6.0 15.0
7.0 4.0
7.0 5.0
7.0 6.0
7.0 12.0
8.0 5.0
8.0 6.0
8.0 10.0
9.0 5.0
9.0 9.0
9.0 12.0
10.0 4.0
11.0 3.0
11.0 4.0
11.0 5.0
11.0 8.0
13.0 4.0
13.0 5.0
15.0 8.0
16.0 3.0
16.0 8.0
17.0 6.0
17.0 8.0
18.0 2.0
19.0 3.0
19.0 8.0
20.0 3.0
20.0 8.0
21.0 3.0
22.0 3.0
22.0 5.0
23.0 5.0
23.0 8.0
24.0 3.0
24.0 8.0
25.0 3.0
25.0 8.0
26.0 3.0
30.0 3.0
32.0 6.0
33.0 2.0
34.0 2.0
34.0 6.0
40.0 1.0
41.0 1.0
42.0 1.0
43.0 1.0
45.0 4.0
47.0 1.0
47.0 4.0
48.0 1.0
48.0 4.0
};
\addplot [semithick, color0, mark size=3.58351893845611, opacity=0.1, only marks] 
table {
27.0 5.0
34.0 4.0
44.0 3.0
};
\addplot [semithick, color0, mark size=1.791759469228055, opacity=0.1, only marks] 
table {
2.0 12.0
3.0 13.0
4.0 9.0
4.0 10.0
5.0 8.0
5.0 13.0
7.0 7.0
7.0 11.0
12.0 4.0
12.0 8.0
16.0 5.0
17.0 5.0
17.0 7.0
18.0 5.0
19.0 5.0
20.0 5.0
20.0 6.0
21.0 4.0
21.0 8.0
22.0 2.0
23.0 7.0
26.0 6.0
27.0 3.0
27.0 6.0
28.0 3.0
29.0 3.0
31.0 6.0
33.0 5.0
33.0 6.0
35.0 2.0
36.0 2.0
45.0 1.0
46.0 1.0
49.0 1.0
};
\addplot [semithick, color0, mark size=3.6635616461296463, opacity=0.1, only marks] 
table {
1.0 10.0
};
\addplot [semithick, color0, mark size=2.1972245773362196, opacity=0.1, only marks] 
table {
4.0 11.0
4.0 14.0
5.0 9.0
5.0 11.0
5.0 14.0
6.0 9.0
6.0 10.0
6.0 14.0
8.0 9.0
13.0 8.0
14.0 8.0
15.0 7.0
17.0 3.0
22.0 4.0
24.0 6.0
24.0 7.0
25.0 4.0
29.0 6.0
32.0 3.0
39.0 1.0
44.0 4.0
};
\addplot [semithick, color0, mark size=2.4849066497880004, opacity=0.1, only marks] 
table {
3.0 12.0
4.0 13.0
6.0 8.0
6.0 13.0
9.0 6.0
9.0 8.0
10.0 6.0
10.0 8.0
15.0 4.0
16.0 6.0
16.0 7.0
18.0 4.0
18.0 6.0
18.0 7.0
19.0 6.0
25.0 6.0
25.0 7.0
27.0 4.0
28.0 6.0
30.0 6.0
39.0 3.0
40.0 3.0
43.0 2.0
46.0 4.0
50.0 3.0
};
\addplot [semithick, color0, mark size=2.70805020110221, opacity=0.1, only marks] 
table {
3.0 10.0
4.0 12.0
5.0 10.0
14.0 5.0
15.0 6.0
18.0 3.0
19.0 4.0
20.0 4.0
23.0 4.0
24.0 5.0
37.0 2.0
38.0 3.0
42.0 2.0
42.0 4.0
43.0 4.0
44.0 2.0
50.0 1.0
};
\addplot [semithick, color0, mark size=2.8903717578961645, opacity=0.1, only marks] 
table {
2.0 11.0
7.0 8.0
7.0 9.0
8.0 8.0
14.0 6.0
14.0 7.0
22.0 6.0
26.0 4.0
28.0 4.0
29.0 4.0
33.0 3.0
34.0 3.0
39.0 2.0
40.0 2.0
41.0 3.0
41.0 4.0
45.0 2.0
};
\addplot [semithick, color0, mark size=3.044522437723423, opacity=0.1, only marks] 
table {
12.0 6.0
13.0 6.0
15.0 5.0
19.0 7.0
21.0 6.0
22.0 7.0
24.0 4.0
25.0 5.0
30.0 4.0
37.0 3.0
38.0 2.0
39.0 4.0
41.0 2.0
46.0 2.0
46.0 3.0
};
\addplot [semithick, color0, mark size=3.1780538303479458, opacity=0.1, only marks] 
table {
3.0 11.0
8.0 7.0
11.0 6.0
11.0 7.0
13.0 7.0
16.0 4.0
21.0 7.0
28.0 5.0
30.0 5.0
32.0 4.0
32.0 5.0
35.0 3.0
37.0 4.0
38.0 4.0
47.0 3.0
48.0 3.0
};
\addplot [semithick, color0, mark size=3.295836866004329, opacity=0.1, only marks] 
table {
9.0 7.0
12.0 7.0
20.0 7.0
29.0 5.0
31.0 4.0
31.0 5.0
33.0 4.0
36.0 3.0
36.0 4.0
40.0 4.0
42.0 3.0
49.0 2.0
49.0 3.0
};
\addplot [semithick, color0, mark size=3.4011973816621555, opacity=0.1, only marks] 
table {
35.0 4.0
43.0 3.0
47.0 2.0
48.0 2.0
50.0 2.0
};
\addplot [semithick, orange]
table { 
0 4
52 4
};
\end{axis}
\end{tikzpicture}

%% file: images/ddpg_n_steps.tex
\begin{tikzpicture}
\definecolor{color0}{rgb}{0.12156862745098,0.466666666666667,0.705882352941177}
\begin{axis}[
height=6cm,
width=6cm,
tick align=outside,
tick pos=left,
x grid style={white!69.0196078431373!black},
xtick style={color=black},
y grid style={white!69.0196078431373!black},
ytick style={color=black},
xlabel={Generation},
ylabel={Number of steps},
xmin=0, 
xmax=52]
\addplot [semithick, color0, mark size=1.0986122886681098, opacity=0.1, only marks] 
table {
1.0 1744.0
1.0 1837.0
1.0 7380.0
1.0 8921.0
1.0 9678.0
1.0 11260.0
1.0 14190.0
1.0 14333.0
2.0 1744.0
2.0 1892.0
2.0 2059.0
2.0 7697.0
2.0 7847.0
2.0 8151.0
2.0 8199.0
2.0 8921.0
2.0 9695.0
2.0 14190.0
2.0 14333.0
3.0 1617.0
3.0 6715.0
3.0 6754.0
3.0 6931.0
3.0 7310.0
3.0 7437.0
3.0 7500.0
3.0 7697.0
3.0 8111.0
3.0 8173.0
3.0 8456.0
3.0 8655.0
3.0 8921.0
3.0 9644.0
3.0 11085.0
3.0 11463.0
3.0 13950.0
3.0 14190.0
4.0 6754.0
4.0 6830.0
4.0 7065.0
4.0 7310.0
4.0 7552.0
4.0 8111.0
4.0 8323.0
4.0 8655.0
4.0 8979.0
4.0 9173.0
4.0 10240.0
4.0 11463.0
5.0 5742.0
5.0 5775.0
5.0 6041.0
5.0 6628.0
5.0 7065.0
5.0 7120.0
5.0 7230.0
5.0 7495.0
5.0 8111.0
5.0 8456.0
5.0 9173.0
6.0 5354.0
6.0 5404.0
6.0 6041.0
6.0 6282.0
6.0 6533.0
6.0 6628.0
6.0 6754.0
6.0 6809.0
6.0 7065.0
6.0 7230.0
6.0 7602.0
6.0 7843.0
6.0 8091.0
6.0 11736.0
7.0 5152.0
7.0 5404.0
7.0 6156.0
7.0 6235.0
7.0 6272.0
7.0 6529.0
7.0 6649.0
7.0 6689.0
7.0 6773.0
7.0 6809.0
7.0 6992.0
7.0 6993.0
7.0 7065.0
7.0 7602.0
7.0 7735.0
7.0 7737.0
7.0 8046.0
7.0 8373.0
7.0 8414.0
7.0 8955.0
8.0 5354.0
8.0 5404.0
8.0 6189.0
8.0 6441.0
8.0 6993.0
8.0 7065.0
8.0 7602.0
8.0 7808.0
8.0 9414.0
8.0 9684.0
8.0 9911.0
9.0 5152.0
9.0 5701.0
9.0 7584.0
9.0 7666.0
9.0 7762.0
9.0 8072.0
9.0 8422.0
9.0 8778.0
9.0 9038.0
9.0 9151.0
9.0 9414.0
10.0 5701.0
10.0 7534.0
10.0 7950.0
10.0 8072.0
10.0 8919.0
10.0 9620.0
11.0 7079.0
11.0 7385.0
11.0 7666.0
11.0 8175.0
11.0 8837.0
11.0 9151.0
11.0 9258.0
11.0 10431.0
11.0 10483.0
11.0 10571.0
12.0 7079.0
12.0 7225.0
12.0 7666.0
12.0 7925.0
12.0 8361.0
12.0 8837.0
12.0 8990.0
12.0 9038.0
12.0 9532.0
12.0 9534.0
12.0 10019.0
12.0 10311.0
12.0 11020.0
12.0 11256.0
13.0 7463.0
13.0 7925.0
13.0 8212.0
13.0 8837.0
13.0 8866.0
13.0 8906.0
13.0 9169.0
13.0 9414.0
13.0 9463.0
13.0 9534.0
13.0 9620.0
13.0 10128.0
13.0 12070.0
13.0 12278.0
14.0 7196.0
14.0 7227.0
14.0 7568.0
14.0 8158.0
14.0 8349.0
14.0 8361.0
14.0 8773.0
14.0 8906.0
14.0 9081.0
14.0 9180.0
14.0 9414.0
14.0 9534.0
14.0 9635.0
14.0 10683.0
14.0 12278.0
15.0 7568.0
15.0 7590.0
15.0 8116.0
15.0 8180.0
15.0 8458.0
15.0 8594.0
15.0 8773.0
15.0 8901.0
15.0 8946.0
15.0 8955.0
15.0 9219.0
15.0 9414.0
15.0 9659.0
15.0 9781.0
15.0 10602.0
15.0 10616.0
16.0 7124.0
16.0 7239.0
16.0 7405.0
16.0 7590.0
16.0 7744.0
16.0 7788.0
16.0 7911.0
16.0 7983.0
16.0 8158.0
16.0 8205.0
16.0 8955.0
16.0 9414.0
16.0 10213.0
16.0 10259.0
16.0 10602.0
16.0 11617.0
16.0 11991.0
17.0 6663.0
17.0 6711.0
17.0 6952.0
17.0 7674.0
17.0 7677.0
17.0 8034.0
17.0 8116.0
17.0 8129.0
17.0 8687.0
17.0 11190.0
18.0 6251.0
18.0 6711.0
18.0 7180.0
18.0 7189.0
18.0 7600.0
18.0 7790.0
18.0 7935.0
18.0 8034.0
18.0 8116.0
18.0 8129.0
18.0 8687.0
18.0 8803.0
19.0 6816.0
19.0 7488.0
19.0 7935.0
19.0 8116.0
19.0 8566.0
19.0 8683.0
19.0 9438.0
19.0 9711.0
19.0 11077.0
19.0 11219.0
19.0 11800.0
19.0 11850.0
19.0 12184.0
19.0 12201.0
20.0 7502.0
20.0 7600.0
20.0 7790.0
20.0 8052.0
20.0 8555.0
20.0 9980.0
20.0 10213.0
20.0 10243.0
20.0 11219.0
20.0 11602.0
20.0 11624.0
20.0 11754.0
20.0 12102.0
21.0 6681.0
21.0 8642.0
21.0 8738.0
21.0 10441.0
21.0 11219.0
21.0 11658.0
21.0 11764.0
21.0 12073.0
21.0 12102.0
21.0 12120.0
21.0 12562.0
21.0 12769.0
21.0 13018.0
21.0 13294.0
21.0 13700.0
22.0 6528.0
22.0 7688.0
22.0 7693.0
22.0 7758.0
22.0 9010.0
22.0 9276.0
22.0 9372.0
22.0 9483.0
22.0 10727.0
22.0 12769.0
22.0 13700.0
22.0 13717.0
22.0 14230.0
23.0 7752.0
23.0 8205.0
23.0 8492.0
23.0 8509.0
23.0 8824.0
23.0 9010.0
23.0 9075.0
23.0 9276.0
23.0 9291.0
23.0 9728.0
23.0 10233.0
23.0 14709.0
24.0 7608.0
24.0 7677.0
24.0 8257.0
24.0 8286.0
24.0 8509.0
24.0 8738.0
24.0 9010.0
24.0 9276.0
24.0 9291.0
24.0 9468.0
24.0 9535.0
24.0 10264.0
24.0 10435.0
24.0 10441.0
24.0 11871.0
24.0 13083.0
25.0 6493.0
25.0 7362.0
25.0 7608.0
25.0 7953.0
25.0 8436.0
25.0 8564.0
25.0 9010.0
25.0 9947.0
25.0 10372.0
25.0 10988.0
25.0 11246.0
25.0 11425.0
25.0 11871.0
25.0 12849.0
25.0 13083.0
25.0 13611.0
25.0 13700.0
25.0 13963.0
25.0 14100.0
25.0 14242.0
26.0 8955.0
26.0 10450.0
26.0 10859.0
26.0 11336.0
26.0 11611.0
26.0 11777.0
26.0 11871.0
26.0 11907.0
26.0 12034.0
26.0 12825.0
26.0 13307.0
26.0 13408.0
26.0 13941.0
26.0 14149.0
27.0 8454.0
27.0 8955.0
27.0 9147.0
27.0 9644.0
27.0 9947.0
27.0 9962.0
27.0 11236.0
27.0 11666.0
27.0 12472.0
27.0 12825.0
27.0 12849.0
27.0 13307.0
27.0 13611.0
27.0 13941.0
27.0 14289.0
27.0 14634.0
28.0 7584.0
28.0 8817.0
28.0 9090.0
28.0 9350.0
28.0 9855.0
28.0 9879.0
28.0 10823.0
28.0 11308.0
28.0 11523.0
28.0 12849.0
28.0 13307.0
28.0 14127.0
28.0 14248.0
29.0 8113.0
29.0 8197.0
29.0 8404.0
29.0 9350.0
29.0 9855.0
29.0 10396.0
29.0 10742.0
29.0 11189.0
29.0 11471.0
29.0 13838.0
30.0 6551.0
30.0 8113.0
30.0 8891.0
30.0 9053.0
30.0 9103.0
30.0 9557.0
30.0 9644.0
30.0 9855.0
30.0 9884.0
30.0 10396.0
30.0 10554.0
30.0 10659.0
30.0 12042.0
30.0 12736.0
30.0 13838.0
31.0 6772.0
31.0 7275.0
31.0 7613.0
31.0 8197.0
31.0 8343.0
31.0 9779.0
31.0 9884.0
31.0 10589.0
31.0 11634.0
31.0 12025.0
31.0 12736.0
31.0 13257.0
31.0 13489.0
31.0 13925.0
31.0 14248.0
32.0 6590.0
32.0 7275.0
32.0 7692.0
32.0 8197.0
32.0 8523.0
32.0 9395.0
32.0 9997.0
32.0 10826.0
32.0 11634.0
32.0 13239.0
32.0 13992.0
32.0 14248.0
33.0 6590.0
33.0 7613.0
33.0 7943.0
33.0 8197.0
33.0 8251.0
33.0 9055.0
33.0 9255.0
33.0 9884.0
33.0 10167.0
33.0 10826.0
33.0 10893.0
33.0 10943.0
33.0 13288.0
33.0 13925.0
33.0 14023.0
33.0 14104.0
34.0 6618.0
34.0 7454.0
34.0 7613.0
34.0 8222.0
34.0 8839.0
34.0 8974.0
34.0 9255.0
34.0 9884.0
34.0 9952.0
34.0 10167.0
34.0 10310.0
34.0 10943.0
34.0 11902.0
34.0 12500.0
34.0 13797.0
34.0 13925.0
35.0 6618.0
35.0 8015.0
35.0 8150.0
35.0 8215.0
35.0 8839.0
35.0 9233.0
35.0 9769.0
35.0 9884.0
35.0 9952.0
35.0 10167.0
35.0 10764.0
35.0 10846.0
35.0 10943.0
35.0 11378.0
35.0 11767.0
35.0 12115.0
36.0 6618.0
36.0 6965.0
36.0 7142.0
36.0 7684.0
36.0 8196.0
36.0 9769.0
36.0 10148.0
36.0 10167.0
36.0 10331.0
36.0 10885.0
36.0 11378.0
37.0 6680.0
37.0 6769.0
37.0 8905.0
37.0 9209.0
37.0 9769.0
37.0 9812.0
37.0 9923.0
37.0 10220.0
37.0 10333.0
37.0 10885.0
37.0 12153.0
37.0 12625.0
38.0 6680.0
38.0 9812.0
38.0 10936.0
38.0 11012.0
38.0 11945.0
38.0 12153.0
38.0 12625.0
38.0 13204.0
38.0 13895.0
39.0 6089.0
39.0 6680.0
39.0 7684.0
39.0 8496.0
39.0 8913.0
39.0 8948.0
39.0 11012.0
39.0 12008.0
39.0 13204.0
39.0 13914.0
40.0 5218.0
40.0 5286.0
40.0 7307.0
40.0 7684.0
40.0 8079.0
40.0 8707.0
40.0 8941.0
40.0 9812.0
40.0 11012.0
40.0 11743.0
41.0 6027.0
41.0 6311.0
41.0 6660.0
41.0 7478.0
41.0 7766.0
41.0 8079.0
41.0 8175.0
41.0 8196.0
41.0 8755.0
41.0 9404.0
41.0 9812.0
41.0 9891.0
41.0 11945.0
41.0 13032.0
42.0 6435.0
42.0 6696.0
42.0 6792.0
42.0 7256.0
42.0 7369.0
42.0 7469.0
42.0 7478.0
42.0 8079.0
42.0 8494.0
42.0 9404.0
42.0 9419.0
42.0 9812.0
42.0 9990.0
43.0 6092.0
43.0 6145.0
43.0 6527.0
43.0 7305.0
43.0 7369.0
43.0 7372.0
43.0 7452.0
43.0 7478.0
43.0 8494.0
43.0 8757.0
43.0 8771.0
43.0 8870.0
43.0 9404.0
43.0 9692.0
44.0 6145.0
44.0 6389.0
44.0 6794.0
44.0 6878.0
44.0 6938.0
44.0 7208.0
44.0 7372.0
44.0 7478.0
44.0 7839.0
44.0 8643.0
44.0 8870.0
44.0 9029.0
44.0 9058.0
44.0 10122.0
45.0 6017.0
45.0 6794.0
45.0 6938.0
45.0 7381.0
45.0 7440.0
45.0 7601.0
45.0 8079.0
45.0 9058.0
45.0 10122.0
46.0 5083.0
46.0 5387.0
46.0 6092.0
46.0 6391.0
46.0 7055.0
46.0 7381.0
46.0 7843.0
46.0 8487.0
46.0 8548.0
46.0 8664.0
46.0 8744.0
46.0 9058.0
46.0 10490.0
47.0 5339.0
47.0 5541.0
47.0 5660.0
47.0 6092.0
47.0 6241.0
47.0 6742.0
47.0 6794.0
47.0 6799.0
47.0 6821.0
47.0 6892.0
47.0 7372.0
47.0 7843.0
47.0 7906.0
47.0 8549.0
47.0 8744.0
48.0 5476.0
48.0 5541.0
48.0 5909.0
48.0 5988.0
48.0 6016.0
48.0 6279.0
48.0 6629.0
48.0 7205.0
48.0 7729.0
48.0 7843.0
48.0 7906.0
48.0 8744.0
49.0 5123.0
49.0 5174.0
49.0 6147.0
49.0 6162.0
49.0 6322.0
49.0 6391.0
49.0 6505.0
49.0 6534.0
49.0 6619.0
49.0 6731.0
49.0 6772.0
49.0 6794.0
49.0 6934.0
49.0 7175.0
49.0 7205.0
49.0 7372.0
49.0 8411.0
49.0 9083.0
50.0 5489.0
50.0 5784.0
50.0 5942.0
50.0 6054.0
50.0 6147.0
50.0 6152.0
50.0 6358.0
50.0 6505.0
50.0 6507.0
50.0 6635.0
50.0 6794.0
50.0 6886.0
50.0 6934.0
50.0 7191.0
};
\addplot [semithick, color0, mark size=3.58351893845611, opacity=0.1, only marks] 
table {
1.0 7500.0
};
\addplot [semithick, color0, mark size=1.791759469228055, opacity=0.1, only marks] 
table {
2.0 6179.0
2.0 7964.0
3.0 6179.0
4.0 7327.0
5.0 6754.0
5.0 6931.0
5.0 10853.0
6.0 7310.0
6.0 9173.0
6.0 10853.0
8.0 5152.0
8.0 6649.0
8.0 7737.0
9.0 6649.0
9.0 7737.0
9.0 8837.0
10.0 7666.0
10.0 8955.0
10.0 9151.0
10.0 9414.0
11.0 9038.0
11.0 9414.0
11.0 9620.0
13.0 7568.0
14.0 9463.0
15.0 8158.0
15.0 9180.0
17.0 7239.0
17.0 7911.0
17.0 10213.0
18.0 7677.0
19.0 10213.0
20.0 7677.0
20.0 9711.0
21.0 8555.0
22.0 7677.0
22.0 8738.0
23.0 13700.0
24.0 6407.0
24.0 13700.0
26.0 9947.0
26.0 12849.0
26.0 14242.0
27.0 9203.0
27.0 14242.0
28.0 9203.0
28.0 11666.0
29.0 9203.0
30.0 8197.0
31.0 8113.0
32.0 7613.0
32.0 9779.0
32.0 9884.0
32.0 13925.0
33.0 10846.0
33.0 12500.0
35.0 11088.0
35.0 11592.0
36.0 8839.0
36.0 11088.0
36.0 11592.0
37.0 7684.0
37.0 10167.0
37.0 10764.0
37.0 11592.0
38.0 7684.0
39.0 10764.0
39.0 11945.0
40.0 8196.0
40.0 8948.0
41.0 8707.0
42.0 6878.0
42.0 8175.0
43.0 6696.0
43.0 7440.0
43.0 8079.0
45.0 5281.0
45.0 7372.0
45.0 8549.0
46.0 6794.0
46.0 7372.0
47.0 5083.0
48.0 6391.0
49.0 7729.0
};
\addplot [semithick, color0, mark size=2.1972245773362196, opacity=0.1, only marks] 
table {
4.0 6715.0
4.0 8456.0
5.0 7310.0
8.0 8955.0
9.0 8955.0
10.0 8837.0
10.0 9038.0
12.0 8955.0
12.0 9414.0
14.0 8955.0
16.0 8458.0
18.0 7911.0
18.0 10213.0
20.0 11850.0
21.0 7677.0
22.0 10441.0
23.0 7677.0
23.0 8738.0
28.0 9644.0
29.0 14248.0
30.0 14248.0
31.0 13838.0
36.0 10764.0
38.0 8196.0
38.0 10764.0
38.0 11592.0
39.0 8196.0
39.0 9812.0
40.0 6089.0
40.0 11945.0
42.0 6089.0
44.0 6092.0
44.0 8079.0
46.0 8079.0
47.0 6391.0
48.0 6794.0
48.0 7372.0
};
\addplot [semithick, color0, mark size=2.4849066497880004, opacity=0.1, only marks] 
table {
11.0 8955.0
13.0 8955.0
17.0 7744.0
19.0 7677.0
34.0 10846.0
41.0 6089.0
};
\addplot [semithick, color0, mark size=2.70805020110221, opacity=0.1, only marks] 
table {
2.0 7500.0
29.0 9644.0
45.0 6092.0
50.0 7372.0
};
\addplot [semithick, orange]
table { 
0 9711
52 9711
};
\end{axis}
\end{tikzpicture}

%% file: chapters/07_conclusion.tex
\section{Conclusions and Limitations}
\label{sec:conclusions}

\textbf{Conclusions}~~This paper introduced the \libname{} framework for automating reinforcement learning by proposing two pipelines, one for the online setting and one for the offline setting. Moreover, we showcased the capabilities of such a framework by creating a Python library, and we tested its performance in both simulated and realistic settings.
While the proposed framework in its current formulation is flexible and allows adding customized \blocks, the complete democratization of RL is far from being achieved. First, the procedures to optimize the different \blocks revealed to be computationally demanding. Thus, adding tools to predict and control the amount of computational time required by a pipeline is of paramount importance to obtaining a flexible tool. Another interesting development, going in the opposite direction of what we have just mentioned, consists in including a \quotes{whole pipeline optimization} procedure, which \emph{jointly} optimize the entire learning process. This direction requires a preliminary development of less computationally demanding algorithms for each \block of the pipeline. Finally, we focused our attention on fully-observable, stationary, single agent, single-objective settings. Developing a more general pipeline to relax some or all the above assumptions would ease the application of RL algorithms in a more wide spectrum of real-world problems.

\textbf{Limitations}~~The goal of AutoRL is to bring RL closer to the non-expert user. This represents a source of opportunities and risks. On the one hand, making RL usable to a wide audience contributes to the \emph{democratization} of the field, overcoming the need for specific education and opening it to the large public. On the other hand, such an abstract approach tends to compromise the transparency of the learning process and traceability of the resulting model. Shadowing the underlying principles, AutoRL might pose the risk of misuse of RL approaches, leading to results not in line with expectations. Furthermore, AutoRL, even more than RL, requires huge amounts of data and computation that might represent a limit of the framework.

%% file: chapters/09_appendix.tex
{\centering{\huge Supplementary Material for the Paper: "\libname{}: A Framework for Automated Reinforcement Learning"}}

\section{Library}
\label{apx:implementation}

\libname{} is a Python library implementing the framework described in this paper. 
It contains all the automation capabilities described in the main paper. It also provides the implementation of specific stages for each phase of the two pipelines we introduced in Section~\ref{sec:pipelines}.

The RL algorithms present in the implementation are wrappers of those implemented in \textit{MushroomRL}~\citep{DBLP:journals/jmlr/DEramoTBRP21}. Moreover, we structured the library so that one has the option to implement wrappers for any other RL library, \eg Stable Baselines, RLLib, Tensorforce.

\textbf{Supported Units}~~
In Table~\ref{tab:units}, we list the currently implemented units for each \block. As mentioned before, this is a non-exhaustive list of the possible methods that can be included in the proposed framework, but only those which we used for experimental purposes. See Section~\ref{sec:components} for some suggestions about the methods which are appropriate for an extension for each \block.

\begin{table}[thp!]
    \caption{Supported units in the current ARLO implementation.}
    \label{tab:units}
    \centering
    \begin{tabular}{cc}
        \toprule
        \textbf{Stage} & \textbf{Implementations} \\
        \cmidrule(lr){1-2}
        \multirow{2}{*}{\texttt{Data Generation}} & Random Uniform Policy \\
        \cmidrule(lr){2-2}
        & MEPOL~\citep{DBLP:conf/aaai/MuttiPR21} \\ 
        \cmidrule(lr){1-2}
        \multirow{2}{*}{\texttt{Data Preparation}} & Mean Imputation  \\
        \cmidrule(lr){2-2}
        & 1-NN Imputation \\ 
        \cmidrule(lr){1-2}
        \multirow{2}{*}{\texttt{Feature Engineering}} &  Recursive Feature Selection \\
        \cmidrule(lr){2-2}
        & Forward Feature Selection via Mutual Information~\citep{DBLP:conf/ijcnn/BerahaMPTR19} \\
        \cmidrule(lr){2-2} &
        Nystroem Map Feature Generation
        \\ 
        \cmidrule(lr){1-2}
        \multirow{7}{*}{\texttt{Policy Generation}} & Fitted-Q Iteration~\citep[FQI,][]{DBLP:journals/jmlr/ErnstGW05} \\
        \cmidrule(lr){2-2}
        & Double Fitted-Q Iteration~\citep[DoubleFQI,][]{d2017estimating} \\
        \cmidrule(lr){2-2}
        & Least Squares Policy Iteration~\cite[LSPI,][]{lagoudakis2003least} \\
        \cmidrule(lr){2-2}
        & Deep Q-Network~\citep[DQN,][]{mnih2015human} \\
        \cmidrule(lr){2-2}
        & Proximal Policy Optimization~\citep[PPO,][]{schulman2017proximal} \\
        \cmidrule(lr){2-2}
        & Deep Deterministic Policy Gradient~\citep[DDPG,][]{DBLP:journals/corr/LillicrapHPHETS15} \\
        \cmidrule(lr){2-2}
        & Soft Actor Critic~\citep[SAC,][]{DBLP:conf/icml/HaarnojaZAL18}\\
        \cmidrule(lr){2-2}
        & GPOMDP~\citep{baxter2001infinite} \\
        \bottomrule
    \end{tabular}
\end{table}

\textbf{Used Libraries}~~
\libname{} requirements, in terms of libraries, are:
catboost (v1.0.3), gym (v0.19.0), joblib (v1.1.0), matplotlib (v3.5.0), mushroom\_rl (v1.7.0), numpy (v1.22.0), optuna (v2.10.0), plotly (v5.4.0), scikit\_learn (v1.0.2), scipy (v1.7.3), torch (v1.10.1), xgboost (v1.5.1).

\section{Details on the Experiments}
\label{apx:details_exp}
All the experiments were run on a Linux-based server with an \textit{AMD Ryzen} $9$ $5950X$ 16-Core Processor with 128GB DDR4 RAM running \textit{Python} $3.8.8$ on \textit{CentOS} $8.5.2111$.

\textbf{Hyper-parameter tuning}~~
The pseudo-code of the genetic tuner is detailed in Algorithm~\ref{algo:genetic_tuner}.
The hyper-parameter tuning of the genetic algorithms are run for $50$ generations, each one including $20$ agents. Throughout each generation, \textit{elitism} is performed, \ie the best agent of the generation is preserved, and the new generation is created via tournament selection. More specifically, we take the best agent, out of a subset of $3$ agents of the previous generation, and we repeat such an operation until $19$ agents are selected (as the remaining spot is reserved for the best performing agent in the previous generation).

\begin{algorithm}[t!]
\caption{Genetic Tuner} \label{algo:genetic_tuner}
\begin{algorithmic}[1]
    \STATE Randomly initialise first generation
    \FOR{$i \in [0, \ldots, n\_generations)$}
        \STATE Fit and evaluate each agent in the generation $i$ 
        \STATE Select the best agent and add it to the new generation $i+1$
        \FOR{$j \in [0, \ldots, n\_agents-1)$}
            \STATE Add best agent, out of a random subset of $3$, to the new generation $i+1$
        \ENDFOR
        \STATE Mutate the new generation $i+1$
    \ENDFOR
\end{algorithmic}
\end{algorithm} 

Each hyper-parameter is mutated with probability $0.5$ and two different types of mutation can take place:
\begin{itemize}[noitemsep]
    \item for \textit{categorical} hyper-parameters and for the ones having a discrete support, we sample from a uniform distribution over the possible values;
    \item for \textit{numerical}, \ie continuous domains, we sample hyper-parameters from a uniform distribution over $0.8$ and $1.2$ times of the current value of the hyper-parameter.
\end{itemize}
\libname{} implements the Genetic Algorithm presented above as well as the hyper-parameter tuning solutions from Optuna~\citep{optuna_2019}.

In all the experiments, we chose reasonable hyper-parameters configuration spaces so that they weren't neither too small, to avoid exploring a space that was too little and thus finding solutions quite far off, nor too large, to avoid increasing the total computational time (as some hyper-parameters have a great impact on the training time of the \texttt{Policy Generation} units).

\textbf{Environment}~~
Whenever an environment is used for the training of a RL algorithm, a deep copy of such an environment is provided to each agent in the generation, while in the case a dataset is used for the training of a RL algorithm, we provided each agent with a bootstrapped dataset coming from the original one. The environments currently available are presented in Table~\ref{tab:implemented_envs}.

\begin{table}[t!]
    \caption{Supported environments in the current \libname{} implementation.}
    \label{tab:implemented_envs}
    \centering
    \begin{tabular}{ccc}
        \toprule
        \textbf{Source} & \textbf{Type} & \textbf{Environment} \\
        \cmidrule(lr){1-3}
        \multirow{14}{*}{Gym} & \multirow{4}{*}{Classic Control} & Grid World  \\
        \cmidrule(lr){3-3} 
        && Mountain Car \\  
        \cmidrule(lr){3-3} 
        && Cart Pole \\ 
        \cmidrule(lr){2-3} 
        & \multirow{9}{*}{MuJoCo} & Inverted Pendulum \\ 
        \cmidrule(lr){3-3} 
        && Walker2d \\
        \cmidrule(lr){3-3} 
        && HalfCheetah \\
        \cmidrule(lr){3-3} 
        && Ant \\
        \cmidrule(lr){3-3} 
        && Hopper \\
        \cmidrule(lr){3-3} 
        && Humanoid \\
        \cmidrule(lr){3-3} 
        && Swimmer \\
        \cmidrule(lr){1-3} Other & Controller & LQG \\
        \bottomrule
    \end{tabular}
\end{table}

\textbf{Loss Function}~~
As loss function for guiding the tuning procedure we used the empirical expected return defined in Equation~\eqref{eq:discrew}. The specific loss functions used in the different experiments for policy evaluation are detailed in the following sections.

\subsection{Linear Quadratic Gaussian Regulator}
\label{sec:supplementary_lqg}

In this experiment, we consider a Linear Quadratic Gaussian (LQG) Regulator characterized as follows:
$$
A = \begin{bmatrix} 1 & 0 \\ 0 & 1\end{bmatrix},
\hspace*{0.5cm}
B = \begin{bmatrix} 1 & 0 & 0 \\ 0 & 0 & 1\end{bmatrix},
\hspace*{0.5cm}
Q = 0.7\cdot \begin{bmatrix} 1 & 0 \\ 0 & 1\end{bmatrix} ,
\hspace*{0.5cm}
R = 0.3\cdot \begin{bmatrix} 1 & 0 & 0 \\  0 & 1 & 0 \\ 0 & 0 & 1\end{bmatrix}.
$$
Along each dimension, and for each time step $t$, the action $a_t$ and the observation $s_t$, can take values in $[-3.5,3.5]$, while the discount factor and the time horizon were set to $\gamma = 0.9$, and $T = 15$, respectively.
We used a noise standard deviation as follows:
$$\sigma = 0.1 \cdot \begin{bmatrix}1 & 0 \\ 0 & 1\end{bmatrix}.$$
We perform three experiments using three different seeds: $2$, $42$, $2022$. We tune the hyper-parameters of SAC through the genetic algorithm described above, using as metric the empirical expected return.
For all the seeds we consider, we used the hyper-parameters configuration space reported in Table~\ref{table:sac_lqg_hp}. Notice that if only a single value is specified for its domain, it means that the hyper-parameter is considered fixed.
The three runs performed (one for each seed) took on average $44 \ (\pm17.2)$ hours each.

\begin{table}
    \caption{Hyper-parameters configuration space for the Linear Quadratic Gaussian Regulator experiment.}
    \label{table:sac_lqg_hp}
    \centering
    \begin{tabular}{c|cc}
        \toprule
        Hyper-parameter & Search Space \\
        \toprule
        Actor Learning Rate & \{$10^{-5}, 10^{-4}, 10^{-3}, 10^{-2}$\} \\
        Actor Network & One layer with 16 neurons and ReLU activation \\
        Critic Learning Rate & \{$10^{-5}, 10^{-4}, 10^{-3}, 10^{-2}$\} \\
        Critic Loss & MSE \\
        Critic Optimizer & Adam \\
        Critic Network & One layer with 16 neurons and ReLU activation \\
        Batch Size & \{8, 16, 32, 64, 128\} \\
        Initial Replay Size & \{10, 100, 300, 500, 1000, 5000\} \\
        Max Replay Size & \{3000, 10000, 30000, 100000\} \\
        Warmup Transitions & \{50, 100, 500\} \\
        Tau & 0.005 \\
        Alpha Learning Rate & \{$10^{-5}, 10^{-4}, 10^{-3}$\} \\
        Log Std Min & -20 \\
        Log Std Max & 3 \\
        N Epochs & [1, 30] \\
        N Episodes & [1, 1600] \\
        N Episodes Per Fit & [1, 500] \\
        \bottomrule
    \end{tabular}

\end{table}

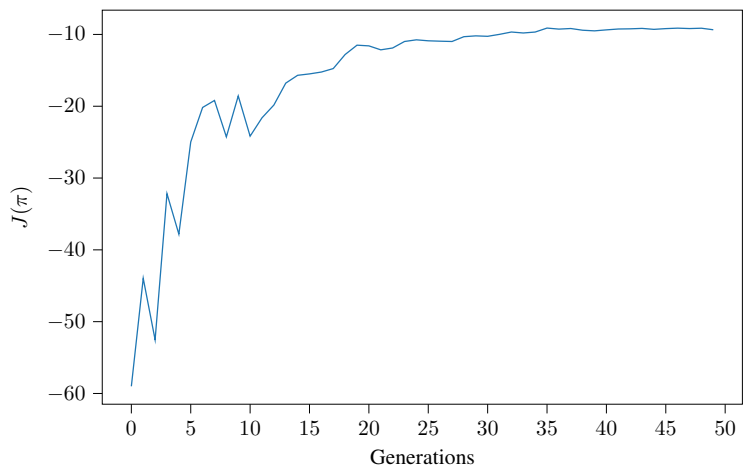
\begin{figure}[t!]
    \centering
        \centering
        \resizebox{0.6\linewidth}{!}{\input{images/performance_sac_supplementary}}
        \caption{SAC best agent performance for each generation.}
        \label{fig:sac_performance}
\end{figure}

In Figure~\ref{fig:sac_performance} we report the value of the performance over time of the best agent of each generation for this experiment.
It shows how the performances are almost constant in the last $\approx 25$ generations, meaning that the optimization procedure converged to a solution near to a local minimum point.
The scripts needed to run these three experiments (each corresponding to a different seed) are available at \url{https://github.com/arlo-lib/ARLO/tree/main/experiments/LQG}.

\subsection{HalfCheetah}
\label{sec:supplementary_halfcheetah}

In this second experiment we used the simulated environment of HalfCheetah to run experiment on the model generation \block. The MDP corresponding to this environment is assumed to have a discount factor and a time horizon of $\gamma = 1$, and $T = 1000$, respectively.
We perform three experiments using three different seeds: $2$, $42$, and $2022$. We tune the hyper-parameters of DDPG using the genetic algorithm described above, considering as metric the Average Reward.

\begin{table}
    \caption{Hyper-parameters configuration space for the HalfCheetah experiment.}
    \label{table:ddpg_halfcheetah_hp}
    \centering
    \begin{tabular}{c|cc}
        \toprule
        Hyper-parameter & Search Space \\
        \toprule
        Actor Learning Rate & [$10^{-5}, 10^{-2}$] \\
        Actor Network & Two layers with 128 neurons and ReLU activations \\
        Critic Learning Rate & [$10^{-5}, 10^{-2}$] \\
        Critic Loss & MSE \\
        Critic Optimizer & Adam \\
        Critic Network & Two layers with 128 neurons and ReLU activations \\
        Batch Size & [8, 256] \\
        Initial Replay Size & [1000, 20000] \\
        Max Replay Size & [10000, 1500000] \\
        Tau & 0.001 \\
        Policy delay & 1 \\
        Policy & OrnsteinUhlenbeckPolicy($\sigma$=0.2, $\theta$=0.15, dt=10$^{-2}$) \\
        N Epochs & [1, 50] \\
        N Steps & [1000, 15000] \\
        N Steps Per Fit & [1, 10000] \\
        \bottomrule
    \end{tabular}
\end{table}

For all three seeds, we consider the hyper-parameters configuration space reported in Table~\ref{table:ddpg_halfcheetah_hp}.
The three runs performed (one for each seed), took on average $124.7 \ (\pm8.8)$ hours each.
The scripts needed to run these three experiments (each corresponding to a different seed) are available at \url{https://github.com/arlo-lib/ARLO/tree/main/experiments/HalfCheetah-v3}.

\subsection{Dam}
\label{sec:dam_supplementary}
In this experiment, we consider the control of a water reservoir (dam) that models the dynamics of a real alpine lake, as described by~\citet{DBLP:conf/adprl/CastellettiGRS11}. The observation space is a continuous space with $31$ dimensions, each of which taking values in $\mathbb{R}^+$. This state space features represent the inflow values for the previous month. The action space is sampled to get a discrete space with $8$ actions, each one corresponding to a different amount of water released in a day.
The discount factor and the time horizon have been set to $\gamma=0.999$ and $T = 360$, respectively.
In this experiment, once we extract the dataset, we perform forward feature selection via mutual information, as described by~\citet{DBLP:conf/ijcnn/BerahaMPTR19}. Hyper-parameters search space for the tested tunable feature selection unit is reported in Table~\ref{table:fqi_dam_fs}.

\begin{table}[h!]
  \centering
  \captionof{table}{Hyper-parameters configuration space for the \texttt{Feature Engineering} stage of the Dam experiment.}
  \label{table:fqi_dam_fs}
  \small
    \begin{tabular}{c|cc}
        \toprule
        Hyper-parameter & Search Space \\
        \toprule
        K & $\{$1, 2, 3, 4, 5, 10, 20, 50$\}$ \\
        N Features & $\{$1, 2, $\ldots$, 31$\}$ \\
        \bottomrule
    \end{tabular}
\end{table}

\begin{table}[h!]
\centering
  \captionof{table}{Hyper-parameters used in the
  \texttt{Policy Generation} stage of the Dam experiment.}
  \label{table:fqi_dam_mg}
  \small
    \begin{tabular}{c|cc}
        \toprule
        Hyper-parameter & Value \\
        \toprule
        N Iterations & 60 \\
        N Estimators & 100 \\
        Criterion & MSE \\
        Min Samples Split & 10 \\
        \bottomrule
    \end{tabular}
\end{table}



Once feature selection is performed, we fit a \texttt{Policy Generation} unit, \ie FQI, using an Extremely Randomized Trees Regressor~\citep{DBLP:journals/ml/GeurtsEW06} with the hyper-parameters present in Table~\ref{table:fqi_dam_mg}.
The entire run took around $2$ hours.
The script needed to run this experiment is available at \url{https://github.com/arlo-lib/ARLO/tree/main/experiments/Dam/dam.py}.

\section{Additional Experiments}

In this section, we present two additional experiments, the former implementing an offline pipeline (Section~\ref{sec:apx_additional_offline}), and the latter an online pipeline (Section~\ref{sec:apx_additional_online}).

\subsection{Offline Pipeline}
\label{sec:apx_additional_offline}

The objective of this experiment to evaluate the capabilities of using the complete Offline Pipeline we defined in Section~\ref{sec:pipelines} in the Dam experiment described above (Section~\ref{sec:dam_supplementary}).

\begin{figure}[h!]
    \centering
    \centering
    \vspace{.5cm}
    \resizebox{\linewidth}{!}{\input{images/apx_offline_pipeline}}
    \caption{Types of the units adopted in the offline pipeline experiment.}
    \label{fig:apx_offline_pipeline}
\end{figure}
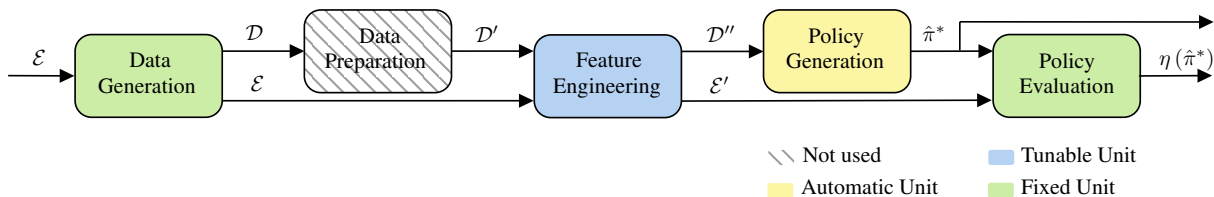

\begin{table}[h!]
  \centering
  \captionof{table}{Hyper-parameters configuration space of XGBoost.}
  \label{table:fqi_dam_tuning_hp_xgb}
  \small
  \begin{tabular}{c|cc}
        \toprule
        Hyper-parameter & Search Space \\
        \toprule
        N Iterations & [2, 60] \\
        N Estimators & [5, 250] \\
        Min Child Weight & [1, 100]  \\
        Subsample & [0.5, 1]  \\
        Learning Rate & [$10^{-3}$, 0.4] \\
        Max Depth & [4, 15]  \\
        \bottomrule
   \end{tabular}
\end{table}

\begin{table}[h!]   
  \centering
  \captionof{table}{Hyper-parameters configuration space of ExtraTrees.}
  \label{table:fqi_dam_tuning_hp_extra_trees}
  \small
  \begin{tabular}{c|cc}
        \toprule
        Hyper-parameter & Search Space \\
        \toprule
        N Iterations & [2, 60] \\
        N Estimators & [5, 250] \\
        Criterion & MSE \\
        Min Samples Split & [1, 50] \\
        \bottomrule
   \end{tabular}
\end{table}

The scheme of the used pipeline and the topology of the different \blocks is presented in Figure~\ref{fig:apx_offline_pipeline}.\textbf{}
We use a fixed unit to generate the data, and a tunable unit to perform the \texttt{Feature Engineering} stage, as the ones we used for the dam experiment in Section~\ref{sec:experiments}.
Subsequently, we use an automatic \texttt{Policy Generation} unit composed two tunable units using different versions of Fitted-Q Iteration, one with XGBoost as regressor, and the other with Extremely Randomized Trees. The two hyper-parameters configurations spaces are those presented in Table~\ref{table:fqi_dam_tuning_hp_xgb} and Table~\ref{table:fqi_dam_tuning_hp_extra_trees}, respectively.

\begin{table}[t!]
    \caption{Results obtained for the additional experiment over the full offline pipeline.}
    \label{table:fqi_dam_tuning_res}
    \centering
    \begin{tabular}{c|cc}
        \toprule
        Method & Empirical Expected Return \\
        \toprule
        Baseline & $-1224.67 \ (124.41)$ \\
        Tuned Configuration & $-1047.97 \ (213.37)$ \\
        \bottomrule
    \end{tabular}
\end{table}

In Table~\ref{table:fqi_dam_tuning_res} we report the empirical expected return over $10$ episodes (standard deviation in brackets).
Even though we have improved over the previously obtained result by about $20\%$, we point out that obtaining a statistically significant result would require a huge computational effort, indeed the entire run took around $16$ hours. We leave to future experiments the test on a larger number of samples to assess the statistical significance of this result.
The script needed to run this experiment is available at \url{https://github.com/arlo-lib/ARLO/blob/main/experiments/Dam/hp_tuning_fqi_dam.py}.

\subsection{Online Pipeline}
\label{sec:apx_additional_online}

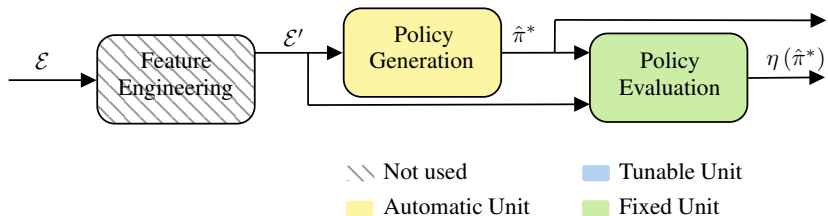
\begin{figure}[h!]
    \centering
    \centering
    \resizebox{0.7\linewidth}{!}{\input{images/apx_online_pipeline}}
    \caption{Types of the units adopted in the online pipeline experiment.}
    \label{fig:apx_online_pipeline}
\end{figure}

Differently from the previous experiments we run on online pipelines, in which we used a single tunable unit for the \texttt{Policy Generation} stage, in what follows, we apply an automatic unit to this \block, to tune the hyper-parameters of different algorithms.
To test this out we run an automatic \texttt{Policy Generation} unit for the Linear Quadratic Gaussian Regulator, in which we tune both Soft Actor Critic~\citep[SAC,][]{DBLP:conf/icml/HaarnojaZAL18} and Proximal Policy Optimization~\citep[PPO,][]{schulman2017proximal}.
To test the performance we consider the Discounted Reward (defined in Equation~\ref{eq:discrew}) in \texttt{Policy Evaluation} stage. Figure~\ref{fig:apx_online_pipeline} represents the pipeline and the topology of the units used in the experiment.

\begin{table}[h!]
    \caption{Performance obtained using the automatic \texttt{Policy Generation} unit on the Linear Quadratic Gaussian Regulator.}
    \label{table:table_lqg_ppo_sac}
    \centering
    \begin{tabular}{c|cc}
        \toprule
        Method & Empirical Expected Return \\
        \midrule
        \citet{van1981generalized} & $-7.2 \ (4.9)$ \\
        \toprule
        Best SAC Tuned Configuration & $-8.2$ \ ($4.7$)\\
        Best Automatic Block Tuned Configuration & $-7.4$ \ ($5.0$) \\
        \bottomrule
    \end{tabular}
\end{table}

Table~\ref{table:table_lqg_ppo_sac} shows the results in terms of discounted reward. By tuning the hyper-parameters of two different \texttt{Policy Generation} tunable units the pipeline further improved the results presented in Section~\ref{sec:experiments}, reaching a level of performance in line with the one of the optimal solution by~\citet{van1981generalized}.

%% file: images/performance_sac_supplementary.tex
\begin{tikzpicture}

\definecolor{darkgray176}{RGB}{176,176,176}
\definecolor{steelblue31119180}{RGB}{31,119,180}

\begin{axis}[
height=8cm,
width=12cm,
tick align=outside,
tick pos=left,
x grid style={darkgray176},
xmin=-2.45, xmax=51.45,
xtick style={color=black},
y grid style={darkgray176},
ymin=-61.4957231853865, ymax=-6.62020789537505,
ytick style={color=black},
xlabel={Generations},
ylabel={$J(\pi)$}
]
\addplot [semithick, steelblue31119180]
table {%
0 -59.0013815812951
1 -43.9666955255919
2 -52.5277642686108
3 -32.1793153104166
4 -37.8148460300134
5 -24.9778702741573
6 -20.1701157001408
7 -19.1945778120519
8 -24.2675567475447
9 -18.5684670631068
10 -24.1685072058796
11 -21.6134537289139
12 -19.8336793384131
13 -16.7831037826655
14 -15.6926558968665
15 -15.4961785933127
16 -15.2378721526786
17 -14.7540959510963
18 -12.7972403752459
19 -11.4968125084208
20 -11.5933053226731
21 -12.1449807266396
22 -11.8891690425168
23 -10.9830105991925
24 -10.7513319507179
25 -10.8874457770093
26 -10.9419756665368
27 -10.9835756875159
28 -10.3156685856611
29 -10.2001881593947
30 -10.2623815297516
31 -9.98878690162022
32 -9.66448911114529
33 -9.79855037435131
34 -9.66769644336881
35 -9.11454949946648
36 -9.25933283157236
37 -9.17569437699644
38 -9.41503834590906
39 -9.50191440403006
40 -9.36517394141991
41 -9.24489204309758
42 -9.22024326393059
43 -9.16006293963497
44 -9.2937504235072
45 -9.18995442197441
46 -9.12594674476139
47 -9.1824820146247
48 -9.13526503488031
49 -9.35027383852138
};
\end{axis}

\end{tikzpicture}

%% file: images/apx_offline_pipeline.tex
 
\tikzset{
pattern size/.store in=\mcSize, 
pattern size = 5pt,
pattern thickness/.store in=\mcThickness, 
pattern thickness = 0.3pt,
pattern radius/.store in=\mcRadius, 
pattern radius = 1pt}
\makeatletter
\pgfutil@ifundefined{pgf@pattern@name@_abfghr2e7}{
\pgfdeclarepatternformonly[\mcThickness,\mcSize]{_abfghr2e7}
{\pgfqpoint{0pt}{-\mcThickness}}
{\pgfpoint{\mcSize}{\mcSize}}
{\pgfpoint{\mcSize}{\mcSize}}
{
\pgfsetcolor{\tikz@pattern@color}
\pgfsetlinewidth{\mcThickness}
\pgfpathmoveto{\pgfqpoint{0pt}{\mcSize}}
\pgfpathlineto{\pgfpoint{\mcSize+\mcThickness}{-\mcThickness}}
\pgfusepath{stroke}
}}
\makeatother

 
\tikzset{
pattern size/.store in=\mcSize, 
pattern size = 5pt,
pattern thickness/.store in=\mcThickness, 
pattern thickness = 0.3pt,
pattern radius/.store in=\mcRadius, 
pattern radius = 1pt}
\makeatletter
\pgfutil@ifundefined{pgf@pattern@name@_nd501v835}{
\pgfdeclarepatternformonly[\mcThickness,\mcSize]{_nd501v835}
{\pgfqpoint{0pt}{-\mcThickness}}
{\pgfpoint{\mcSize}{\mcSize}}
{\pgfpoint{\mcSize}{\mcSize}}
{
\pgfsetcolor{\tikz@pattern@color}
\pgfsetlinewidth{\mcThickness}
\pgfpathmoveto{\pgfqpoint{0pt}{\mcSize}}
\pgfpathlineto{\pgfpoint{\mcSize+\mcThickness}{-\mcThickness}}
\pgfusepath{stroke}
}}
\makeatother
\tikzset{every picture/.style={line width=0.75pt}} 

\begin{tikzpicture}[x=0.75pt,y=0.75pt,yscale=-1,xscale=1]

\draw    (-20.32,155.15) -- (17.02,155.3) ;
\draw [shift={(20.02,155.31)}, rotate = 180.22] [fill={rgb, 255:red, 0; green, 0; blue, 0 }  ][line width=0.08]  [draw opacity=0] (8.93,-4.29) -- (0,0) -- (8.93,4.29) -- cycle    ;
\draw  [fill={rgb, 255:red, 74; green, 144; blue, 226 }  ,fill opacity=0.4 ] (300.05,140.41) .. controls (300.05,134.84) and (304.56,130.33) .. (310.13,130.33) -- (379.57,130.33) .. controls (385.13,130.33) and (389.65,134.84) .. (389.65,140.41) -- (389.65,170.65) .. controls (389.65,176.21) and (385.13,180.73) .. (379.57,180.73) -- (310.13,180.73) .. controls (304.56,180.73) and (300.05,176.21) .. (300.05,170.65) -- cycle ;
\draw    (668.21,155.34) -- (709.84,155.14) ;
\draw [shift={(712.84,155.13)}, rotate = 179.72] [fill={rgb, 255:red, 0; green, 0; blue, 0 }  ][line width=0.08]  [draw opacity=0] (8.93,-4.29) -- (0,0) -- (8.93,4.29) -- cycle    ;
\draw  [pattern=_abfghr2e7,pattern size=6pt,pattern thickness=0.75pt,pattern radius=0pt, pattern color={rgb, 255:red, 155; green, 155; blue, 155}] (160.05,125.21) .. controls (160.05,119.64) and (164.56,115.13) .. (170.13,115.13) -- (239.57,115.13) .. controls (245.13,115.13) and (249.65,119.64) .. (249.65,125.21) -- (249.65,155.45) .. controls (249.65,161.01) and (245.13,165.52) .. (239.57,165.52) -- (170.13,165.52) .. controls (164.56,165.52) and (160.05,161.01) .. (160.05,155.45) -- cycle ;
\draw  [fill={rgb, 255:red, 126; green, 211; blue, 33 }  ,fill opacity=0.4 ] (20.45,140.21) .. controls (20.45,134.64) and (24.96,130.13) .. (30.53,130.13) -- (99.97,130.13) .. controls (105.53,130.13) and (110.05,134.64) .. (110.05,140.21) -- (110.05,170.45) .. controls (110.05,176.01) and (105.53,180.53) .. (99.97,180.53) -- (30.53,180.53) .. controls (24.96,180.53) and (20.45,176.01) .. (20.45,170.45) -- cycle ;
\draw    (110.05,140.21) -- (157.05,140.21) ;
\draw [shift={(160.05,140.21)}, rotate = 180] [fill={rgb, 255:red, 0; green, 0; blue, 0 }  ][line width=0.08]  [draw opacity=0] (8.93,-4.29) -- (0,0) -- (8.93,4.29) -- cycle    ;
\draw    (249.65,140.21) -- (296.65,140.2) ;
\draw [shift={(299.65,140.2)}, rotate = 180] [fill={rgb, 255:red, 0; green, 0; blue, 0 }  ][line width=0.08]  [draw opacity=0] (8.93,-4.29) -- (0,0) -- (8.93,4.29) -- cycle    ;
\draw    (110.05,170.45) -- (296.65,170.44) ;
\draw [shift={(299.65,170.44)}, rotate = 180] [fill={rgb, 255:red, 0; green, 0; blue, 0 }  ][line width=0.08]  [draw opacity=0] (8.93,-4.29) -- (0,0) -- (8.93,4.29) -- cycle    ;
\draw  [fill={rgb, 255:red, 248; green, 231; blue, 28 }  ,fill opacity=0.4 ] (439.65,125.2) .. controls (439.65,119.64) and (444.16,115.12) .. (449.73,115.12) -- (519.17,115.12) .. controls (524.73,115.12) and (529.25,119.64) .. (529.25,125.2) -- (529.25,155.44) .. controls (529.25,161.01) and (524.73,165.52) .. (519.17,165.52) -- (449.73,165.52) .. controls (444.16,165.52) and (439.65,161.01) .. (439.65,155.44) -- cycle ;
\draw    (389.65,140.41) -- (436.65,140.4) ;
\draw [shift={(439.65,140.4)}, rotate = 180] [fill={rgb, 255:red, 0; green, 0; blue, 0 }  ][line width=0.08]  [draw opacity=0] (8.93,-4.29) -- (0,0) -- (8.93,4.29) -- cycle    ;
\draw    (389.65,170.65) -- (575.64,170.13) ;
\draw [shift={(578.64,170.13)}, rotate = 179.84] [fill={rgb, 255:red, 0; green, 0; blue, 0 }  ][line width=0.08]  [draw opacity=0] (8.93,-4.29) -- (0,0) -- (8.93,4.29) -- cycle    ;
\draw  [fill={rgb, 255:red, 126; green, 211; blue, 33 }  ,fill opacity=0.4 ] (579.45,140.2) .. controls (579.45,134.64) and (583.96,130.12) .. (589.53,130.12) -- (658.97,130.12) .. controls (664.53,130.12) and (669.05,134.64) .. (669.05,140.2) -- (669.05,170.44) .. controls (669.05,176.01) and (664.53,180.52) .. (658.97,180.52) -- (589.53,180.52) .. controls (583.96,180.52) and (579.45,176.01) .. (579.45,170.44) -- cycle ;
\draw    (529.15,140.31) -- (576.45,140.21) ;
\draw [shift={(579.45,140.2)}, rotate = 179.88] [fill={rgb, 255:red, 0; green, 0; blue, 0 }  ][line width=0.08]  [draw opacity=0] (8.93,-4.29) -- (0,0) -- (8.93,4.29) -- cycle    ;
\draw    (559.19,122.35) -- (710.75,122.46) ;
\draw [shift={(713.75,122.47)}, rotate = 180.04] [fill={rgb, 255:red, 0; green, 0; blue, 0 }  ][line width=0.08]  [draw opacity=0] (8.93,-4.29) -- (0,0) -- (8.93,4.29) -- cycle    ;
\draw    (559.19,122.35) -- (559.16,139.7) ;
\draw  [color={rgb, 255:red, 255; green, 255; blue, 255 }  ,draw opacity=1 ][fill={rgb, 255:red, 248; green, 231; blue, 28 }  ,fill opacity=0.4 ] (441.54,222.18) .. controls (441.54,221.04) and (442.46,220.13) .. (443.59,220.13) -- (456.24,220.13) .. controls (457.37,220.13) and (458.29,221.04) .. (458.29,222.18) -- (458.29,228.33) .. controls (458.29,229.46) and (457.37,230.38) .. (456.24,230.38) -- (443.59,230.38) .. controls (442.46,230.38) and (441.54,229.46) .. (441.54,228.33) -- cycle ;
\draw  [color={rgb, 255:red, 255; green, 255; blue, 255 }  ,draw opacity=1 ][fill={rgb, 255:red, 74; green, 144; blue, 226 }  ,fill opacity=0.4 ] (576.04,201.93) .. controls (576.04,200.79) and (576.96,199.88) .. (578.09,199.88) -- (590.74,199.88) .. controls (591.87,199.88) and (592.79,200.79) .. (592.79,201.93) -- (592.79,208.08) .. controls (592.79,209.21) and (591.87,210.13) .. (590.74,210.13) -- (578.09,210.13) .. controls (576.96,210.13) and (576.04,209.21) .. (576.04,208.08) -- cycle ;
\draw  [color={rgb, 255:red, 255; green, 255; blue, 255 }  ,draw opacity=1 ][fill={rgb, 255:red, 126; green, 211; blue, 33 }  ,fill opacity=0.4 ] (575.79,221.93) .. controls (575.79,220.79) and (576.71,219.88) .. (577.84,219.88) -- (590.49,219.88) .. controls (591.62,219.88) and (592.54,220.79) .. (592.54,221.93) -- (592.54,228.08) .. controls (592.54,229.21) and (591.62,230.13) .. (590.49,230.13) -- (577.84,230.13) .. controls (576.71,230.13) and (575.79,229.21) .. (575.79,228.08) -- cycle ;
\draw  [color={rgb, 255:red, 255; green, 255; blue, 255 }  ,draw opacity=1 ][pattern=_nd501v835,pattern size=6pt,pattern thickness=0.75pt,pattern radius=0pt, pattern color={rgb, 255:red, 155; green, 155; blue, 155}] (442.04,201.93) .. controls (442.04,200.79) and (442.96,199.88) .. (444.09,199.88) -- (456.74,199.88) .. controls (457.87,199.88) and (458.79,200.79) .. (458.79,201.93) -- (458.79,208.08) .. controls (458.79,209.21) and (457.87,210.13) .. (456.74,210.13) -- (444.09,210.13) .. controls (442.96,210.13) and (442.04,209.21) .. (442.04,208.08) -- cycle ;

\draw (-7,138) node [anchor=north west][inner sep=0.75pt]    {$\mathcal{E}$};
\draw (33,155) node [anchor=north west][inner sep=0.75pt]   [align=left] {Generation};
\draw (52,140) node [anchor=north west][inner sep=0.75pt]   [align=left] {Data};
\draw (122.96,123) node [anchor=north west][inner sep=0.75pt]    {$\mathcal{D}$};
\draw (171,140) node [anchor=north west][inner sep=0.75pt]   [align=left] {Preparation};
\draw (192,125) node [anchor=north west][inner sep=0.75pt]   [align=left] {Data};
\draw (126,154) node [anchor=north west][inner sep=0.75pt]    {$\mathcal{E}$};
\draw (262.73,123) node [anchor=north west][inner sep=0.75pt]    {$\mathcal{D} '$};
\draw (310,155) node [anchor=north west][inner sep=0.75pt]   [align=left] {Engineering};
\draw (324,140) node [anchor=north west][inner sep=0.75pt]   [align=left] {Feature};
\draw (405.5,154) node [anchor=north west][inner sep=0.75pt]    {$\mathcal{E} '$};
\draw (403.7,123) node [anchor=north west][inner sep=0.75pt]    {$\mathcal{D} ''$};
\draw (465.48,125) node [anchor=north west][inner sep=0.75pt]   [align=left] {Policy};
\draw (606,140) node [anchor=north west][inner sep=0.75pt]   [align=left] {Policy};
\draw (535,123) node [anchor=north west][inner sep=0.75pt]    {$\hat{\pi }^{*}$};
\draw (594,155) node [anchor=north west][inner sep=0.75pt]   [align=left] {Evaluation};
\draw (452,140) node [anchor=north west][inner sep=0.75pt]   [align=left] {Generation};
\draw (679,137) node [anchor=north west][inner sep=0.75pt]    {$\eta \left(\hat{\pi }^{*}\right)$};
\draw (461,217) node [anchor=north west][inner sep=0.75pt]   [align=left] {Automatic Unit};
\draw (595,197) node [anchor=north west][inner sep=0.75pt]   [align=left] {Tunable Unit};
\draw (595,217) node [anchor=north west][inner sep=0.75pt]   [align=left] {Fixed Unit};
\draw (462,197) node [anchor=north west][inner sep=0.75pt]   [align=left] {Not used};

\end{tikzpicture}

%% file: images/apx_online_pipeline.tex
 
\tikzset{
pattern size/.store in=\mcSize, 
pattern size = 5pt,
pattern thickness/.store in=\mcThickness, 
pattern thickness = 0.3pt,
pattern radius/.store in=\mcRadius, 
pattern radius = 1pt}
\makeatletter
\pgfutil@ifundefined{pgf@pattern@name@_har2ykjgj}{
\pgfdeclarepatternformonly[\mcThickness,\mcSize]{_har2ykjgj}
{\pgfqpoint{0pt}{-\mcThickness}}
{\pgfpoint{\mcSize}{\mcSize}}
{\pgfpoint{\mcSize}{\mcSize}}
{
\pgfsetcolor{\tikz@pattern@color}
\pgfsetlinewidth{\mcThickness}
\pgfpathmoveto{\pgfqpoint{0pt}{\mcSize}}
\pgfpathlineto{\pgfpoint{\mcSize+\mcThickness}{-\mcThickness}}
\pgfusepath{stroke}
}}
\makeatother

 
\tikzset{
pattern size/.store in=\mcSize, 
pattern size = 5pt,
pattern thickness/.store in=\mcThickness, 
pattern thickness = 0.3pt,
pattern radius/.store in=\mcRadius, 
pattern radius = 1pt}
\makeatletter
\pgfutil@ifundefined{pgf@pattern@name@_bpeyddhwz}{
\pgfdeclarepatternformonly[\mcThickness,\mcSize]{_bpeyddhwz}
{\pgfqpoint{0pt}{-\mcThickness}}
{\pgfpoint{\mcSize}{\mcSize}}
{\pgfpoint{\mcSize}{\mcSize}}
{
\pgfsetcolor{\tikz@pattern@color}
\pgfsetlinewidth{\mcThickness}
\pgfpathmoveto{\pgfqpoint{0pt}{\mcSize}}
\pgfpathlineto{\pgfpoint{\mcSize+\mcThickness}{-\mcThickness}}
\pgfusepath{stroke}
}}
\makeatother
\tikzset{every picture/.style={line width=0.75pt}} 

\begin{tikzpicture}[x=0.75pt,y=0.75pt,yscale=-1,xscale=1]

\draw  [pattern=_har2ykjgj,pattern size=6pt,pattern thickness=0.75pt,pattern radius=0pt, pattern color={rgb, 255:red, 155; green, 155; blue, 155}] (300.66,291.66) .. controls (300.66,286.1) and (305.18,281.58) .. (310.74,281.58) -- (380.18,281.58) .. controls (385.75,281.58) and (390.26,286.1) .. (390.26,291.66) -- (390.26,321.9) .. controls (390.26,327.47) and (385.75,331.98) .. (380.18,331.98) -- (310.74,331.98) .. controls (305.18,331.98) and (300.66,327.47) .. (300.66,321.9) -- cycle ;
\draw    (670.92,305.99) -- (711.7,305.64) ;
\draw [shift={(714.7,305.61)}, rotate = 179.5] [fill={rgb, 255:red, 0; green, 0; blue, 0 }  ][line width=0.08]  [draw opacity=0] (8.93,-4.29) -- (0,0) -- (8.93,4.29) -- cycle    ;
\draw    (250.26,307.21) -- (297.26,307.21) ;
\draw [shift={(300.26,307.21)}, rotate = 180] [fill={rgb, 255:red, 0; green, 0; blue, 0 }  ][line width=0.08]  [draw opacity=0] (8.93,-4.29) -- (0,0) -- (8.93,4.29) -- cycle    ;
\draw    (390.26,291.66) -- (437.26,291.66) ;
\draw [shift={(440.26,291.66)}, rotate = 180] [fill={rgb, 255:red, 0; green, 0; blue, 0 }  ][line width=0.08]  [draw opacity=0] (8.93,-4.29) -- (0,0) -- (8.93,4.29) -- cycle    ;
\draw    (420.6,321.41) -- (577.47,320.95) -- (577.91,320.95) ;
\draw [shift={(580.91,320.95)}, rotate = 180] [fill={rgb, 255:red, 0; green, 0; blue, 0 }  ][line width=0.08]  [draw opacity=0] (8.93,-4.29) -- (0,0) -- (8.93,4.29) -- cycle    ;
\draw    (530.62,291.66) -- (577.92,291.57) ;
\draw [shift={(580.92,291.56)}, rotate = 179.88] [fill={rgb, 255:red, 0; green, 0; blue, 0 }  ][line width=0.08]  [draw opacity=0] (8.93,-4.29) -- (0,0) -- (8.93,4.29) -- cycle    ;
\draw    (561.03,272.91) -- (712.41,273.19) ;
\draw [shift={(715.41,273.19)}, rotate = 180.1] [fill={rgb, 255:red, 0; green, 0; blue, 0 }  ][line width=0.08]  [draw opacity=0] (8.93,-4.29) -- (0,0) -- (8.93,4.29) -- cycle    ;
\draw    (561.03,272.91) -- (560.9,291.79) ;
\draw    (420.63,291.74) -- (420.6,321.41) ;
\draw [color={rgb, 255:red, 255; green, 255; blue, 255 }  ,draw opacity=1 ]   (681.8,321.41) -- (723.8,320.74) ;
\draw  [fill={rgb, 255:red, 248; green, 231; blue, 28 }  ,fill opacity=0.4 ] (440.85,276.41) .. controls (440.85,270.84) and (445.36,266.33) .. (450.93,266.33) -- (520.37,266.33) .. controls (525.93,266.33) and (530.45,270.84) .. (530.45,276.41) -- (530.45,306.65) .. controls (530.45,312.21) and (525.93,316.73) .. (520.37,316.73) -- (450.93,316.73) .. controls (445.36,316.73) and (440.85,312.21) .. (440.85,306.65) -- cycle ;
\draw  [fill={rgb, 255:red, 126; green, 211; blue, 33 }  ,fill opacity=0.4 ] (580.91,290.71) .. controls (580.91,285.14) and (585.42,280.63) .. (590.99,280.63) -- (660.43,280.63) .. controls (665.99,280.63) and (670.51,285.14) .. (670.51,290.71) -- (670.51,320.95) .. controls (670.51,326.51) and (665.99,331.03) .. (660.43,331.03) -- (590.99,331.03) .. controls (585.42,331.03) and (580.91,326.51) .. (580.91,320.95) -- cycle ;
\draw  [color={rgb, 255:red, 255; green, 255; blue, 255 }  ,draw opacity=1 ][fill={rgb, 255:red, 248; green, 231; blue, 28 }  ,fill opacity=0.4 ] (441.54,377.38) .. controls (441.54,376.24) and (442.46,375.33) .. (443.59,375.33) -- (456.24,375.33) .. controls (457.37,375.33) and (458.29,376.24) .. (458.29,377.38) -- (458.29,383.53) .. controls (458.29,384.66) and (457.37,385.58) .. (456.24,385.58) -- (443.59,385.58) .. controls (442.46,385.58) and (441.54,384.66) .. (441.54,383.53) -- cycle ;
\draw  [color={rgb, 255:red, 255; green, 255; blue, 255 }  ,draw opacity=1 ][fill={rgb, 255:red, 74; green, 144; blue, 226 }  ,fill opacity=0.4 ] (576.04,357.13) .. controls (576.04,355.99) and (576.96,355.08) .. (578.09,355.08) -- (590.74,355.08) .. controls (591.87,355.08) and (592.79,355.99) .. (592.79,357.13) -- (592.79,363.28) .. controls (592.79,364.41) and (591.87,365.33) .. (590.74,365.33) -- (578.09,365.33) .. controls (576.96,365.33) and (576.04,364.41) .. (576.04,363.28) -- cycle ;
\draw  [color={rgb, 255:red, 255; green, 255; blue, 255 }  ,draw opacity=1 ][fill={rgb, 255:red, 126; green, 211; blue, 33 }  ,fill opacity=0.4 ] (575.79,377.13) .. controls (575.79,375.99) and (576.71,375.08) .. (577.84,375.08) -- (590.49,375.08) .. controls (591.62,375.08) and (592.54,375.99) .. (592.54,377.13) -- (592.54,383.28) .. controls (592.54,384.41) and (591.62,385.33) .. (590.49,385.33) -- (577.84,385.33) .. controls (576.71,385.33) and (575.79,384.41) .. (575.79,383.28) -- cycle ;
\draw  [color={rgb, 255:red, 255; green, 255; blue, 255 }  ,draw opacity=1 ][pattern=_bpeyddhwz,pattern size=6pt,pattern thickness=0.75pt,pattern radius=0pt, pattern color={rgb, 255:red, 155; green, 155; blue, 155}] (442.04,357.13) .. controls (442.04,355.99) and (442.96,355.08) .. (444.09,355.08) -- (456.74,355.08) .. controls (457.87,355.08) and (458.79,355.99) .. (458.79,357.13) -- (458.79,363.28) .. controls (458.79,364.41) and (457.87,365.33) .. (456.74,365.33) -- (444.09,365.33) .. controls (442.96,365.33) and (442.04,364.41) .. (442.04,363.28) -- cycle ;

\draw (263.81,290) node [anchor=north west][inner sep=0.75pt]    {$\mathcal{E}$};
\draw (311,305) node [anchor=north west][inner sep=0.75pt]   [align=left] {Engineering};
\draw (324,290) node [anchor=north west][inner sep=0.75pt]   [align=left] {Feature};
\draw (405.11,276.46) node [anchor=north west][inner sep=0.75pt]    {$\mathcal{E} '$};
\draw (468,275) node [anchor=north west][inner sep=0.75pt]   [align=left] {Policy};
\draw (608,290) node [anchor=north west][inner sep=0.75pt]   [align=left] {Policy};
\draw (596,305) node [anchor=north west][inner sep=0.75pt]   [align=left] {Evaluation};
\draw (454,290) node [anchor=north west][inner sep=0.75pt]   [align=left] {Generation};
\draw (535.21,274.61) node [anchor=north west][inner sep=0.75pt]    {$\hat{\pi }^{*}$};
\draw (679.83,287.23) node [anchor=north west][inner sep=0.75pt]    {$\eta \left(\hat{\pi }^{*}\right)$};
\draw (462,373) node [anchor=north west][inner sep=0.75pt]   [align=left] {Automatic Unit};
\draw (596,353) node [anchor=north west][inner sep=0.75pt]   [align=left] {Tunable Unit};
\draw (596,373) node [anchor=north west][inner sep=0.75pt]   [align=left] {Fixed Unit};
\draw (462,353) node [anchor=north west][inner sep=0.75pt]   [align=left] {Not used};

\end{tikzpicture}